\documentclass{article}

\usepackage[final]{neurips_2025}

\usepackage[utf8]{inputenc} 
\usepackage[T1]{fontenc}    
\usepackage{hyperref}       
\usepackage{url}            
\usepackage{booktabs}       
\usepackage{amsfonts}       
\usepackage{nicefrac}       
\usepackage{microtype}      
\usepackage[table]{xcolor}
\usepackage{graphicx}
\usepackage{caption}
\usepackage{subcaption}
\usepackage{amsmath}
\usepackage{amssymb}
\usepackage{mathtools}
\usepackage{amsthm}
\usepackage[capitalize,noabbrev]{cleveref}

\usepackage{amsmath,amsfonts,bm}

\def\eqref#1{equation~\ref{#1}}

\def\1{\bm{1}}

\DeclareMathAlphabet{\mathsfit}{\encodingdefault}{\sfdefault}{m}{sl}
\SetMathAlphabet{\mathsfit}{bold}{\encodingdefault}{\sfdefault}{bx}{n}

\newcommand{\R}{\mathbb{R}}

\usepackage{comment}
\usepackage{wrapfig}
\usepackage{multirow}
\definecolor{cornflowerblue}{rgb}{0.39, 0.58, 0.93}
\hypersetup{
    colorlinks=true,
    linkcolor=cornflowerblue,
    filecolor=magenta,      
    urlcolor=teal,
    citecolor=cornflowerblue,
    pdftitle={A title},
    pdfpagemode=FullScreen,
    }

\title{Dense Backpropagation Improves Training \\
for Sparse Mixture-of-Experts}

\author{\textbf{Ashwinee Panda}$^{1}$\thanks{* denotes equal contribution. Correspondence to: ashwinee@umd.edu} \quad \textbf{Vatsal Baherwani}$^{1*}$ \quad \textbf{Zain Sarwar}$^{2,3}$ \quad \textbf{Benjamin Therien}$^{2,4}$ \\ \textbf{Sambit Sahu}$^{2}$ \quad \textbf{Tom Goldstein}$^{1}$ \quad \textbf{Supriyo Chakraborty}$^{2}$ \\
$^{1}$University of Maryland \quad \ $^{2}$Capital One \quad \ $^{3}$University of Chicago  \quad $^4$\ Mila -- Quebec AI Institute
}

\begin{document}

\maketitle

\begin{abstract}
    Mixture of Experts (MoE) pretraining is more scalable than dense Transformer pretraining, because MoEs learn to route inputs to a sparse set of their feedforward parameters. However, this means that MoEs only receive a sparse backward update, leading to training instability and suboptimal performance. We present a lightweight approximation method that gives the MoE router a dense gradient update while continuing to sparsely activate its parameters. Our method, which we refer to as Default MoE, substitutes missing expert activations with default outputs consisting of an exponential moving average of expert outputs previously seen over the course of training. This allows the router to receive signals from every expert for each token, leading to significant improvements in training performance. Our Default MoE outperforms standard TopK routing in a variety of settings without requiring significant computational overhead.
    
\end{abstract}
\section{Introduction}
The sparsely activated Mixture-of-Experts~(MoE) Transformer architecture~\citep{shazeer2017outrageouslylargeneuralnetworks} has been used by many industry  deployments~\citep{deepseekv3, grok, DBRX, jiang2024mixtralexperts, arctic, deepseekai2024deepseekv2strongeconomicalefficient} because MoEs have been shown to scale even better than dense Transformers~\citep{clark2022unifiedscalinglawsrouted, du2022glam, lepikhin2020gshard, fedus2022switchtransformersscalingtrillion}. 
MoEs learn a \textit{routing} function that selectively activates the Top-K subset of their modules, or \textit{experts}, most relevant to a given input. This conditionally sparse activation~\citep{jacobs1991, jordanjacobs1994} allows us to multiplicatively increase the model parameter count without significantly increasing the cost of training or inference.
At train time, the sparse router enables very large MoEs to be trained with relatively little computation per token, but it also presents a challenge. The router does not receive a gradient update from experts that it does not activate. This may slow down learning, as the gradient update is unable to adjust the router to promote the optimal expert for each token. 
This can be corrected by activating all experts during training, enabling all experts and routing directions to be optimized at once, although this requires much more compute.

\textbf{In this work} we propose \textbf{DefaultMoE}: a new method that strikes a middle ground between dense and sparse routing; our proposed router can receive and differentiate contributions from all experts, while the computational cost remains virtually identical to a standard Top-K router. 
In our proposed method, the router receives contributions from all experts for every token.  However, a forward pass is only computed on the Top-K experts for each token, while other experts contribute a ``default vector" that represents the expected output of a typical token. Our method adds minimal computational overhead compared to a standard Top-K router while improving performance on a range of standard benchmarks for 2 billion total parameter MoEs trained on 160 billion tokens. 
\section{Background \& Related Work}
\textbf{MoEs. }
The MoE layer replaces the feedforward networks (FFN) of Transformers and consists of two components : \textbf{1)} $N$ FFNs (\textit{experts}), $E_0(x), E_1(x), \dots E_N(x)$ and \textbf{2)} a router that assigns tokens to experts. Each input to the MoE layer is processed by $K$ experts where $K < N$, and this is the source of sparsity in MoEs. The $K$ experts are chosen by the router, which is a learnable component that maps each token to a set of weights over the experts. The router performs a linear transformation 
$\R^{d_\text{token}}\to\R^N$ which produces logits; these are normalized using softmax, resulting in a probability distribution over the experts. With the router's linear transformation parameterized by a matrix $W$, we can represent the expert weights $\pi$ in the following way:

\begin{equation}\label{expertweights}
    \pi\in\R^N = \text{Softmax}(Wx)
\end{equation}


Once we have these expert weights, we apply a routing function to decide which of $K$ experts to route and process this token through.
We consider Top-K routing because it is the most popular.

\textbf{Top-K routing. }
A standard method to select $K$ out of $N$ experts given the expert weights is to select the experts corresponding to the $K$ highest weights. Top-K routing~\citep{fedus2022switchtransformersscalingtrillion} passes the token to the $K$ selected experts and averages the expert outputs using these weights to produce the final output. Experts not selected by the Top-K routing function do not process the token, and this introduces sparsity in MoEs. 
By representing the $K$ chosen experts as the set $\mathcal{A}$, we can express the output of the MoE layer as an average of expert outputs weighted by the router scores:
\begin{equation}\label{moeoutput}
    y = \Sigma_{i \in \mathcal{A}}\pi_{i}E_{i}(x).
\end{equation}    

The expert weights serve two roles. They are used by the routing function to decide which of the $K$ experts to process a token through, and also provide the weights for combining the expert outputs. 
Top-K routing makes the MoE layer desirable for training large, compute-efficient neural networks. It allows models to be scaled up, by way of increasing the total number of experts, while keeping the compute per token constant (as it is a function of $K$ and not $N$).



\textbf{The Router Gradient.}
Consider the gradient of the MoE layer's output $y$ with respect to the router parameters $W$. We express $y$ as a function of $W$ by combining \cref{expertweights} and \cref{moeoutput}. With the chain rule, we can backpropagate through this function by considering the gradient at each respective step: 
\begin{equation}\label{chainrule}
    \frac{\partial y}{\partial W} = \frac{\partial y}{\partial \pi}\frac{\partial \pi}{\partial W}
\end{equation}
 
The second term in \cref{chainrule}, $\frac{\partial \pi}{\partial W}$, is straightforward to compute because the steps in \cref{expertweights} are easily differentiable, as they consist of linear operations and activations. But the first term, $\frac{\partial y}{\partial \pi}$, is not differentiable because in \cref{moeoutput} Top-K expert selection transforms the continuous router weights $\pi\in\R^N$ into a discrete set of selected experts $\mathcal{A}$ with $\binom{N}{K}$ possible values.
One way to address backpropagation of nondifferentiable operations is to use the straight-through estimator \citep{bengio2013estimatingpropagatinggradientsstochastic}, which treats the operation as the identity function. With straight-through we bypass the Top-K routing function and \cref{moeoutput} becomes the dot product between $\pi$ and the vector of all $E_i(x)$ with the following gradient:

\begin{equation}\label{densegradient}
    \frac{\partial y}{\partial \pi} = 
    \begin{bmatrix}
        E_1(x), &
        E_2(x) &
        \cdots & 
        E_N(x)
    \end{bmatrix}^T
\end{equation}

This \textit{dense} gradient requires the output of \textit{all} of the experts for a given token. Passing a token through all the experts will destroy the sparsity of the MoE layer, thus impeding the scalability of this architecture.
In this work, we develop a method for applying the straight-through estimator while maintaining the sparsity of the MoE layer by substituting the non-activated expert outputs in the dense gradient term with a {\em default} vector. This default vector is a running average of previously computed expert values, and it approximates the missing expert output without incurring the cost of a forward pass. Notably, this enables non-activated experts to contribute to the router's gradient update.

\begin{figure*}
\centering
\begin{subfigure}{0.5\textwidth}
  \centering
  \includegraphics[width=0.9\linewidth]{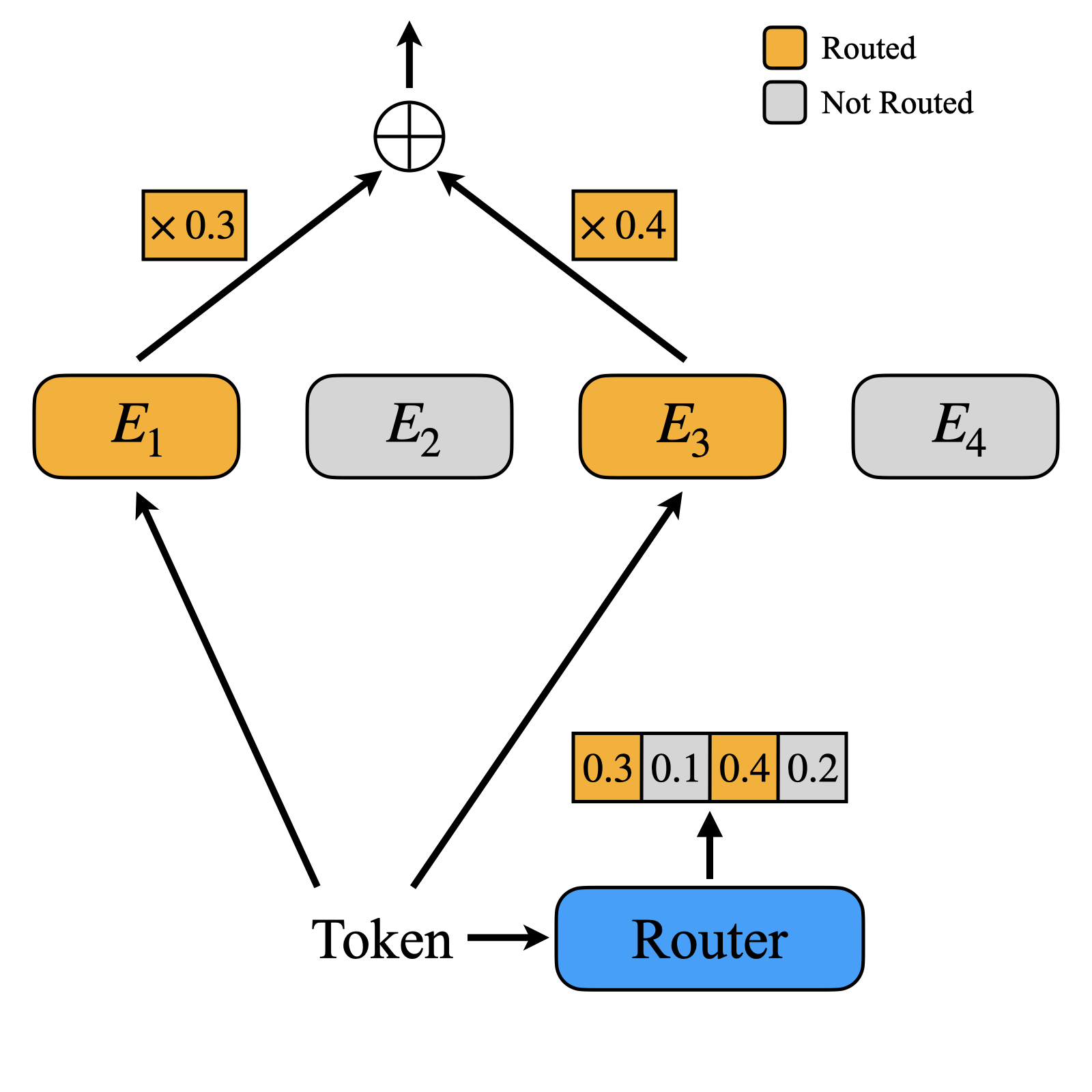}
  \caption{Original Router}
  \label{fig:origrouter}
\end{subfigure}%
\begin{subfigure}{0.5\textwidth}
  \centering
  \includegraphics[width=0.9\linewidth]{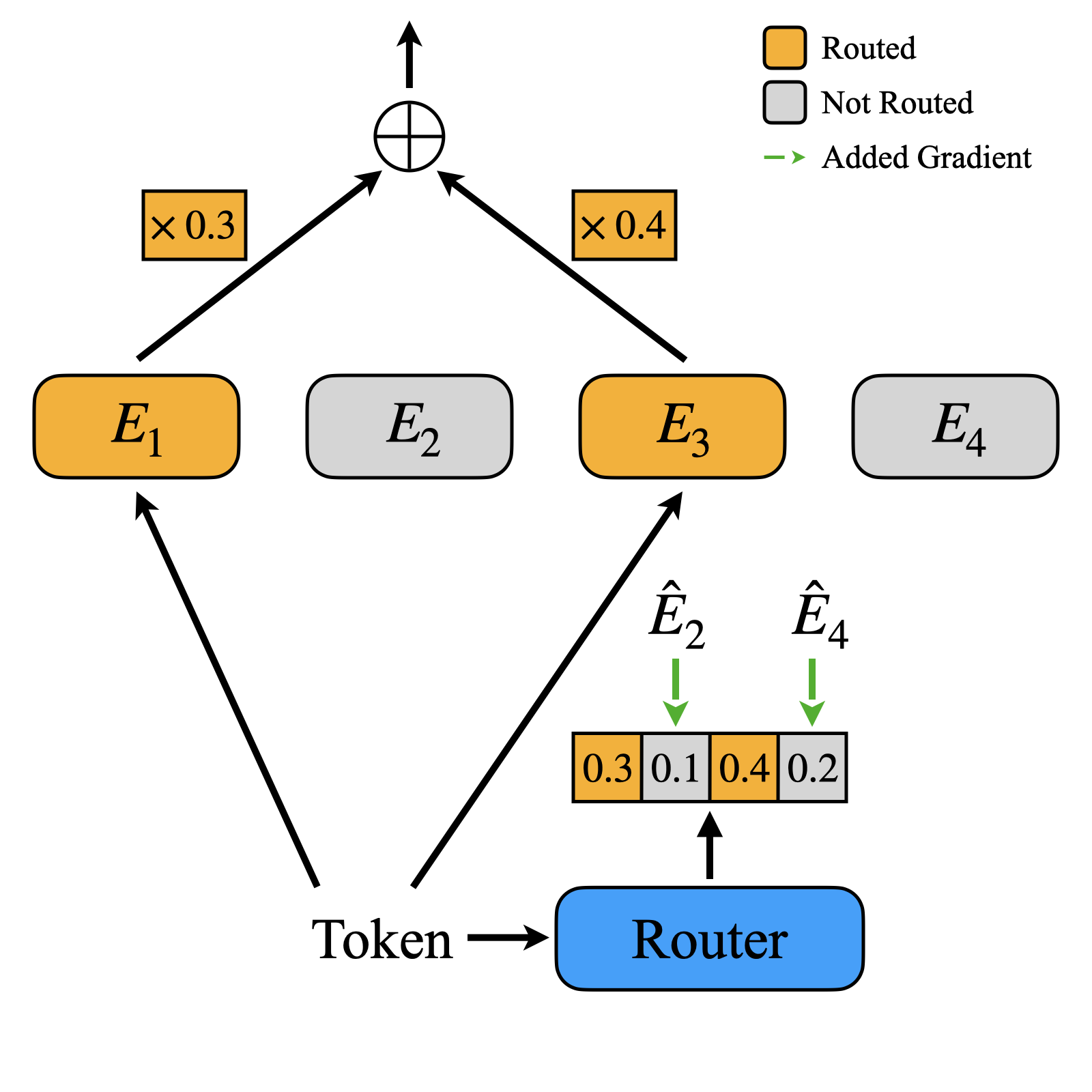}
  \caption{Dense Approximation Router (ours)}
  \label{fig:ourrouter}
\end{subfigure}
\caption{\textbf{Overview of Routing with Dense Approximations}. The original MoE router only receives gradients corresponding to experts the token is routed to, because there is no output from other experts. Our approach provides the router with a complete (dense) gradient by letting non-activated experts contribute a {\em default} vector that approximates its output without the cost of a forward pass. As indicated by the dashed green arrows, the approximated gradients are not actually connected to the token in the computation graph; instead, they are artificially applied in the backward pass.}
\label{fig:router}
\end{figure*}

\section{Designing Dense Backpropagation}


In this section we design a new MoE router that can receive a dense gradient without requiring additional forward passes through experts. \cref{densegradient} describes the dense gradient of the router for an individual token. Each expert output $E_i(x)$ in the router gradient corresponds to updating the embedding for expert $i$ in the routing layer (i.e. the $i$th row of the router parameters $W$). In a standard sparse MoE, the gradient only includes expert outputs $E_i$ for the experts $i\in\mathcal{A}$ selected by the Top-K routing function. The other terms are essentially $0$, which means that non-activated experts can not influence the router's update for a token. 

Our goal is to \textit{approximate} the dense gradient in \cref{densegradient} without directly computing the missing terms $E_{i'}(x)$ for $i'\notin\mathcal{A}$. This enables the router to receive a signal from all experts in its gradient update, without actually activating all of these experts. The router can then consider information from all experts when learning to route tokens instead of being limited to only receiving feedback from experts that were activated. As a result, we expect an improved router which learns to allocate tokens to different experts more effectively, leading to improved MoE training performance.

\cref{fig:router} presents an overview of our method for updating the router. Since we wish to avoid additional computational overhead, we fill in the missing terms in \cref{densegradient} with estimates $\hat{E_i}$ for non-activated experts $E_i$. This simple addition ensures that the router updates its weights for all experts instead of only those selected by the Top-K routing.

\subsection{Approximating Missing Expert Outputs}
Backpropagating through the sparse MoE activation in \cref{moeoutput} does not align with the router's true dense gradient specified in \cref{densegradient}. Specifically, the gradient in TopK routing leads to an error based on the non-activated experts. Let us consider the gradient of the router parameters $W$ with respect to the loss $\mathcal{L}$. For clarity, we examine the gradient term for a single input $x$; our approach also translates to practically training the router with a batch of inputs.

Using the chain rule, we decompose the router gradient to isolate the nondifferentiable $\frac{\partial y}{\partial \pi}$, similar to \cref{chainrule}:
\begin{equation}
    \frac{\partial \mathcal{L}}{\partial W} = \frac{\partial \mathcal{L}}{\partial y}\frac{\partial y}{\partial \pi}\frac{\partial \pi}{\partial W}
\end{equation}

In standard Top-K routing, the gradient term effectively uses zero for non-activated experts:
\begin{equation}
    \frac{\partial y}{\partial \pi_i} = \begin{cases}
        E_i(x) & \text{if } i \in \mathcal{A} \\
        0 & \text{if } i \notin \mathcal{A}
    \end{cases}
\end{equation}

The error is introduced by the missing terms in $\frac{\partial y}{\partial \pi}$:
\begin{equation}
    \epsilon_{\text{TopK}} = \frac{\partial \mathcal{L}}{\partial y} \sum_{i' \notin \mathcal{A}} E_{i'}(x)\frac{\partial \pi}{\partial W}
\end{equation}

To address this error, we need a meaningful estimate for $E_{i'}(x)$ without activating the non-activated experts $E_{i'}$. Our approach to this involves maintaining a \textit{default vector} for each expert $\hat{E}_i = \mathbb{E}_x[E_i(x)]$ that represents the expected value of the expert output. In practice, we will compute this estimate by taking the sample average of existing expert outputs $E_i(x')$ from other tokens $x'$ for which $E_i$ was actually activated. Considering this estimator, our gradient term for non-activated experts is now nonzero:
\begin{equation}
    \frac{\partial y}{\partial \pi_i} = \begin{cases}
        E_i(x) & \text{if } i \in \mathcal{A} \\
        \hat{E}_i & \text{if } i \notin \mathcal{A}
    \end{cases}
\end{equation}

Note that $\hat{E}_i$ is not a function of $x$; it is universally applied to complete the router gradient for all such tokens in the batch. The gradient error term with this new default vector is as follows:
\begin{equation}
    \epsilon_{\text{default}} = \frac{\partial \mathcal{L}}{\partial y} \sum_{i \notin \mathcal{A}} (E_i(x) - \mathbb{E}_x[E_i(x)])\frac{\partial \pi}{\partial W}
\end{equation}

The error term corresponding to $\frac{\partial y}{\partial \pi}$, $\sum_{i \notin \mathcal{A}} (E_i(x) - \mathbb{E}_x[E_i(x)])$, is $0$ in expectation for all $i$. As a result, our default vector method corrects the gradient error by filling in missing expert activations with the mean expert output.

\subsection{Default Expert Outputs with Exponential Moving Averages}
To enable dense backpropagation without sacrificing the computational efficiency of sparse MoE forward passes, we propose \textbf{Default MoE}. Default MoE provides a default value for the outputs of non-selected experts using exponential moving averages (EMAs). The EMA accounts for the fact that experts are being updated, and thus their outputs for inputs $x$ will change over the course of training. For each expert $E_i$, we maintain an EMA of its average output:

\begin{equation}
    \hat{E}_i^{(t)} = \beta \hat{E}_i^{(t-1)} + (1-\beta) \overline{E_i(x)}
\end{equation}

where \(\beta \in [0,1]\) is the decay rate and $\overline{E_i(x)}$ is the sample average of expert outputs $E_i$ for all tokens $x$ for which expert $i$ was activated. We compute $\overline{E_i(x)}$ during the forward pass by collecting all such expert outputs $E_i(x)$ and taking the average. This operation is computationally trivial, since the outputs themselves are already computed as part of the standard MoE forward pass. Thus, the EMA serves as a lightweight proxy for the expert's expected output.

During the forward pass, for a given input $x$, and an expert \(i\)~(where there are \(N\) total experts) we compute the output of the MoE as:
\begin{equation}\label{defaultvector}
    y = \sum_{i=1}^N \pi_i \cdot \begin{cases}
        E_i(x) & \text{if } i \in \text{TopK}(\pi) \\
        \hat{E}_i^{(t)} & \text{otherwise}
    \end{cases}
\end{equation}

Specifically, we first compute the forward pass for activated experts $E_i(x)$ for each token. Then, we perform the EMA update step using these outputs. After applying the EMA update, we use $\hat{E}_i^{(t)}$ to approximate missing outputs for non-activated experts. 
Our method ensures that computed expert outputs at the current step are factored into the EMA update before applying the EMA as a substitute for other tokens that did not activate the expert. 

This formulation allows the router to receive meaningful gradients for all experts while maintaining the computational benefits of sparse activation. The EMA provides a reasonable approximation of what non-activated experts would have computed, based on their historical outputs for other tokens. Importantly, this approximation requires only $\mathcal{O}(1)$ additional memory per expert and requires minimal additional forward pass computation for the EMA update.


This approach enables dense backpropagation through the router while preserving the sparse computational pattern that makes MoE architectures efficient. The router receives gradient information about all possible routing decisions, not just the selected experts. We now evaluate DefaultMoE.
\section{Evaluation\label{sec:eval}}

We describe the experimental setup in~\cref{subsec:experimental_setup}, benchmark our Default MoE in~\cref{subsec:main_results}, show that our improvements hold across multiple configurations in~\cref{subsec:ablations}, discuss prior work in~\cref{subsec:prior_work_comparisons}, analyze our Default MoE in~\cref{subsec:analysis}, and find in~\cref{subsec:efficiency} that our method does not significantly impact throughput or the memory footprint of training, so the improvements we observe are -as far as we can tell- a free lunch. We conclude with a cohesive explanation of why Default MoE beats TopKMoE in~\cref{sec:discussion}.

\subsection{Experimental Setup}\label{subsec:experimental_setup}
\textbf{Model Architectures. } 
We use \texttt{NcK} to refer to a model with $N$ total experts and $K$ active experts. We train both standard and finegrained~\citep{deepseekai2024deepseekv2strongeconomicalefficient} MoEs. All of our models have 1.96 billion total parameters, 366 million of which are non-MoE parameters. This leaves 1.6B MoE parameters, and the number of active parameters depends on sparsity.
All parameter counts are specified in~\cref{tab:param-counts}.
We ablate the model architecture, hidden dim, number of total experts, and number of active experts in~\cref{subsec:ablations}.

\begin{table}[hbtp]
    \centering
    \begin{tabular}{c|ccccc}
    \toprule
         MoE Config. & 8c1 & 8c2 & 32c1 & 32c2 & 32c4  \\
         \midrule
         Active Params. & 565M & 764M & 416M & 466M & 565M \\
         \bottomrule
    \end{tabular}
    \caption{Active parameter counts for each MoE configuration; the total parameter count is 1.96B.}
    \label{tab:param-counts}
\end{table}
\textbf{Dataset.} We train on 
FineWeb-Edu~\citep{lozhkov2024fineweb-edu}~(for~\cref{tab:benchmarks} and~\cref{fig:tokens-to-target-ppl}) and FineWeb~\citep{penedo2024finewebdatasetsdecantingweb}~(for all other plots and results) with the Llama3 tokenizer~\citep{dubey2024llama3herdmodels}. For our main results, we train for 160B tokens, which is a tokens-per-parameter ratio of \(\approx 283\); at this level of overtraining, many spurious improvements should wash out, so we can be confident that any improvements we observe are real gains.

\href{https://anonymous.4open.science/r/default-moe-6C74/}{We have open-sourced our training code}.
All hyperparameters for the models trained in~\cref{tab:benchmarks} can be found in our \href{https://anonymous.4open.science/r/default-moe-6C74/configs/default-moe-2B.yml}{config file}, and all hyperparameters are the same between the baseline and our DefaultMoE. All further experimental details can be found in~\cref{sec:appen_setup}.
\subsection{Main Results}\label{subsec:main_results}

\definecolor{forestgreen}{HTML}{228B22} \begin{table}[h] \centering \resizebox{\textwidth}{!}{ 
\begin{tabular}{@{}l|ccc|ccc|ccc|ccc|c@{}}
\toprule
\multirow{3}{*}{Benchmark} & \multicolumn{6}{c|}{8c1} & \multicolumn{6}{c|}{32c4} & \multirow{3}{*}{\begin{tabular}{@{}c@{}}Random\\Baseline\\(Score)\end{tabular}} \\
\cmidrule(lr){2-4} \cmidrule(lr){5-7} \cmidrule(lr){8-10} \cmidrule(lr){11-13}
 & \multicolumn{3}{c|}{Score} & \multicolumn{3}{c|}{Improvement (\%)} & \multicolumn{3}{c|}{Score} & \multicolumn{3}{c|}{Improvement (\%)} &  \\
 & TopK & Default & Diff (\%) & TopK & Default & Diff (\%) & TopK & Default & Diff (\%) & TopK & Default & Diff (\%) &  \\
\midrule
LogiQA & 26.0$_{1.7}$ & \cellcolor{gray!20}26.9$_{1.7}$ & \textcolor{forestgreen}{+3.6\%} & 3.8 & \cellcolor{gray!20}7.5 & \textcolor{forestgreen}{+96.0\%} & 29.0$_{1.8}$ & \cellcolor{gray!20}28.3$_{1.8}$ & \textcolor{red}{-2.6\%} & 16.1 & \cellcolor{gray!20}13.1 & \textcolor{red}{-19.0\%} & 25.0 \\
MathQA & 26.2$_{0.8}$ & \cellcolor{gray!20}25.8$_{0.8}$ & \textcolor{red}{-1.8\%} & 4.9 & \cellcolor{gray!20}3.0 & \textcolor{red}{-38.1\%} & 25.0$_{0.8}$ & \cellcolor{gray!20}26.1$_{0.8}$ & \textcolor{forestgreen}{+4.3\%} & 0.1 & \cellcolor{gray!20}4.4 & \textcolor{forestgreen}{+4266.7\%} & 25.0 \\
MMLU & 31.8$_{0.4}$ & \cellcolor{gray!20}32.3$_{0.4}$ & \textcolor{forestgreen}{+1.6\%} & 27.3 & \cellcolor{gray!20}29.4 & \textcolor{forestgreen}{+7.5\%} & 33.7$_{0.4}$ & \cellcolor{gray!20}32.5$_{0.4}$ & \textcolor{red}{-3.6\%} & 34.7 & \cellcolor{gray!20}29.9 & \textcolor{red}{-14.0\%} & 25.0 \\
OpenBookQA & 37.2$_{2.2}$ & \cellcolor{gray!20}38.2$_{2.2}$ & \textcolor{forestgreen}{+2.7\%} & 48.8 & \cellcolor{gray!20}52.8 & \textcolor{forestgreen}{+8.2\%} & 39.6$_{2.2}$ & \cellcolor{gray!20}39.0$_{2.2}$ & \textcolor{red}{-1.5\%} & 58.4 & \cellcolor{gray!20}56.0 & \textcolor{red}{-4.1\%} & 25.0 \\
Lambada & 38.6$_{0.7}$ & \cellcolor{gray!20}41.2$_{0.7}$ & \textcolor{forestgreen}{+6.6\%} & 3761.8 & \cellcolor{gray!20}4016.0 & \textcolor{forestgreen}{+6.8\%} & 44.1$_{0.7}$ & \cellcolor{gray!20}43.8$_{0.7}$ & \textcolor{red}{-0.6\%} & 4307.1 & \cellcolor{gray!20}4281.9 & \textcolor{red}{-0.6\%} & 1.0\footnotemark{} \\
SocialIQA & 39.7$_{1.1}$ & \cellcolor{gray!20}41.0$_{1.1}$ & \textcolor{forestgreen}{+3.2\%} & 20.3 & \cellcolor{gray!20}24.2 & \textcolor{forestgreen}{+19.1\%} & 41.1$_{1.1}$ & \cellcolor{gray!20}40.7$_{1.1}$ & \textcolor{red}{-1.1\%} & 24.7 & \cellcolor{gray!20}23.3 & \textcolor{red}{-5.7\%} & 33.0 \\
HellaSwag & 40.4$_{0.5}$ & \cellcolor{gray!20}41.2$_{0.5}$ & \textcolor{forestgreen}{+2.0\%} & 61.6 & \cellcolor{gray!20}64.9 & \textcolor{forestgreen}{+5.4\%} & 42.9$_{0.5}$ & \cellcolor{gray!20}43.3$_{0.5}$ & \textcolor{forestgreen}{+0.9\%} & 71.4 & \cellcolor{gray!20}73.0 & \textcolor{forestgreen}{+2.2\%} & 25.0 \\
ARC & 45.7$_{0.9}$ & \cellcolor{gray!20}47.4$_{0.9}$ & \textcolor{forestgreen}{+3.7\%} & 82.8 & \cellcolor{gray!20}89.6 & \textcolor{forestgreen}{+8.3\%} & 52.0$_{0.9}$ & \cellcolor{gray!20}49.9$_{0.9}$ & \textcolor{red}{-4.2\%} & 108.2 & \cellcolor{gray!20}99.5 & \textcolor{red}{-8.1\%} & 25.0 \\
Winogrande & 53.6$_{1.4}$ & \cellcolor{gray!20}54.8$_{1.4}$ & \textcolor{forestgreen}{+2.2\%} & 7.2 & \cellcolor{gray!20}9.6 & \textcolor{forestgreen}{+33.0\%} & 55.5$_{1.4}$ & \cellcolor{gray!20}56.3$_{1.4}$ & \textcolor{forestgreen}{+1.4\%} & 11.0 & \cellcolor{gray!20}12.5 & \textcolor{forestgreen}{+14.4\%} & 50.0 \\
PubMedQA & 55.2$_{2.2}$ & \cellcolor{gray!20}52.4$_{2.2}$ & \textcolor{red}{-5.1\%} & 67.3 & \cellcolor{gray!20}58.8 & \textcolor{red}{-12.6\%} & 54.4$_{2.2}$ & \cellcolor{gray!20}57.4$_{2.2}$ & \textcolor{forestgreen}{+5.5\%} & 64.8 & \cellcolor{gray!20}73.9 & \textcolor{forestgreen}{+14.0\%} & 33.0 \\
BoolQ & 58.5$_{0.9}$ & \cellcolor{gray!20}62.0$_{0.8}$ & \textcolor{forestgreen}{+6.1\%} & 17.0 & \cellcolor{gray!20}24.1 & \textcolor{forestgreen}{+41.7\%} & 62.8$_{0.8}$ & \cellcolor{gray!20}63.1$_{0.8}$ & \textcolor{forestgreen}{+0.4\%} & 25.6 & \cellcolor{gray!20}26.2 & \textcolor{forestgreen}{+2.1\%} & 50.0 \\
PIQA & 70.8$_{1.1}$ & \cellcolor{gray!20}71.9$_{1.0}$ & \textcolor{forestgreen}{+1.5\%} & 41.7 & \cellcolor{gray!20}43.7 & \textcolor{forestgreen}{+5.0\%} & 72.3$_{1.0}$ & \cellcolor{gray!20}73.1$_{1.0}$ & \textcolor{forestgreen}{+1.1\%} & 44.6 & \cellcolor{gray!20}46.1 & \textcolor{forestgreen}{+3.4\%} & 50.0 \\
SciQ & 86.2$_{1.1}$ & \cellcolor{gray!20}87.8$_{1.0}$ & \textcolor{forestgreen}{+1.9\%} & 244.8 & \cellcolor{gray!20}251.2 & \textcolor{forestgreen}{+2.6\%} & 89.1$_{1.0}$ & \cellcolor{gray!20}90.3$_{0.9}$ & \textcolor{forestgreen}{+1.3\%} & 256.4 & \cellcolor{gray!20}261.2 & \textcolor{forestgreen}{+1.9\%} & 25.0 \\
\midrule
Average & 46.9$_{0.4}$ & \cellcolor{gray!20}47.9$_{0.4}$ & \textcolor{forestgreen}{+2.1\%} & 52.3 & \cellcolor{gray!20}54.9 & \textcolor{forestgreen}{+5.0\%} & 49.4$_{0.4}$ & \cellcolor{gray!20}49.5$_{0.4}$ & \textcolor{forestgreen}{+0.3\%} & 59.7 & \cellcolor{gray!20}59.9 & \textcolor{forestgreen}{+0.4\%} & - \\
\bottomrule
\end{tabular}
} 
\caption{\textbf{Comparison of Pretraining Benchmark Scores for TopK MoE and Default MoE. }After training for 160 billion tokens, Default MoE outperforms TopK MoE across a range of benchmarks for both a standard MoE (right) and a finegrained MoE (left).} \label{tab:benchmarks} 
\end{table} 
\footnotetext{We use the `Random word from passage" as a baseline for LAMBADA \citep{paperno2016lambadadatasetwordprediction}.}

\textbf{Pretraining Benchmarks.} In~\cref{tab:benchmarks} we compare Default MoEs to TopK MoEs. All models are trained for 160B tokens on FineWeb-Edu and have 1.96B params. For both finegrained MoEs~(32c4) and standard MoEs~(8c1), our Default MoEs outperform TopK MoEs on standard benchmarks. We conduct all evaluations with the lm-eval harness~\citep{eval-harness}.


Due to space constraints we defer pretraining curves to the Appendix, but discuss them here.
In~\cref{fig:tokens-to-target-ppl} we compare each MoE's loss curves. Without introducing significant overhead, our method reduces the tokens required to reach a target perplexity of \(12.18\) by \(9\%\). As the model has just \(500M\) active parameters, we are training for significantly more than the compute-optimal number of tokens~\citet{hoffmann2022trainingcomputeoptimallargelanguage}.
Therefore, we can be confident that we are not just seeing our method converge faster to a worse minima; this is a real improvement in terms of both speed of convergence and the quality of the converged model.

\paragraph{Pretraining on FineWeb.}
In~\cref{tab:benchmarks} and~\cref{fig:tokens-to-target-ppl} we pretrain on FineWeb-Edu~\citep{lozhkov2024fineweb-edu}. We also pretrain models on FineWeb~\citep{penedo2024finewebdatasetsdecantingweb}, which is less curated. The rest of the plots in this paper~(our ablations) are on FineWeb, in order to enforce that on some level we are not simply overfitting to the pretraining dataset that we use to collect benchmark scores.
In~\cref{fig:lm-loss-fineweb} we show that when using the learning rate of \(7 \times 10^{-4}\) that we sweep in~\cref{fig:all_ablation}, our DefaultMoE outperforms the baseline TopK. Therefore, our improvements are not due to any unfair hyperparameter tuning when compared to the baseline.

\subsection{Ablations \label{subsec:ablations}}

We now ablate every setting in our experimental setup, from the MoE config. to the learning rate.

\textbf{Tuning \(\beta\).}
We initially tuned \(\beta\) for every configuration independently, and the results of this can be found in~\cref{sec:appen_beta}. We found that more sparse configurations, such as 32c1, required a lower \(\beta=0.65\) whereas a much higher beta of \(\beta=0.999\) was necessary for 32c4 to achieve its best performance. However, we can actually automatically account for the impact of sparsity and granularity on the default vector update by weighting the updates to the default vector by the router logits. When we do this, different values of \(\beta\), that would otherwise be suboptimal, all converge to the same good performance. Therefore, in our final method, the value of \(\beta\) does not seem to matter as shown in ~\cref{fig:scored_beta_ablation}.

\begin{figure}[htbp]
  \centering
  \begin{subfigure}[b]{0.48\textwidth}
    \centering
    \includegraphics[width=\linewidth]{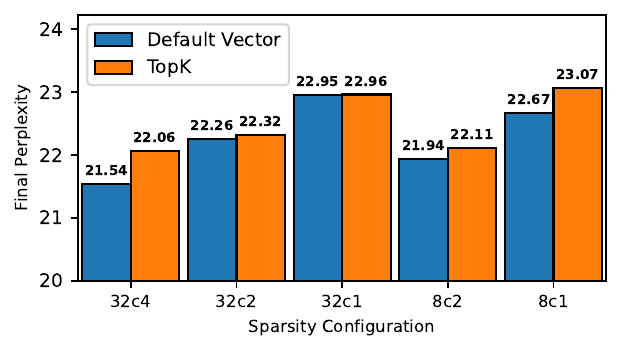}
    \subcaption{MoE Configurations}
    \label{fig:comp_plots_ablation}
  \end{subfigure}
  \hfill
  \begin{subfigure}[b]{0.48\textwidth}
    \centering
    \includegraphics[width=\linewidth]{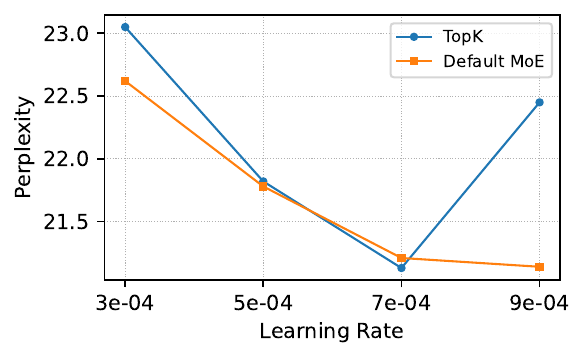}
    \subcaption{Learning Rate Sweep}
    \label{fig:lr_sweep_ablation}
  \end{subfigure}

  \vspace{1em}  

  \begin{subfigure}[b]{0.48\textwidth}
    \centering
    \includegraphics[width=\linewidth]{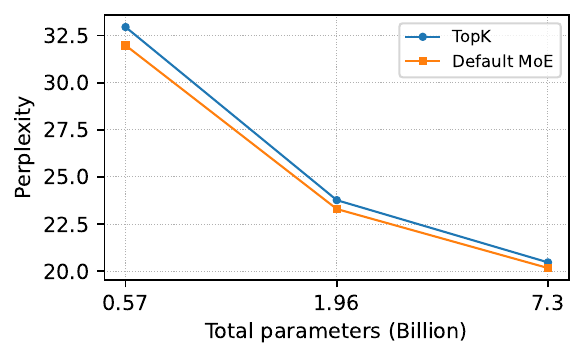}
    \subcaption{Ablation over model parameters}
    \label{fig:hidden_dim_ablation}
  \end{subfigure}
  \hfill
  \begin{subfigure}[b]{0.48\textwidth}
    \centering
    \includegraphics[width=\linewidth]{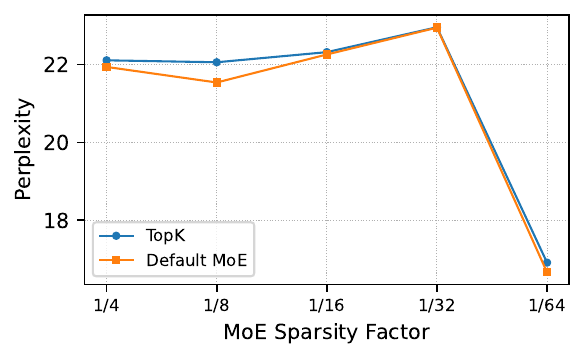}
    \subcaption{Sparsity Factor Ablation}
    \label{fig:new_ablation}
  \end{subfigure}

  \caption{\textbf{Comprehensive Ablations.}
  We report the perplexity after training for 10B tokens for Default MoE and Top-K while varying:
    \textbf{(a)} MoE configuration across five MoE setups.
    \textbf{(b)} Learning rate while keeping the MoE configuration at 8c1.
    \textbf{(c)} Total model size while keeping the MoE configuration at 8c1.
    \textbf{(d)} Sparsity, where the final 1/64 sparsity is for a 7B model, therefore the perplexity is lower than the other sparsity factors that are for 2B models.
  }
  \label{fig:all_ablation}
\end{figure}

\textbf{MoE Architecture Ablations.}
In \cref{fig:all_ablation}(a), we train both a standard TopK model and Default MoE with the configurations 8c1, 8c2, 32c1, 32c2, and 32c4, which correspond to sparsity factors of $1/8$, $1/4$, $1/32$, $1/16$ and $1/8$ respectively. 
Default MoE outperforms a standard TopK MoE at all sparsity configurations. Our improvement is more pronounced for the lowest sparsity (8c2) but still significant for higher sparsity MoEs such as 32c1. Notably, Default MoE takes longer to "warm up" for sparser models; for example, we notice in \cref{fig:comp_plots_ablation_curve} that the 32c1 Default MoE rapidly catches up and surpasses the TopK MoE around the 10 billion token mark, while the 32c2 MoE is clearly ahead of TopK even at 2 billion tokens. 

\textbf{Tuning the Learning Rate.} We want to ensure that we are comparing against a tuned baseline, so we tune the learning rates in~\cref{fig:all_ablation}(b). We use the learning rate that achieves the best performance for the TopK baseline, \(7\times 10^{-4}\), for our results in~\cref{fig:lm-loss-fineweb}. We observe that while it performs better than Default MoE earlier in training (across our learning rate sweep experiments), the results in ~\cref{fig:lm-loss-fineweb} show that even in this case, Default MoE eventually surpasses TopK. Moreover, we note that the best learning rate for the Default MoE is  the larger learning rate of \(9\times 10^{-4}\). By contrast, \(9\times 10^{-4}\) is much too large for the baseline. This indicates that our Default MoE is more stable to train, likely because we are updating the entire router. If we train the baseline with such a large learning rate, we see that a single iteration with a very imbalanced load will lead to a large loss spike; this iteration is very noticeable because, since we train dropless MoEs, it is also slower. We never observe this for the Default MoE. It is unsurprising that our Default MoE's best learning rates may be different from the baseline, but importantly, our method outperforms the baseline at \emph{all} learning rates we consider. 

\textbf{Default MoE Remains Superior As We Increase the Model Size.} ~\cref{fig:hidden_dim_ablation} compares the two methods as we increase the size of the model from \(557M\) total parameters at a hidden dimension of \(512\), to \(7.33B\) total parameters when the hidden dimension is \(2048\). Our method outperforms the baseline across all model sizes. All other results in the paper use a hidden dimension of \(1024\).

We present additional ablations regarding the EMA, including the hyperparameter $\beta$, the EMA initialization, and passing the EMA value forward, in~\cref{sec:appen_beta}.

\subsection{Comparison to Other Routing Methods}\label{subsec:prior_work_comparisons}
We consider three alternatives to TopK routing for comparison: Sparesemixer~\citep{liu2023sparsebackpropagationmoetraining, liu2024gringradientinformedmoe}, ReMoE~\citep{wang2025remoefullydifferentiablemixtureofexperts}, and Loss-Free Balancing~\citep{wang2024auxiliarylossfreeloadbalancingstrategy}.

\textbf{Default MoE Beats Sparsemixer.} ~\citet{liu2023sparsebackpropagationmoetraining} proposes Sparsemixer, which estimates the true router gradient without straight-through. 
~\citet{liu2024gringradientinformedmoe} note that Sparsemixer lags behind TopK~(which our method always outperforms) for the first \(0.5T\) tokens, likely due to the noise that Sparsemixer adds. After 24B tokens, Sparsemixer obtains a validation perplexity of \(19.77\),  worse than both TopK and our method. We present the training curve in~\cref{fig:def_vector_vs_sparsemixer_ablation}.

\textbf{ReMoE Fails to Converge.}
We cannot compare our zero-shot benchmark scores to ReMoE because they train on The Pile, which is unavailable. We train ReMoEs on FineWeb-Edu. ReMoE reports that the first stage of their method is expected to run slowly while their method starts to work, as it is essentially acting as a dense model for the first few hundred steps. Following that, they report that ReMoE exhibits a significant loss spike around 2 billion tokens, after which it recovers and starts to outperform the baseline. In our setting, we indeed were able to replicate the first two results, where ReMoE runs very slowly while acting as a dense model, but the sparsity coefficients start to converge, and the loss spikes at 2 billion tokens. However, the loss never recovered. We invested over 1000 GPU hours into trying to get the ReMoE model to train, but did not have any success. 

\textbf{Loss-Free Balancing Underperforms.}
We first sweep the bias update coefficient, which is the hyperparameter in the loss-free balancing method used in~\citet{wang2024auxiliarylossfreeloadbalancingstrategy}, and find that \(0.0001\) provides the best results. However, after training the 8c1 MoE for 10B tokens, Loss-Free Balancing obtains a validation perplexity of \(24.22\), whereas our Default MoE obtains \(23.48\).

\textbf{Why do Prior Methods Underperform?} Prior work claims to improve upon standard dMoEs, but there is one difference in our experimental setup: we use globally reduced load balancing loss~\citep{qiu2025demonsdetailimplementingload} in all experiments. Corroborating their conclusions, we find that globally reducing the auxiliary loss significantly improves the load balancing, and subsequently the performance, of the baseline, therefore making the improvement of prior methods less noticeable. This is because the ``free lunch'' of slightly improving routing is more or less eaten by the global auxiliary loss reduction, which is trivial to implement and uses little compute, and prior work frequently targets this same free lunch as their source of improvement.~\citet{qiu2025demonsdetailimplementingload} directly evaluate Loss-Free Balancing and report that it underperforms a proper globally reduced auxiliary loss, supporting our results.

\subsection{Empirical Analysis}\label{subsec:analysis}
In~\cref{sec:appen-analysis} we validate that the load balance, expert coactivation, and domain specialization of the trained MoEs do not deteriorate when we implement our DefaultMoE. Our results corroborate those of ~\citet{qiu2025demonsdetailimplementingload}, that even a baseline model trained with globally reduced auxiliary loss will have good load balance and domain specialization. 

\begin{figure}[htbp]
  \centering
  \begin{subfigure}[b]{0.32\textwidth}
    \includegraphics[width=\linewidth]{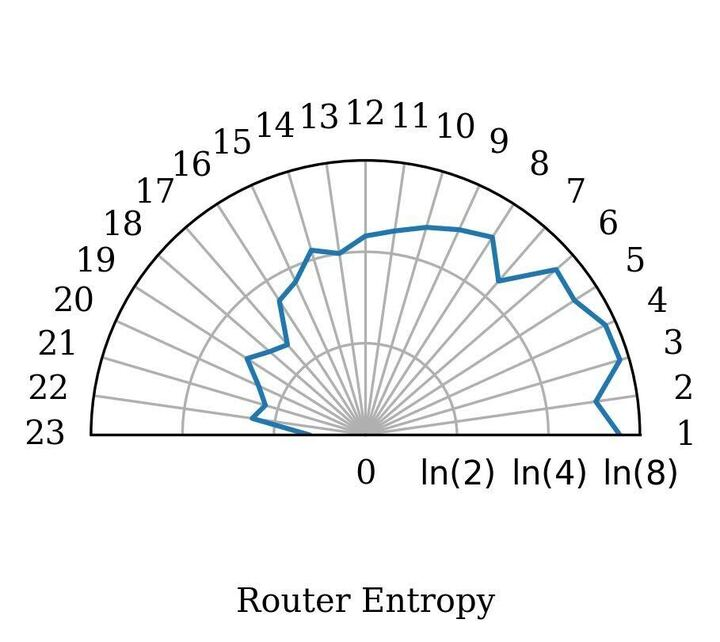}
    \label{fig:router-entropy}
\end{subfigure}
\hfill
  \begin{subfigure}[b]{0.32\textwidth}
    \centering
    \includegraphics[width=\linewidth]{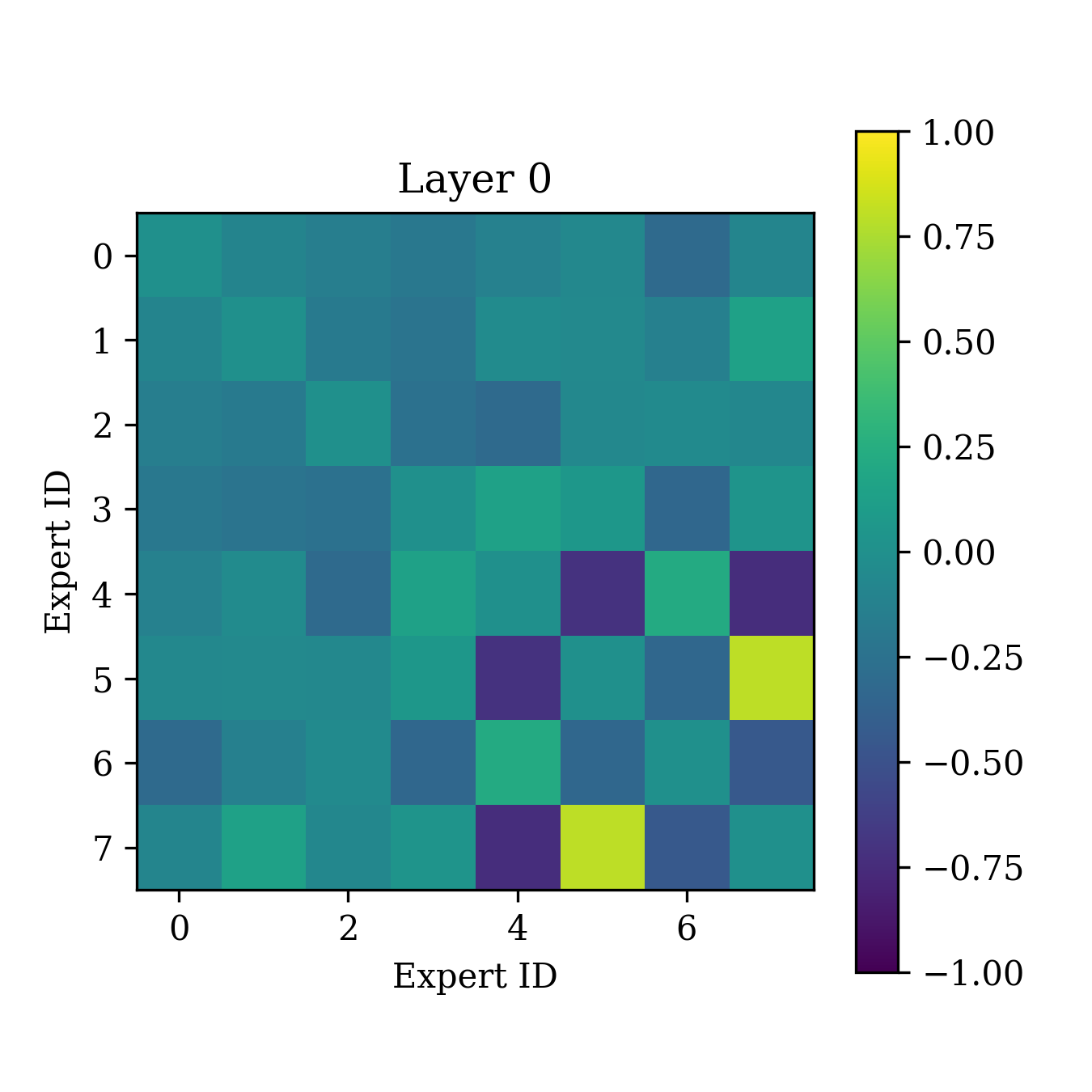}
    \label{fig:similarities-layer0}
  \end{subfigure}
  \hfill
  \begin{subfigure}[b]{0.32\textwidth}
    \centering
    \includegraphics[width=\linewidth]{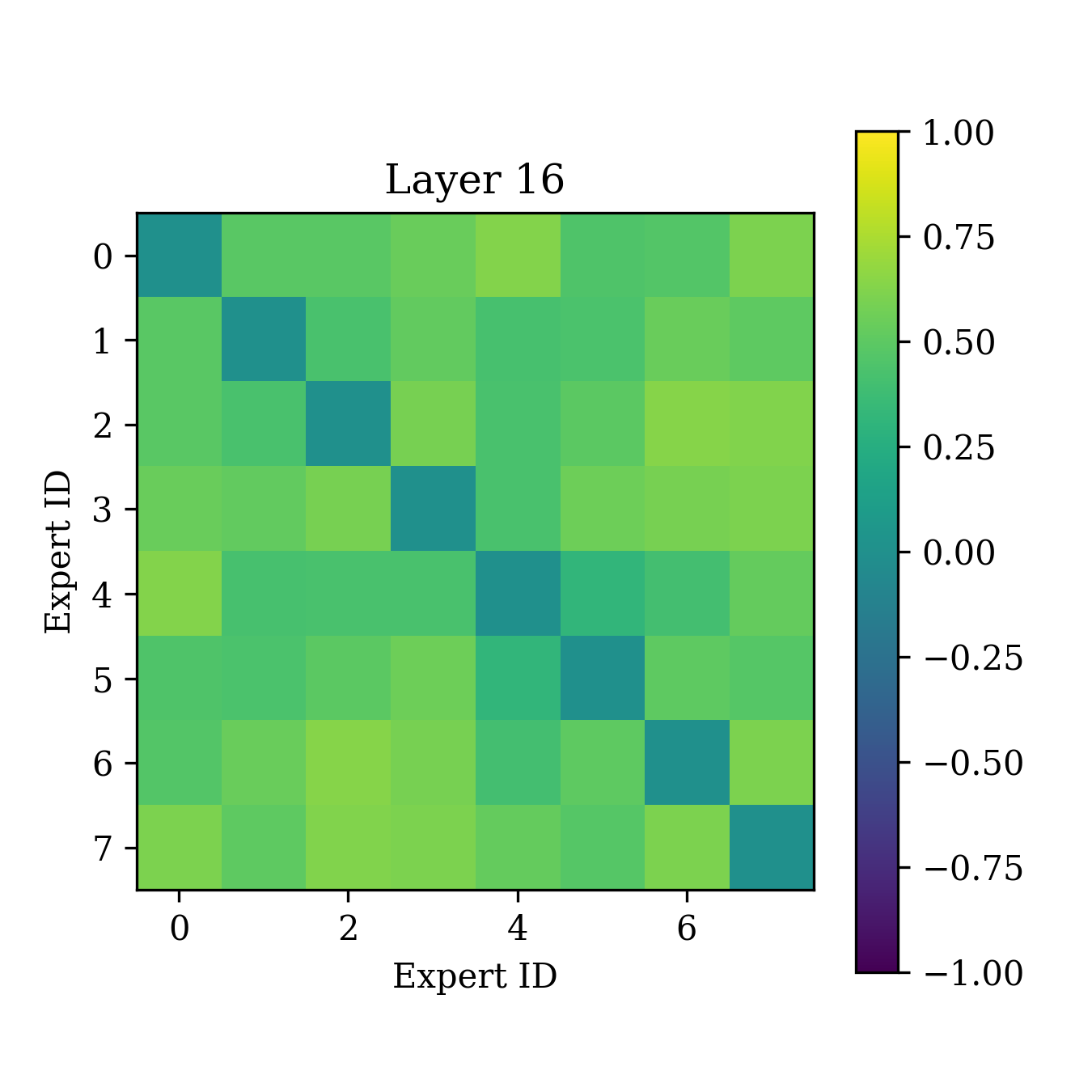}
    \label{fig:similarities-layer16}
  \end{subfigure}
  \caption{\textbf{Analysis.} Left: Router Entropy after training for 10B tokens. Middle, Right: Heatmap of the cosine similarities between default vectors for the 8c1 DefaultMoE at layers 0 and 16.}
  \label{fig:default_vector_similarity}
\end{figure}
In~\cref{fig:default_vector_similarity}(a) we observe that the router entropy is highest for the early layers and consistently goes down, with the final layer's router entropy being very low. Intuitively, the router shouldn't quite know where to send a token at the very start of the model, but the routing decisions should become more and more fixed as we proceed through the model. 

\textbf{Similarity of Default Vectors.}
In~\cref{fig:default_vector_similarity}(a) we plot heatmaps of the cosine similarities between default vectors for experts in different layers of the 8c1 DefaultMoE. The default vectors at layer 0 are not very similar, indicating that the default vectors are learning something nontrivial. However, at layer 16 the default vectors are actually very similar. This follows the trend of the decline in router entropy of~\cref{fig:router-entropy}; if the router is very confident where a token should go, all the default vectors may learn similarly trivial things, but if the router has high entropy, the default vectors should learn faithful nontrivial approximations of the expert activations.

\begin{figure}[htbp]
  \centering
  \begin{subfigure}[b]{0.40\textwidth}
    \centering
    \includegraphics[width=\linewidth]{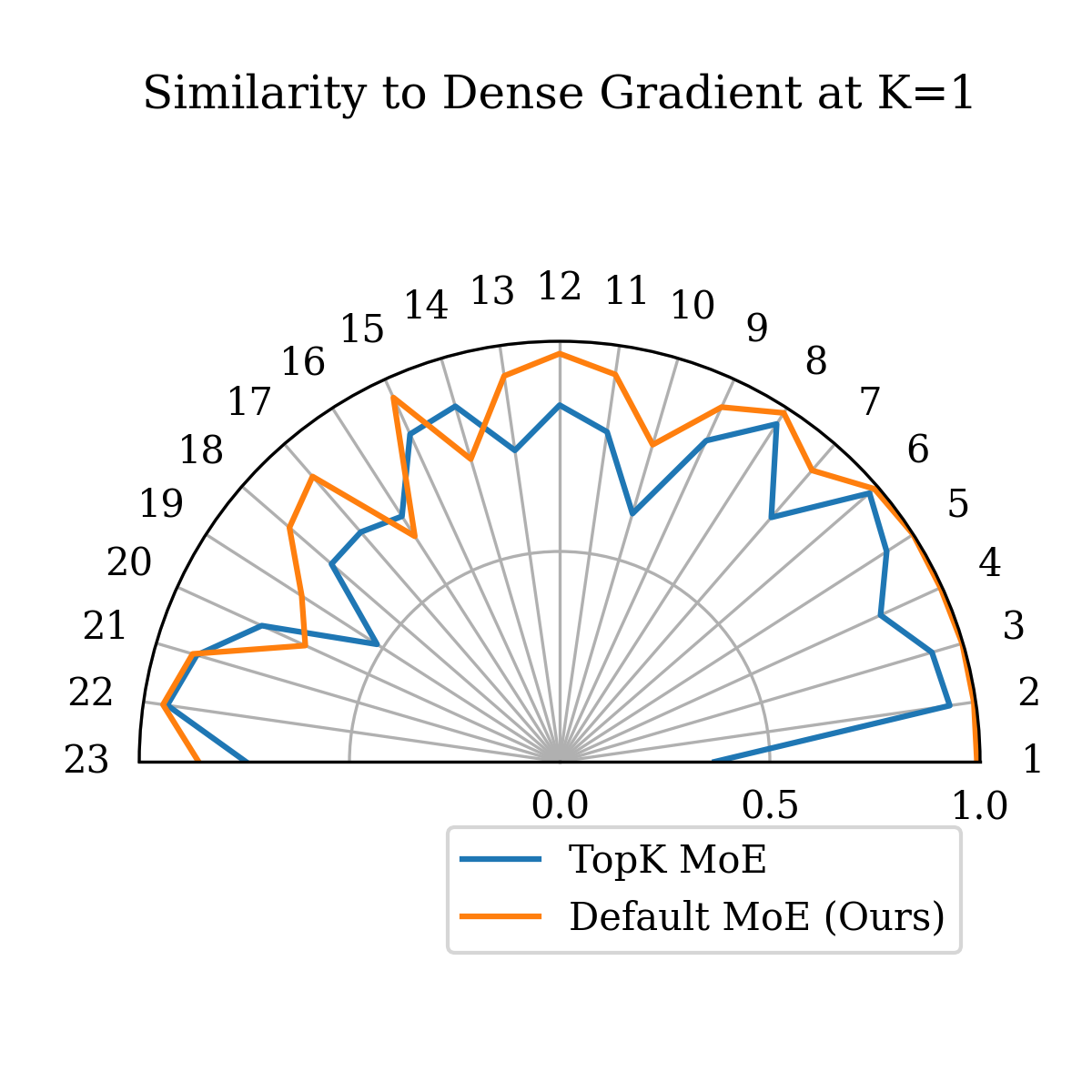}
    \label{fig:comp_plots_ablation}
  \end{subfigure}
  \hfill
  \begin{subfigure}[b]{0.40\textwidth}
    \centering
    \includegraphics[width=\linewidth]{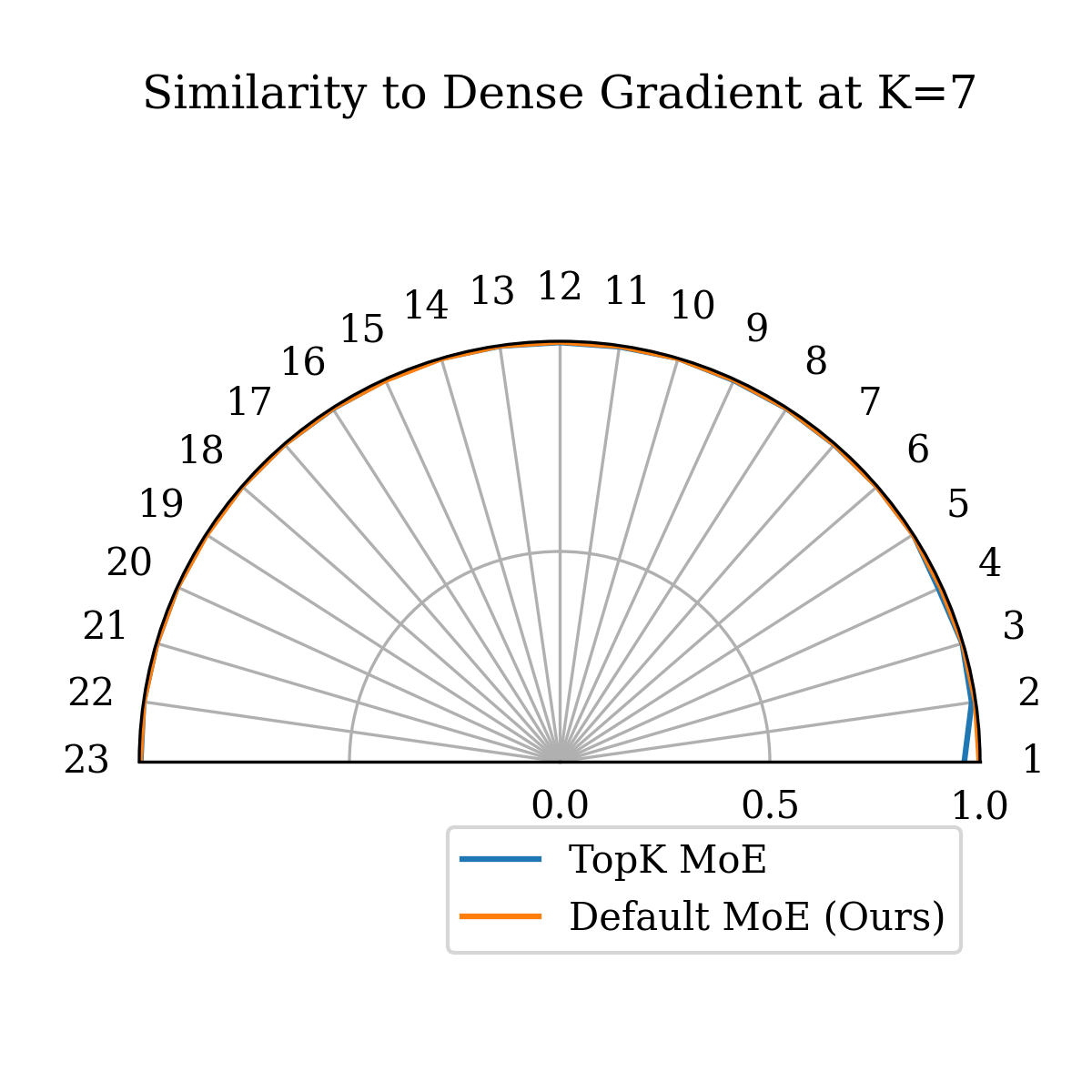}
    \label{fig:lr_sweep_ablation}
  \end{subfigure}
  \caption{\textbf{Dense Router Gradient Similarity.} We plot similarity to the dense router gradient~(K=8) for our 8ck Top-K and DefaultMoEs pretrained on 160B tokens. The DefaultMoE's router gradient is much more similar to the dense router gradient.}
  \label{fig:dense_gradient_similarity}
\end{figure}
\textbf{Similarity to Dense Gradient.}
Our entire design of DefaultMoE is based on the premise that we can better approximate the dense gradient corresponding to simultaneous activation of all the 8 experts. In~\cref{fig:dense_gradient_similarity}, we validate that our DefaultMoE's router gradient is more similar to the dense gradient than the TopK MoE's router gradient is. We take the 8c1 MoE pretrained on 160B tokens and compute the dense router gradient, by activating all the 8 experts. We then vary K between 1 and 7~(the full range can be found in~\cref{sec:appen-analysis}) and report the similarity of the sparse router gradient to the dense gradient, for both DefaultMoE and TopK MoE. At K=1, DefaultMoE is clearly much more similar to the dense gradient for the first few layers, while TopK is not close at all. This is because the routing distribution has high entropy at the early layers, so in DefaultMoE, the early layers will make significant use of the default vectors. This increases the gradient flow to the router which is reflected in the closer approximation of the dense gradient. Whereas even if the router is allocating near-equal mass on all experts and TopK MoE only selects one, its router gradient will be quite dissimilar from the dense gradient. Even at K=7, TopK MoE is still not quite approximating the dense gradient, indicating importance of the missing information from the least activated expert.

\subsection{Efficiency \label{subsec:efficiency}}
\noindent \textbf{Our Method Does Not Significantly Reduce Throughput.}
When we train the 7.33B 8-expert MoE, the per-node GPU throughput is 1,393 tokens per second for TopKMoE and 1,391 tokens per second for DefaultMoE, with a variation of about 1-2 tokens per second on each of these numbers; this is a throughput decrease of just \(0.18\%\) which is just variance. For larger models, more of the time will be spent in matmuls and the overhead of our method will be insignificant.

\noindent \textbf{Our Method Does Not Significantly Increase the Memory Footprint.}
The only additional memory required by our method is the EMA buffers themselves. For each expert in each layer, we store a buffer of the same size as the model's hidden dimension. The total number of parameters in an expert is \(hidden\_size \times intermediate\_size\); for our MoE this is \(1024 \times 2816\). We increase this by \(1024\), which is a negligible \(1/2816=0.03\%\) increase in the number of MoE parameters. 
\begin{table}[h!]
\centering
\caption{\textbf{Throughput Comparison. }We compare the throughput between DefaultMoE and TopKMoE for different model sizes, measured in tokens per second (sequence length 2048) on 1 GPU.}
\resizebox{0.5\linewidth}{!}{
\begin{tabular}{cc|cc|c}
\toprule
Hidden & Model & \multicolumn{2}{c|}{Tokens per second} & Overhead vs TopK \\
Dim & Size & TopK & Ours &  \\
\midrule
1024 & 1.96B & 26,393 & 25,913 & -1.85\% \\
2048 & 7.33B & 1,393 & 1,391 & -0.18\% \\
\bottomrule
\end{tabular}
}
\label{tab:throughput}
\end{table}
\section{Discussion}\label{sec:discussion}
We have validated that DefaultMoEs beat tuned TopKMoEs across multiple settings, but \emph{why?} Our explanation is as follows. Our default vectors are serving as nontrivial~(\cref{fig:default_vector_similarity}) approximations of their respective expert activations. Our gradient error should be corrected, which is validated by DefaultMoE's router gradient being closer to the dense router gradient than that of TopKMoE~(\cref{fig:dense_gradient_similarity}). Less gradient error means that DefaultMoE can tolerate a higher learning rate than TopKMoE~(\cref{fig:all_ablation}), or simply converge faster at TopKMoE's best learning rate~(\cref{fig:tokens-to-target-ppl}) and score higher on benchmarks~(\cref{tab:benchmarks}). This improvement is efficient~(\cref{tab:throughput}), and does not require additional hyperparameter tuning~(\cref{fig:scored_beta_ablation}). We hope that our work will help the community train better MoEs.

\bibliographystyle{plainnat}
\bibliography{neurips}

\begin{thebibliography}{39}
\providecommand{\natexlab}[1]{#1}
\providecommand{\url}[1]{\texttt{#1}}
\expandafter\ifx\csname urlstyle\endcsname\relax
  \providecommand{\doi}[1]{doi: #1}\else
  \providecommand{\doi}{doi: \begingroup \urlstyle{rm}\Url}\fi

\bibitem[Andonian et~al.(2023)Andonian, Anthony, Biderman, Black, Gali, Gao, Hallahan, Levy-Kramer, Leahy, Nestler, Parker, Pieler, Phang, Purohit, Schoelkopf, Stander, Songz, Tigges, Thérien, Wang, and Weinbach]{gpt-neox-library}
Alex Andonian, Quentin Anthony, Stella Biderman, Sid Black, Preetham Gali, Leo Gao, Eric Hallahan, Josh Levy-Kramer, Connor Leahy, Lucas Nestler, Kip Parker, Michael Pieler, Jason Phang, Shivanshu Purohit, Hailey Schoelkopf, Dashiell Stander, Tri Songz, Curt Tigges, Benjamin Thérien, Phil Wang, and Samuel Weinbach.
\newblock {GPT-NeoX: Large Scale Autoregressive Language Modeling in PyTorch}, 9 2023.
\newblock URL \url{https://www.github.com/eleutherai/gpt-neox}.

\bibitem[Ba et~al.(2016)Ba, Kiros, and Hinton]{ba2016layernormalization}
Jimmy~Lei Ba, Jamie~Ryan Kiros, and Geoffrey~E. Hinton.
\newblock Layer normalization, 2016.
\newblock URL \url{https://arxiv.org/abs/1607.06450}.

\bibitem[Bengio et~al.(2013)Bengio, Léonard, and Courville]{bengio2013estimatingpropagatinggradientsstochastic}
Yoshua Bengio, Nicholas Léonard, and Aaron Courville.
\newblock Estimating or propagating gradients through stochastic neurons for conditional computation, 2013.
\newblock URL \url{https://arxiv.org/abs/1308.3432}.

\bibitem[Clark et~al.(2022)Clark, de~las Casas, Guy, Mensch, Paganini, Hoffmann, Damoc, Hechtman, Cai, Borgeaud, van~den Driessche, Rutherford, Hennigan, Johnson, Millican, Cassirer, Jones, Buchatskaya, Budden, Sifre, Osindero, Vinyals, Rae, Elsen, Kavukcuoglu, and Simonyan]{clark2022unifiedscalinglawsrouted}
Aidan Clark, Diego de~las Casas, Aurelia Guy, Arthur Mensch, Michela Paganini, Jordan Hoffmann, Bogdan Damoc, Blake Hechtman, Trevor Cai, Sebastian Borgeaud, George van~den Driessche, Eliza Rutherford, Tom Hennigan, Matthew Johnson, Katie Millican, Albin Cassirer, Chris Jones, Elena Buchatskaya, David Budden, Laurent Sifre, Simon Osindero, Oriol Vinyals, Jack Rae, Erich Elsen, Koray Kavukcuoglu, and Karen Simonyan.
\newblock Unified scaling laws for routed language models, 2022.
\newblock URL \url{https://arxiv.org/abs/2202.01169}.

\bibitem[Databricks(2024)]{DBRX}
Databricks.
\newblock Dbrx, 2024.
\newblock URL \url{https://www.databricks.com/blog/introducing-dbrx-new-state-art-open-llm}.

\bibitem[{DeepSeek-AI Team}(2024{\natexlab{a}})]{deepseekai2024deepseekv2strongeconomicalefficient}
{DeepSeek-AI Team}.
\newblock Deepseek-v2: A strong, economical, and efficient mixture-of-experts language model, 2024{\natexlab{a}}.
\newblock URL \url{https://arxiv.org/abs/2405.04434}.

\bibitem[{DeepSeek-AI Team}(2024{\natexlab{b}})]{deepseekv3}
{DeepSeek-AI Team}.
\newblock Deepseek-v3 technical report, 2024{\natexlab{b}}.
\newblock URL \url{https://arxiv.org/abs/2412.19437}.

\bibitem[Du et~al.(2022)Du, Huang, Dai, Tong, Lepikhin, Xu, Krikun, Zhou, Yu, Firat, et~al.]{du2022glam}
Nan Du, Yanping Huang, Andrew~M Dai, Simon Tong, Dmitry Lepikhin, Yuanzhong Xu, Maxim Krikun, Yanqi Zhou, Adams~Wei Yu, Orhan Firat, et~al.
\newblock Glam: Efficient scaling of language models with mixture-of-experts.
\newblock In \emph{International Conference on Machine Learning}, pages 5547--5569. PMLR, 2022.

\bibitem[Fedus et~al.(2022)Fedus, Zoph, and Shazeer]{fedus2022switchtransformersscalingtrillion}
William Fedus, Barret Zoph, and Noam Shazeer.
\newblock Switch transformers: Scaling to trillion parameter models with simple and efficient sparsity, 2022.
\newblock URL \url{https://arxiv.org/abs/2101.03961}.

\bibitem[Gale et~al.(2022)Gale, Narayanan, Young, and Zaharia]{gale2022megablocksefficientsparsetraining}
Trevor Gale, Deepak Narayanan, Cliff Young, and Matei Zaharia.
\newblock Megablocks: Efficient sparse training with mixture-of-experts, 2022.
\newblock URL \url{https://arxiv.org/abs/2211.15841}.

\bibitem[Gao et~al.(2024)Gao, Tow, Abbasi, Biderman, Black, DiPofi, Foster, Golding, Hsu, Le~Noac'h, Li, McDonell, Muennighoff, Ociepa, Phang, Reynolds, Schoelkopf, Skowron, Sutawika, Tang, Thite, Wang, Wang, and Zou]{eval-harness}
Leo Gao, Jonathan Tow, Baber Abbasi, Stella Biderman, Sid Black, Anthony DiPofi, Charles Foster, Laurence Golding, Jeffrey Hsu, Alain Le~Noac'h, Haonan Li, Kyle McDonell, Niklas Muennighoff, Chris Ociepa, Jason Phang, Laria Reynolds, Hailey Schoelkopf, Aviya Skowron, Lintang Sutawika, Eric Tang, Anish Thite, Ben Wang, Kevin Wang, and Andy Zou.
\newblock A framework for few-shot language model evaluation, 07 2024.
\newblock URL \url{https://zenodo.org/records/12608602}.

\bibitem[{Gemini Team}(2024)]{geminiteam2024gemini15unlockingmultimodal}
{Gemini Team}.
\newblock Gemini 1.5: Unlocking multimodal understanding across millions of tokens of context, 2024.
\newblock URL \url{https://arxiv.org/abs/2403.05530}.

\bibitem[Hoffmann et~al.(2022)Hoffmann, Borgeaud, Mensch, Buchatskaya, Cai, Rutherford, de~Las~Casas, Hendricks, Welbl, Clark, Hennigan, Noland, Millican, van~den Driessche, Damoc, Guy, Osindero, Simonyan, Elsen, Rae, Vinyals, and Sifre]{hoffmann2022trainingcomputeoptimallargelanguage}
Jordan Hoffmann, Sebastian Borgeaud, Arthur Mensch, Elena Buchatskaya, Trevor Cai, Eliza Rutherford, Diego de~Las~Casas, Lisa~Anne Hendricks, Johannes Welbl, Aidan Clark, Tom Hennigan, Eric Noland, Katie Millican, George van~den Driessche, Bogdan Damoc, Aurelia Guy, Simon Osindero, Karen Simonyan, Erich Elsen, Jack~W. Rae, Oriol Vinyals, and Laurent Sifre.
\newblock Training compute-optimal large language models, 2022.
\newblock URL \url{https://arxiv.org/abs/2203.15556}.

\bibitem[Hsu et~al.(2024)Hsu, Dai, Kothapalli, Song, Tang, Zhu, Shimizu, Sahni, Ning, and Chen]{hsu2024ligerkernelefficienttriton}
Pin-Lun Hsu, Yun Dai, Vignesh Kothapalli, Qingquan Song, Shao Tang, Siyu Zhu, Steven Shimizu, Shivam Sahni, Haowen Ning, and Yanning Chen.
\newblock Liger kernel: Efficient triton kernels for llm training.
\newblock \emph{arXiv preprint arXiv:2410.10989}, 2024.
\newblock URL \url{https://arxiv.org/abs/2410.10989}.

\bibitem[{Hunyuan Team}(2024)]{sun2024hunyuanlargeopensourcemoemodel}
{Hunyuan Team}.
\newblock Hunyuan-large: An open-source moe model with 52 billion activated parameters by tencent, 2024.
\newblock URL \url{https://arxiv.org/abs/2411.02265}.

\bibitem[Jacobs et~al.(1991)Jacobs, Jordan, Nowlan, and Hinton]{jacobs1991}
Robert~A. Jacobs, Michael~I. Jordan, Steven~J. Nowlan, and Geoffrey~E. Hinton.
\newblock {Adaptive Mixtures of Local Experts}.
\newblock \emph{Neural Computation}, 3\penalty0 (1):\penalty0 79--87, 03 1991.
\newblock ISSN 0899-7667.
\newblock \doi{10.1162/neco.1991.3.1.79}.
\newblock URL \url{https://doi.org/10.1162/neco.1991.3.1.79}.

\bibitem[Jordan and Jacobs(1994)]{jordanjacobs1994}
M.~I. Jordan and R.~A. Jacobs.
\newblock Hierarchical mixtures of experts and the em algorithm.
\newblock In Maria Marinaro and Pietro~G. Morasso, editors, \emph{ICANN '94}, pages 479--486, London, 1994. Springer London.
\newblock ISBN 978-1-4471-2097-1.

\bibitem[Lepikhin et~al.(2020)Lepikhin, Lee, Xu, Chen, Firat, Huang, Krikun, Shazeer, and Chen]{lepikhin2020gshard}
Dmitry Lepikhin, HyoukJoong Lee, Yuanzhong Xu, Dehao Chen, Orhan Firat, Yanping Huang, Maxim Krikun, Noam Shazeer, and Zhifeng Chen.
\newblock Gshard: Scaling giant models with conditional computation and automatic sharding.
\newblock \emph{arXiv preprint arXiv:2006.16668}, 2020.

\bibitem[Liu et~al.(2023)Liu, Gao, and Chen]{liu2023sparsebackpropagationmoetraining}
Liyuan Liu, Jianfeng Gao, and Weizhu Chen.
\newblock Sparse backpropagation for moe training, 2023.
\newblock URL \url{https://arxiv.org/abs/2310.00811}.

\bibitem[Liu et~al.(2024)Liu, Kim, Wang, Liang, Shen, Cheng, Liu, Tanaka, Wu, Hu, Chaudhary, Lin, Zhang, Xue, Awadalla, Gao, and Chen]{liu2024gringradientinformedmoe}
Liyuan Liu, Young~Jin Kim, Shuohang Wang, Chen Liang, Yelong Shen, Hao Cheng, Xiaodong Liu, Masahiro Tanaka, Xiaoxia Wu, Wenxiang Hu, Vishrav Chaudhary, Zeqi Lin, Chenruidong Zhang, Jilong Xue, Hany Awadalla, Jianfeng Gao, and Weizhu Chen.
\newblock Grin: Gradient-informed moe, 2024.
\newblock URL \url{https://arxiv.org/abs/2409.12136}.

\bibitem[{Llama 3 Team}(2024)]{dubey2024llama3herdmodels}
{Llama 3 Team}.
\newblock The llama 3 herd of models, 2024.
\newblock URL \url{https://arxiv.org/abs/2407.21783}.

\bibitem[Loshchilov and Hutter(2019)]{loshchilov2019decoupledweightdecayregularization}
Ilya Loshchilov and Frank Hutter.
\newblock Decoupled weight decay regularization, 2019.
\newblock URL \url{https://arxiv.org/abs/1711.05101}.

\bibitem[Lozhkov et~al.(2024)Lozhkov, Ben~Allal, von Werra, and Wolf]{lozhkov2024fineweb-edu}
Anton Lozhkov, Loubna Ben~Allal, Leandro von Werra, and Thomas Wolf.
\newblock Fineweb-edu: the finest collection of educational content, 2024.
\newblock URL \url{https://huggingface.co/datasets/HuggingFaceFW/fineweb-edu}.

\bibitem[{Mistral Team}(2024)]{jiang2024mixtralexperts}
{Mistral Team}.
\newblock Mixtral of experts, 2024.
\newblock URL \url{https://arxiv.org/abs/2401.04088}.

\bibitem[Nguyen and Salazar(2019)]{https://doi.org/10.5281/zenodo.3525484}
Toan~Q. Nguyen and Julian Salazar.
\newblock Transformers without tears: Improving the normalization of self-attention.
\newblock 2019.
\newblock \doi{10.5281/ZENODO.3525484}.
\newblock URL \url{https://zenodo.org/record/3525484}.

\bibitem[{OpenAI Team}(2024)]{openai2024gpt4technicalreport}
{OpenAI Team}.
\newblock Gpt-4 technical report, 2024.
\newblock URL \url{https://arxiv.org/abs/2303.08774}.

\bibitem[Paperno et~al.(2016)Paperno, Kruszewski, Lazaridou, Pham, Bernardi, Pezzelle, Baroni, Boleda, and Fernández]{paperno2016lambadadatasetwordprediction}
Denis Paperno, Germán Kruszewski, Angeliki Lazaridou, Quan~Ngoc Pham, Raffaella Bernardi, Sandro Pezzelle, Marco Baroni, Gemma Boleda, and Raquel Fernández.
\newblock The lambada dataset: Word prediction requiring a broad discourse context, 2016.
\newblock URL \url{https://arxiv.org/abs/1606.06031}.

\bibitem[Penedo et~al.(2024)Penedo, Kydlíček, allal, Lozhkov, Mitchell, Raffel, Werra, and Wolf]{penedo2024finewebdatasetsdecantingweb}
Guilherme Penedo, Hynek Kydlíček, Loubna~Ben allal, Anton Lozhkov, Margaret Mitchell, Colin Raffel, Leandro~Von Werra, and Thomas Wolf.
\newblock The fineweb datasets: Decanting the web for the finest text data at scale, 2024.
\newblock URL \url{https://arxiv.org/abs/2406.17557}.

\bibitem[{Phi Team}(2024)]{abdin2024phi3technicalreporthighly}
{Phi Team}.
\newblock Phi-3 technical report: A highly capable language model locally on your phone, 2024.
\newblock URL \url{https://arxiv.org/abs/2404.14219}.

\bibitem[Qiu et~al.(2025)Qiu, Huang, Zheng, Wen, Wang, Men, Titov, Liu, Zhou, and Lin]{qiu2025demonsdetailimplementingload}
Zihan Qiu, Zeyu Huang, Bo~Zheng, Kaiyue Wen, Zekun Wang, Rui Men, Ivan Titov, Dayiheng Liu, Jingren Zhou, and Junyang Lin.
\newblock Demons in the detail: On implementing load balancing loss for training specialized mixture-of-expert models.
\newblock \emph{arXiv preprint arXiv:2501.11873}, 2025.

\bibitem[Shazeer(2020)]{shazeer2020gluvariantsimprovetransformer}
Noam Shazeer.
\newblock Glu variants improve transformer, 2020.
\newblock URL \url{https://arxiv.org/abs/2002.05202}.

\bibitem[Shazeer et~al.(2017)Shazeer, Mirhoseini, Maziarz, Davis, Le, Hinton, and Dean]{shazeer2017outrageouslylargeneuralnetworks}
Noam Shazeer, Azalia Mirhoseini, Krzysztof Maziarz, Andy Davis, Quoc Le, Geoffrey Hinton, and Jeff Dean.
\newblock Outrageously large neural networks: The sparsely-gated mixture-of-experts layer, 2017.
\newblock URL \url{https://arxiv.org/abs/1701.06538}.

\bibitem[Snowflake(2024)]{arctic}
Snowflake.
\newblock Arctic, 2024.
\newblock URL \url{https://www.snowflake.com/en/blog/arctic-open-efficient-foundation-language-models-snowflake/}.

\bibitem[Su et~al.(2023)Su, Lu, Pan, Murtadha, Wen, and Liu]{su2023roformerenhancedtransformerrotary}
Jianlin Su, Yu~Lu, Shengfeng Pan, Ahmed Murtadha, Bo~Wen, and Yunfeng Liu.
\newblock Roformer: Enhanced transformer with rotary position embedding, 2023.
\newblock URL \url{https://arxiv.org/abs/2104.09864}.

\bibitem[Touvron et~al.(2023)Touvron, Lavril, Izacard, Martinet, Lachaux, Lacroix, Rozière, Goyal, Hambro, Azhar, Rodriguez, Joulin, Grave, and Lample]{touvron2023llamaopenefficientfoundation}
Hugo Touvron, Thibaut Lavril, Gautier Izacard, Xavier Martinet, Marie-Anne Lachaux, Timothée Lacroix, Baptiste Rozière, Naman Goyal, Eric Hambro, Faisal Azhar, Aurelien Rodriguez, Armand Joulin, Edouard Grave, and Guillaume Lample.
\newblock Llama: Open and efficient foundation language models, 2023.
\newblock URL \url{https://arxiv.org/abs/2302.13971}.

\bibitem[Wang et~al.(2022)Wang, Ma, Dong, Huang, Zhang, and Wei]{wang2022deepnetscalingtransformers1000}
Hongyu Wang, Shuming Ma, Li~Dong, Shaohan Huang, Dongdong Zhang, and Furu Wei.
\newblock Deepnet: Scaling transformers to 1,000 layers, 2022.
\newblock URL \url{https://arxiv.org/abs/2203.00555}.

\bibitem[Wang et~al.(2024)Wang, Gao, Zhao, Sun, and Dai]{wang2024auxiliarylossfreeloadbalancingstrategy}
Lean Wang, Huazuo Gao, Chenggang Zhao, Xu~Sun, and Damai Dai.
\newblock Auxiliary-loss-free load balancing strategy for mixture-of-experts, 2024.
\newblock URL \url{https://arxiv.org/abs/2408.15664}.

\bibitem[Wang et~al.(2025)Wang, Zhu, and Chen]{wang2025remoefullydifferentiablemixtureofexperts}
Ziteng Wang, Jun Zhu, and Jianfei Chen.
\newblock Remoe: Fully differentiable mixture-of-experts with relu routing, 2025.
\newblock URL \url{https://arxiv.org/abs/2412.14711}.

\bibitem[xAI(2024)]{grok}
xAI.
\newblock Grok-1, 2024.
\newblock URL \url{https://github.com/xai-org/grok-1?tab=readme-ov-file}.

\end{thebibliography}

\newpage
\appendix

\appendix
\section{Appendix}
\subsection{Limitations}
Throughout our work we only train dropless MoEs. These are ideal for the academic research setting, because they train efficiently in data-parallel and so do not require cumbersome implementations of model or pipeline parallelism. However, dropless MoEs may not be well-suited for industry, because reshaping the tensors to provide consistent shapes to the MoE kernels incurs many host-device syncs. This is a limitation that we share with prior work, but an important future direction will be extending our results to token-dropping MoEs. 

We report results on a range of standard language modeling benchmarks, but we have not invested enough compute to see performance on LLM benchmarks such as MMLU-Pro, and our data is not sufficiently domain-specific to report scores on coding benchmarks (HumanEval, MBPP) or math (GSM8k). In the same vein, our models have not been trained to reason. We focus heavily on pretraining in this paper, but an important future direction would be seeing whether we can finetune MoEs with our method.
\subsection{Experimental Setup Details}\label{sec:appen_setup}
\textbf{Architecture. }
We use SwiGLU~\citep{shazeer2020gluvariantsimprovetransformer} MLPs following Llama~\citep{touvron2023llamaopenefficientfoundation}, \(16\) attention heads with dimension of \(64\), LayerNorm~\citep{ba2016layernormalization} and RoPE~\citep{su2023roformerenhancedtransformerrotary}.

\textbf{Hyperparameters. } 
We use the initialization from ~\citet{wang2022deepnetscalingtransformers1000} for residual branch merge layers and the initialization from~\citet{https://doi.org/10.5281/zenodo.3525484} for all other layers.
We use the AdamW optimizer~\citep{loshchilov2019decoupledweightdecayregularization}. We use a sequence length of \(2048\) and a global batch size of \(1024\), resulting in a global token batch size of \(2^{21}\). We use a standard cosine decay schedule.

\textbf{MoE-Specific Training Details. }We set the auxiliary loss~\citep{fedus2022switchtransformersscalingtrillion} to \(0.01\). We do not use z-loss or jitter, because we find that at this scale they do not improve training for the baseline model. Following~\citet{deepseekv3}, we set the first layer to be dense. Following~\citet{liu2024gringradientinformedmoe}, we compute the aux loss across nodes. We train dropless MoEs.

\textbf{Implementation.} We train with the gpt-neox library~\citep{gpt-neox-library} integrated with Megablocks~\citep{gale2022megablocksefficientsparsetraining} and augmented with Triton kernels from~\citep{hsu2024ligerkernelefficienttriton}. 

\subsection{More details on other Routing Mechanism Comparisons}

\begin{wrapfigure}{r}{0.475\textwidth}
\vspace{-15pt}
    \includegraphics[width=0.475\textwidth]{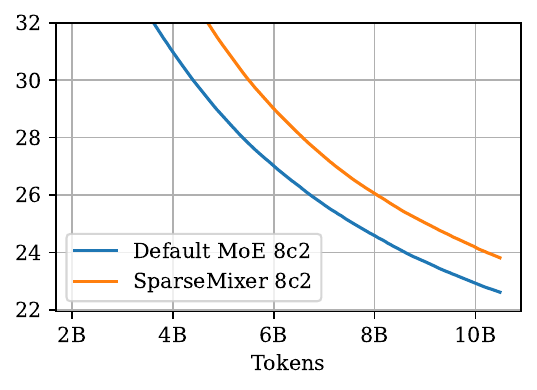}
    \caption{\textbf{Comparison of Default MoE and SparseMixer.} We compare Default MoE with SparseMixer, both configured with 8 experts and Top-K=2 active experts. The results report training perplexity throughout training, demonstrating that Default MoE consistently outperforms SparseMixer.}
    \label{fig:def_vector_vs_sparsemixer_ablation}
    \vspace{-25pt}
\end{wrapfigure}

We see in~\cref{fig:def_vector_vs_sparsemixer_ablation} that our method significantly outperforms SparseMixer for at least the first \(10B\) tokens.

\subsection{Ablating EMA design choices}\label{sec:appen_beta}

\textbf{Tuning $\beta$. }The lone hyperparameter introduced by our method is the \(\beta\) parameter of the EMA. 
While we use $\beta=0.9$ for the 8c1 and 8c2 MoEs in \cref{fig:comp_plots_ablation}, $\beta$ requires more careful tuning for the 32 expert MoEs. For $N=32$, we use $\beta=0.65$, $\beta=0.95$, and $\beta=0.999$ for $K=1$, $K=2$, and $K=4$ respectively. In other words, sparser MoEs require a lower $\beta$. We believe this is due to each expert receiving less tokens at each step, which leads to a sparser "history" for each default vector. As a result, the default vector's estimate for the average expert output improves when assigning higher weight to the current batch.

\begin{figure}[hbtp]
    \centering
    \includegraphics[width=0.8\linewidth]{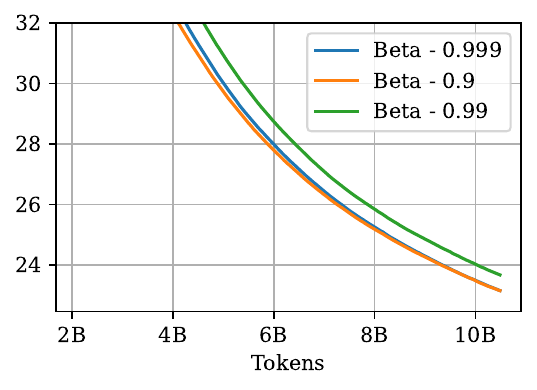}
    \caption{\textbf{Selecting the Best EMA Beta Parameter.} We compare different values of the exponential moving average (EMA) smoothing factor \(\beta\) for an 8c1 Default MoE, evaluating \(\beta = 0.999\), \(\beta = 0.99\), and \(\beta = 0.9\). The results show that \(\beta = 0.9\) provides the most stable and effective training behavior, leading to lower perplexity. Based on these findings, we use \(\beta = 0.9\) to train our 8c1 model for 320B tokens. }
    
    \label{fig:ema_beta_ablation}
\end{figure}

\begin{figure}[btp]
    \centering
    \begin{subfigure}[b]{0.49\linewidth}
        \includegraphics[width=\textwidth]{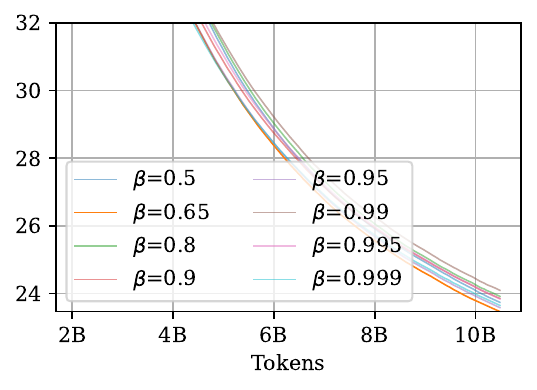}
    \caption{$K=1$}
    \end{subfigure}
    \begin{subfigure}[b]{0.49\linewidth}
        \includegraphics[width=\textwidth]{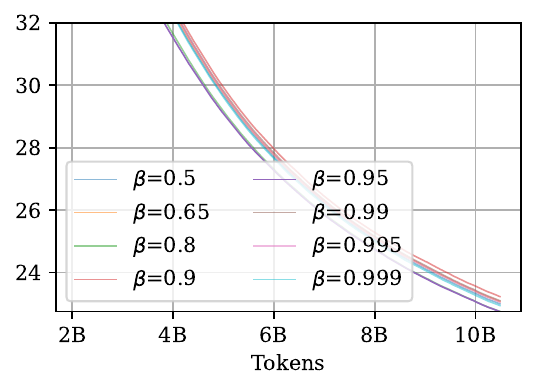}
    \caption{$K=2$}
    \end{subfigure}
    \begin{subfigure}[b]{0.49\linewidth}
        \includegraphics[width=\textwidth]{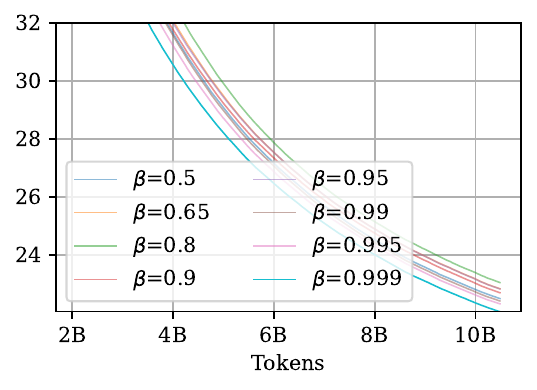}
    \caption{$K=4$}
    \end{subfigure}
    \caption{\textbf{Sweeps for the $\beta$ hyperparameter for Default MoE with 32 experts.} For $K=1$ the optimal $\beta$ is $0.65$; for $K=2$ it is $0.95$, for $K=4$ it is $0.999$. Notably, as $K$ decreases, $\beta$ must also decrease. With roughly 10 billion tokens of training, we can clearly distinguish the best $\beta$ for any given configuration.}
    \label{fig:beta_sweep}
\end{figure}

We vary the \(\beta\) used in the Default MoE in~\cref{fig:ema_beta_ablation} for an 8 expert MoE and find that both \(\beta=0.9\) and \(\beta=0.999\) perform equally well. Intuitively, \(\beta\) parametrizes how fast our EMA adjusts its model of the sample mean as we train the model. In principle the optimal value of \(\beta\) might depend on the learning rate, which defines how fast the model is changing, the batch size, which controls how many tokens are actually incorporated into the EMA, and even the data distribution. For example, we need to tune $\beta$ more carefully for a 32 expert MoE. In~\cref{fig:beta_sweep} we sweep across multiple values of $\beta$ for a 32c1, 32c2, and 32c4 Default MoE.

\begin{figure}
    \centering
    \includegraphics[width=0.8\linewidth]{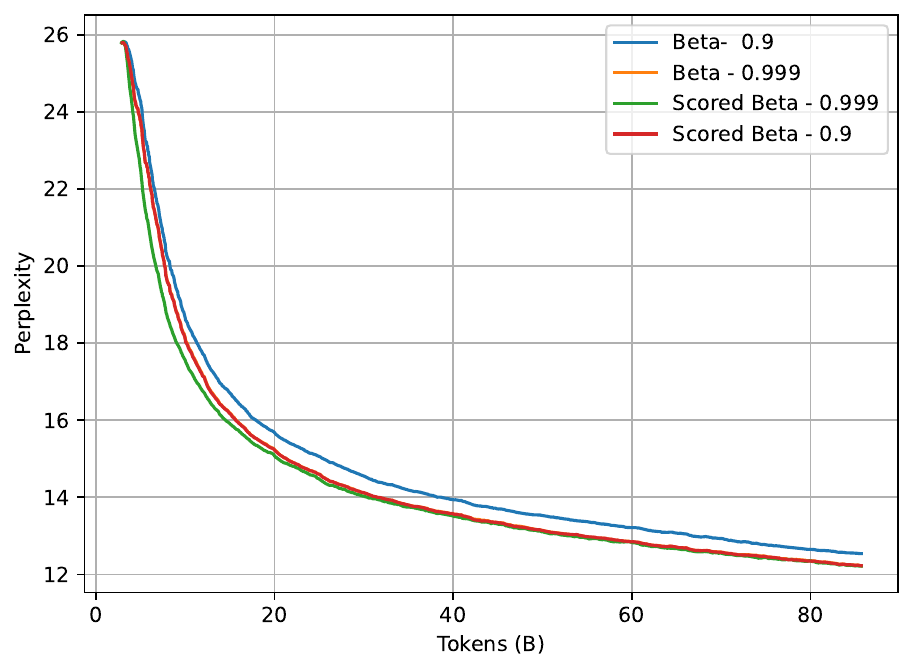}
    \caption{We compare exponential moving average (EMA) updates to the default vector with and without scoring. In the scored version, each expert’s contribution is weighted by its router probability before being added to the EMA update. Without scoring, the choice of $\beta$ (e.g., 0.9 vs. 0.999) significantly affects performance. With scoring, this sensitivity disappears—both scored variants converge to the same strong final performance. The unscored $\beta = 0.999$ curve (orange) is nearly hidden behind the scored curves.}

    \label{fig:scored_beta_ablation}
\end{figure}
\textbf{Weighted Updates Remove Sensitivity to EMA Decay Rate ($\beta$)}.
We explore different values of the EMA decay coefficient $\beta$ used to update the default vector. Naively, performance is sensitive to $\beta$. However, we find that if we accumulate activations in the the default vector weighted by their router score performance becomes robust to the choice of $\beta$. In fact, multiple $\beta$ values converge to the a similar final performance as shown in ~\cref{fig:scored_beta_ablation}.

\begin{figure}[h]
    \centering
    \includegraphics[width=0.8\linewidth]{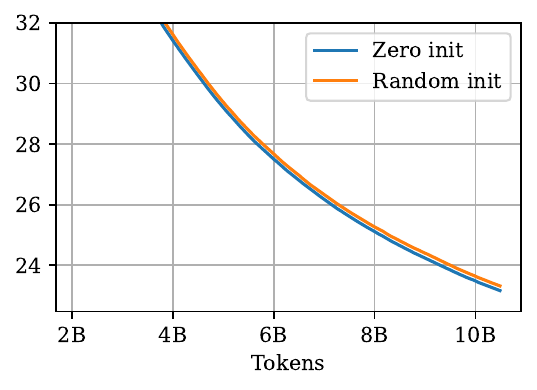}
    \caption{\textbf{EMA Buffer Initialization Strategies for Default MoE.} We compare two approaches for initializing the exponential moving average (EMA) buffer for Default MoE: zero initialization and random initialization from a Gaussian distribution. The results indicate that zero initialization is more effective, leading to lower  perplexity. Based on this finding, we adopt zero initialization for our experiments.}

    \label{fig:ema_random_zero_init}
\end{figure}
\textbf{Initializing the EMA. }
We experiment with initializing default vector EMA in two ways: zero initialization and random (Gaussian) initialization. Zero initialization leads to the default vector starting off with zero signal in the earliest training steps. Random initialization, on the other hand, immediately provides some signal to fill in missing expert outputs but this signal starts as pure noise. Given these tradeoffs, we try both approaches and outline our results in \cref{fig:ema_random_zero_init}. Initializing our EMA with zeroes demonstrates slight improvement compared to random initialization. We believe this is due to the adverse effects from providing a random signal to the router early in training. However, the impact is minor, even without the use of any explicit bias correction term in our EMA.

\begin{figure}
    \centering
    \includegraphics[width=0.8\linewidth]{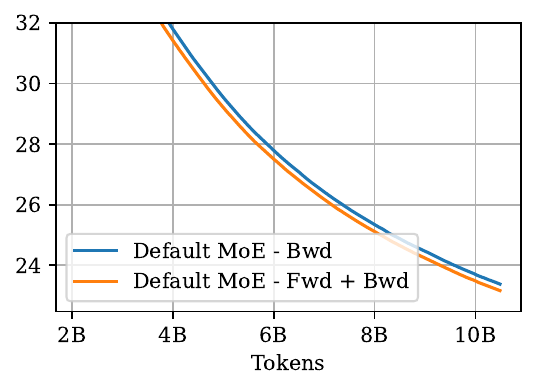}
    \caption{\textbf{The importance of applying the EMA during the forward pass.} We perform an ablation study to examine the effect of applying the exponential moving average (EMA) during both the forward and backward passes versus only the backward pass. The results show that using the EMA in both passes leads to consistently lower training perplexity, whereas applying the EMA only in the backward pass—which affects only the gradient update—results in inferior performance. This highlights the importance of incorporating EMA throughout training to maximize its stabilizing effect.}

    \label{fig:ema_fwd_bwd_ablation}
\end{figure}
\textbf{Passing the Default Vector Forward. }
Our goal is to provide a default expert activation for unactivated experts so that the router can receive a gradient for all experts. We do this by passing the default vector forward through the network, and the router gradient is automatically computed. However, we can in principle do this without passing the default vector forward, and just manually writing the gradient update for the backward pass. One reason why we pass the default vector forward is because our error in estimating the true dense gradient, as written in Eq. 9, is scaled by the loss. Therefore if the default vector activations actually improve the model's output, our error is smaller. We validate this in~\cref{fig:ema_fwd_bwd_ablation}, where we find that passing the default vector forward does improve over only updating the router gradients in the backward pass with the default vector.


\begin{figure}[htbp]
    \centering
    \begin{minipage}{0.475\textwidth}
        \centering
        \includegraphics[width=\linewidth]{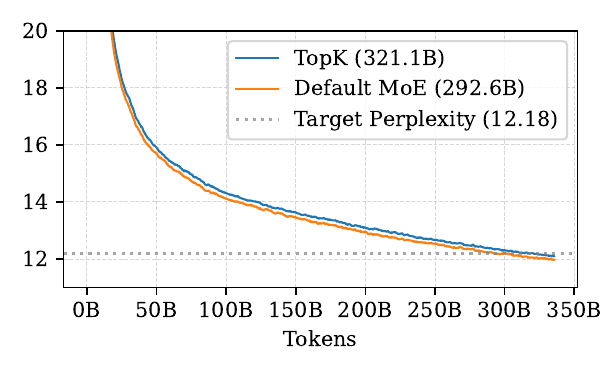}
        \caption{\textbf{Default MoE Beats TopK.} Our Default MoE reaches a perplexity of \(\approx 12\) about \(9\%\) faster than the baseline TopK MoE, without introducing any additional overhead. Both MoEs are configured with 8 experts and Top-K=1 active experts.}
        \label{fig:tokens-to-target-ppl}
    \end{minipage}\hfill
    \begin{minipage}{0.475\textwidth}
    \centering
    \includegraphics[width=\linewidth]{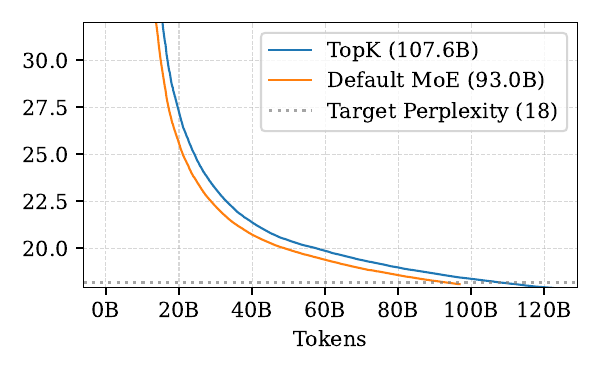}
    \caption{\textbf{Default MoE beats a tuned TopK baseline on FineWeb.} Our method reaches the target perplexity of 18 about 15\% faster than the baseline TopK method, without introducing any additional overhead.}
    \label{fig:lm-loss-fineweb}
    \end{minipage}
\end{figure}

\begin{figure}[htbp]
    \centering

        \centering        \includegraphics[width=0.8\linewidth]{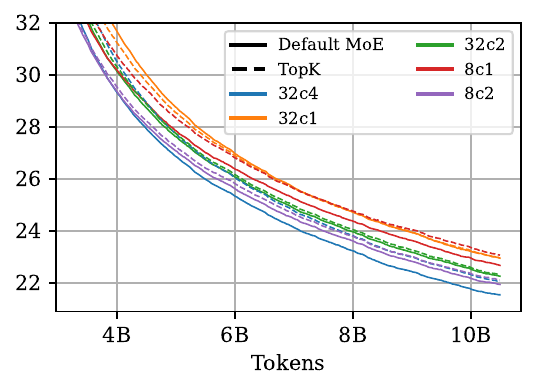}
        \caption{\textbf{Comparison of MoE Configurations.} We compare Default MoE and Top-K across five configurations: 8c1, 8c2, 32c1, 32c2, and 32c4. The results show that Default MoE outperforms Top-K in all MoE configurations. Both 8c1 and 8c2 use $\beta=0.9$; however, 32c1, 32c2, and 32c4 require more careful tuning of $\beta$. We use $\beta=0.65$, $\beta=0.95$, and $\beta=0.999$ for 32c1, 32c2, and 32c4, respectively. We detail our choices of $\beta$ in~\cref{sec:appen_beta}.}
        \label{fig:comp_plots_ablation_curve}

        \centering
\end{figure}

\section{Detailed Empirical Analysis}\label{sec:appen-analysis}

\subsection{Domain Specialization}

\begin{figure}[hbtp]
    \centering
    \includegraphics[width=0.5\linewidth]{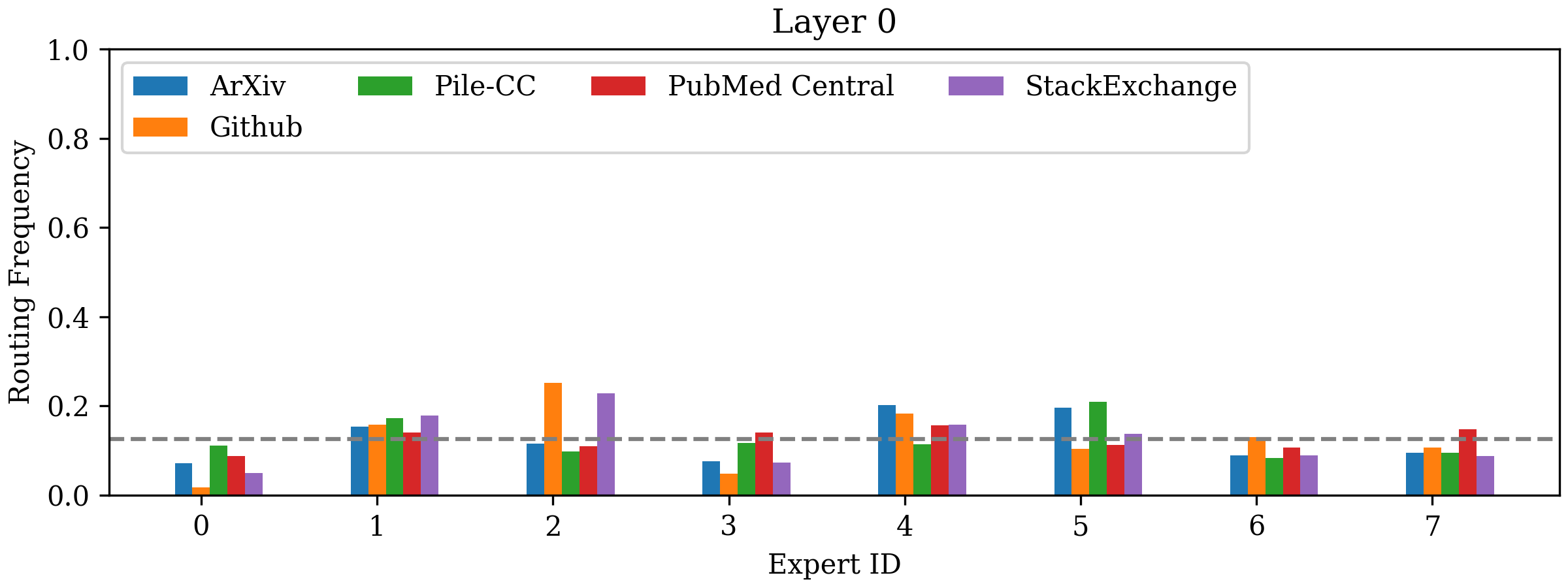}
    \caption{Domain Specialization for 8c1 TopKMoE at Layer 0.}
    \label{fig:appen-spec-topk-8c1-layer0}
\end{figure}

\begin{figure}[hbtp]
    \centering
    \includegraphics[width=0.5\linewidth]{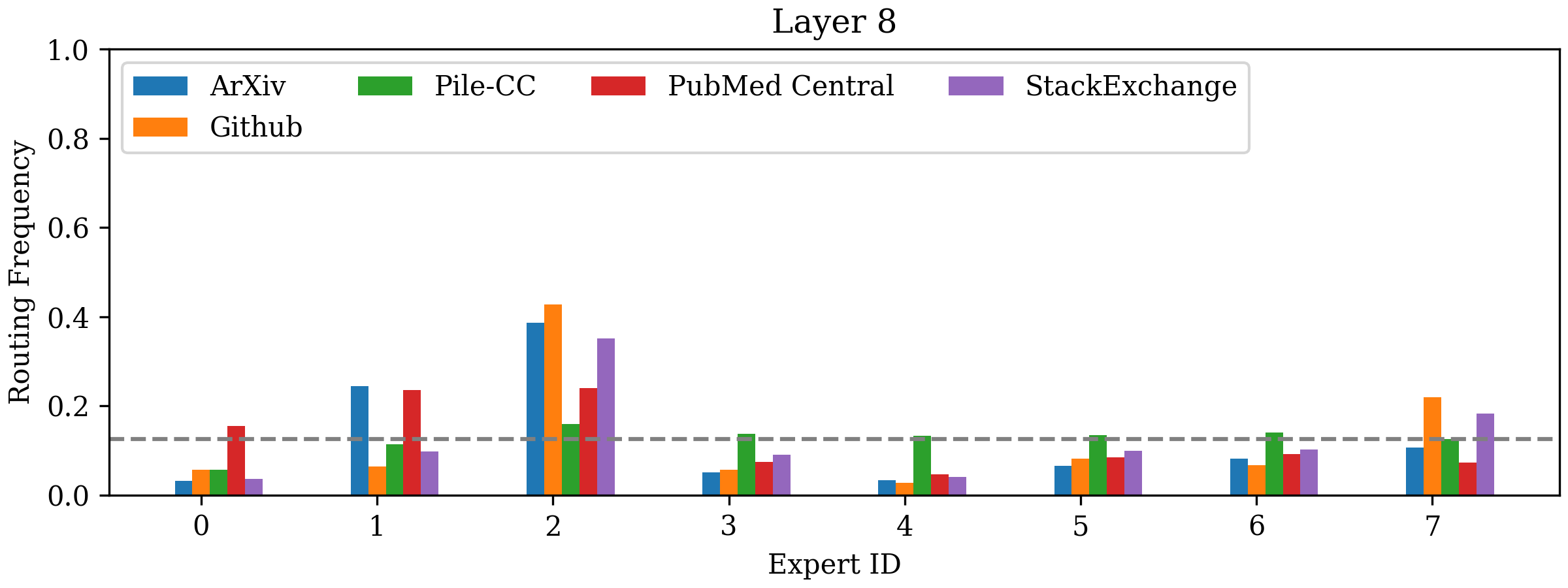}
    \caption{Domain Specialization for 8c1 TopKMoE at Layer 8.}
    \label{fig:appen-spec-topk-8c1-layer8}
\end{figure}

\begin{figure}[hbtp]
    \centering
    \includegraphics[width=0.5\linewidth]{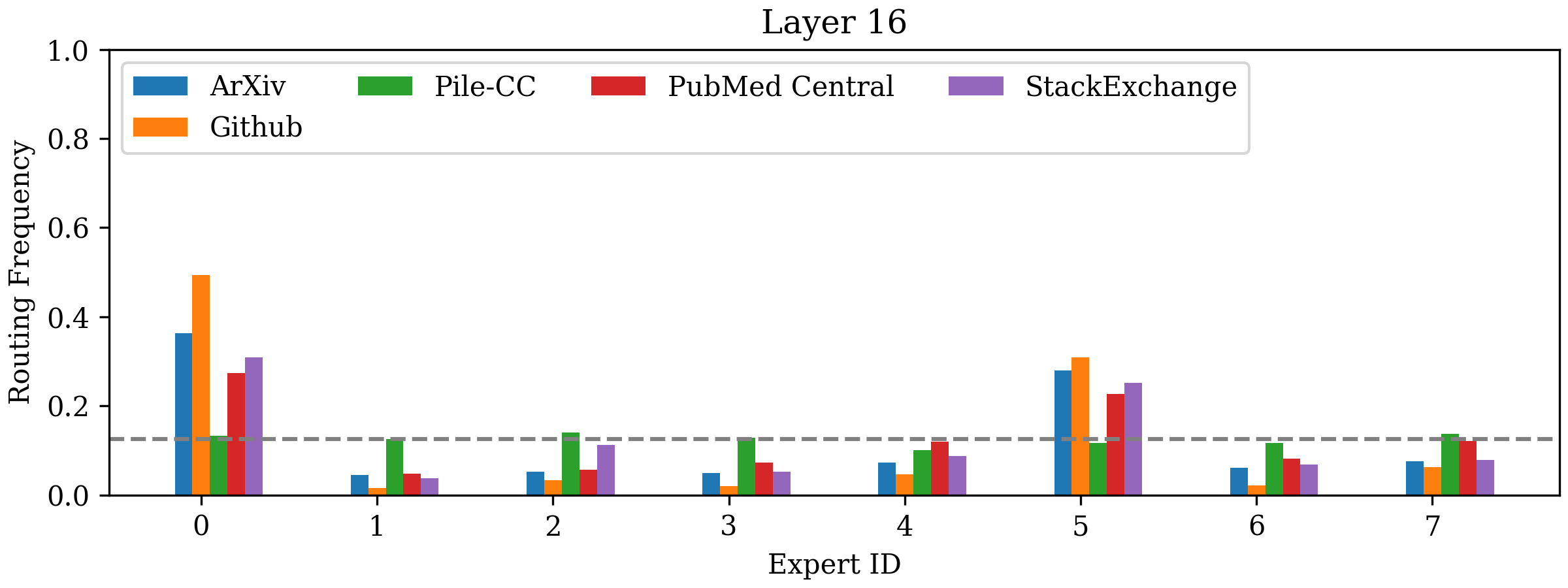}
    \caption{Domain Specialization for 8c1 TopKMoE at Layer 16.}
    \label{fig:appen-spec-topk-8c1-layer16}
\end{figure}

\begin{figure}[hbtp]
    \centering
    \includegraphics[width=0.5\linewidth]{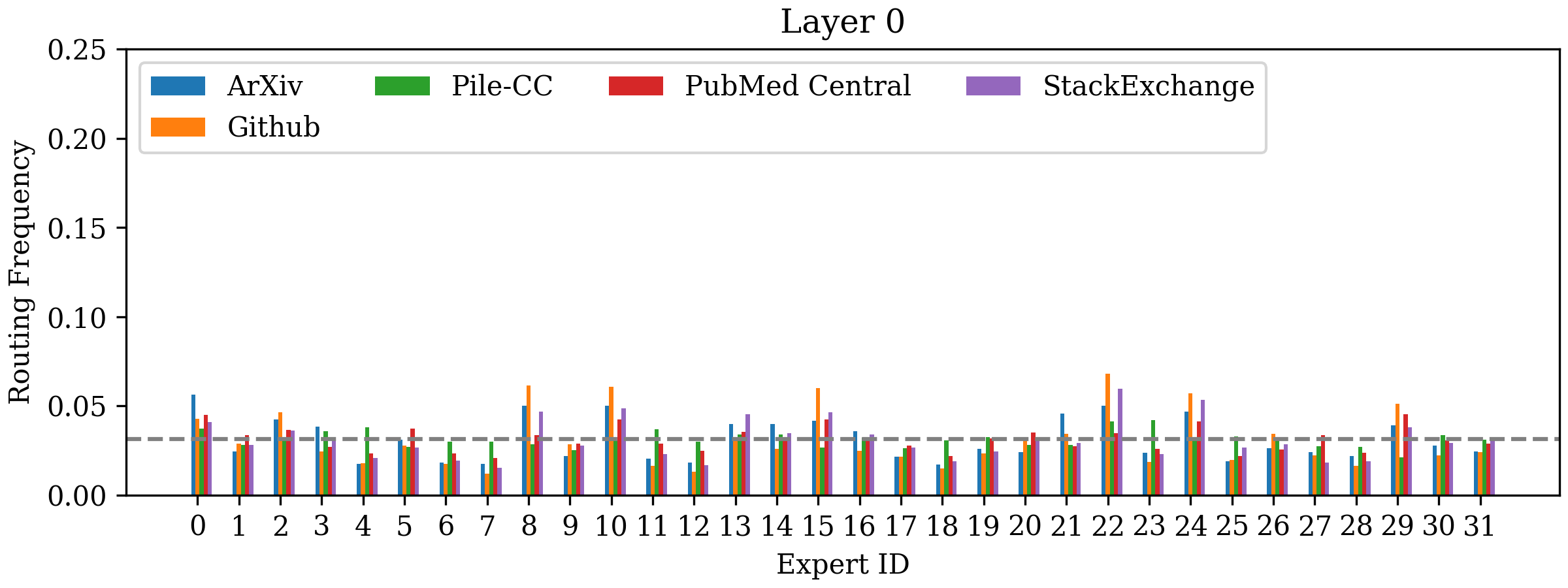}
    \caption{Domain Specialization for 32c4 TopKMoE at Layer 0.}
    \label{fig:appen-spec-topk-32c4-layer0}
\end{figure}

\begin{figure}[hbtp]
    \centering
    \includegraphics[width=0.5\linewidth]{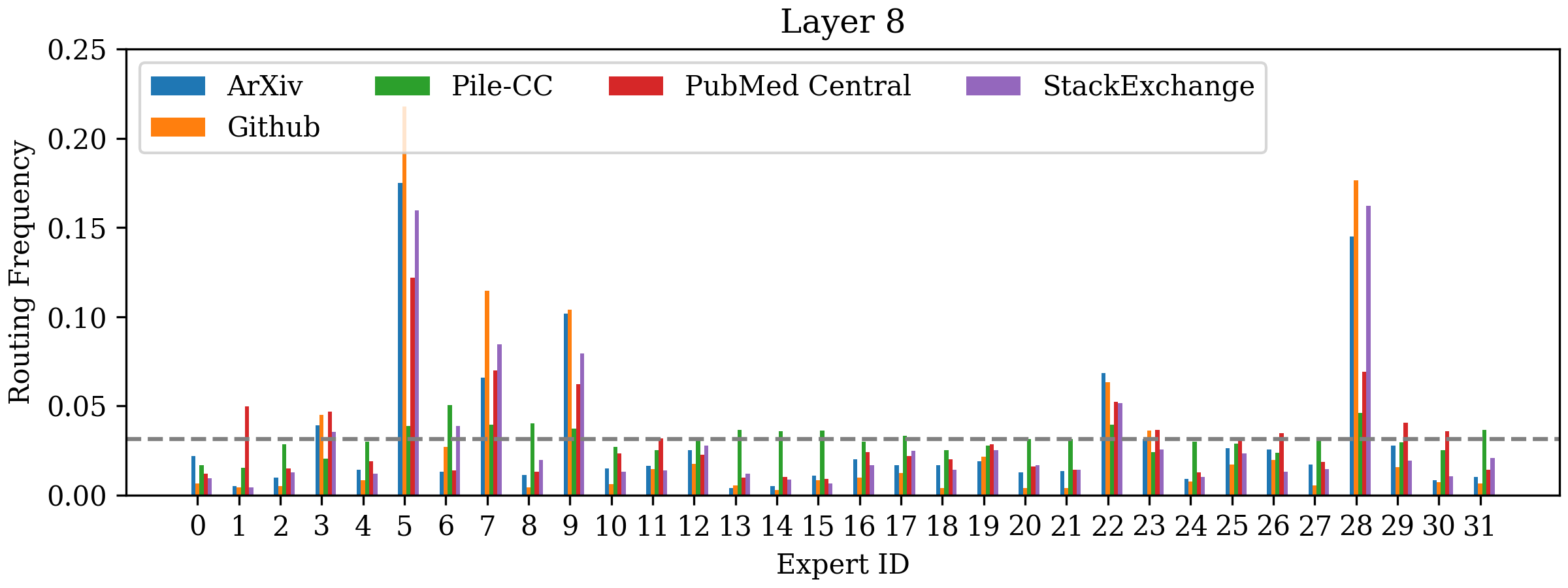}
    \caption{Domain Specialization for 32c4 TopKMoE at Layer 8.}
    \label{fig:appen-spec-topk-32c4-layer8}
\end{figure}

\begin{figure}[hbtp]
    \centering
    \includegraphics[width=0.5\linewidth]{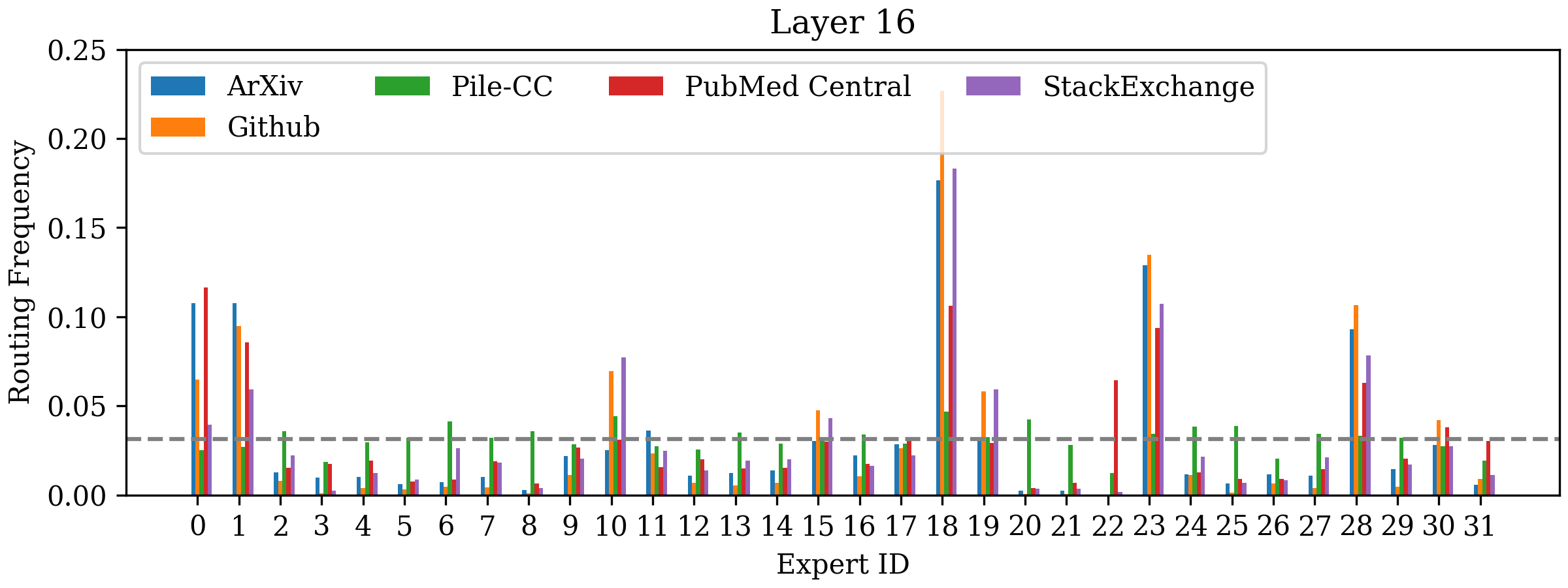}
    \caption{Domain Specialization for 32c4 TopKMoE at Layer 16.}
    \label{fig:appen-spec-topk-32c4-layer16}
\end{figure}

\begin{figure}[hbtp]
    \centering
    \includegraphics[width=0.5\linewidth]{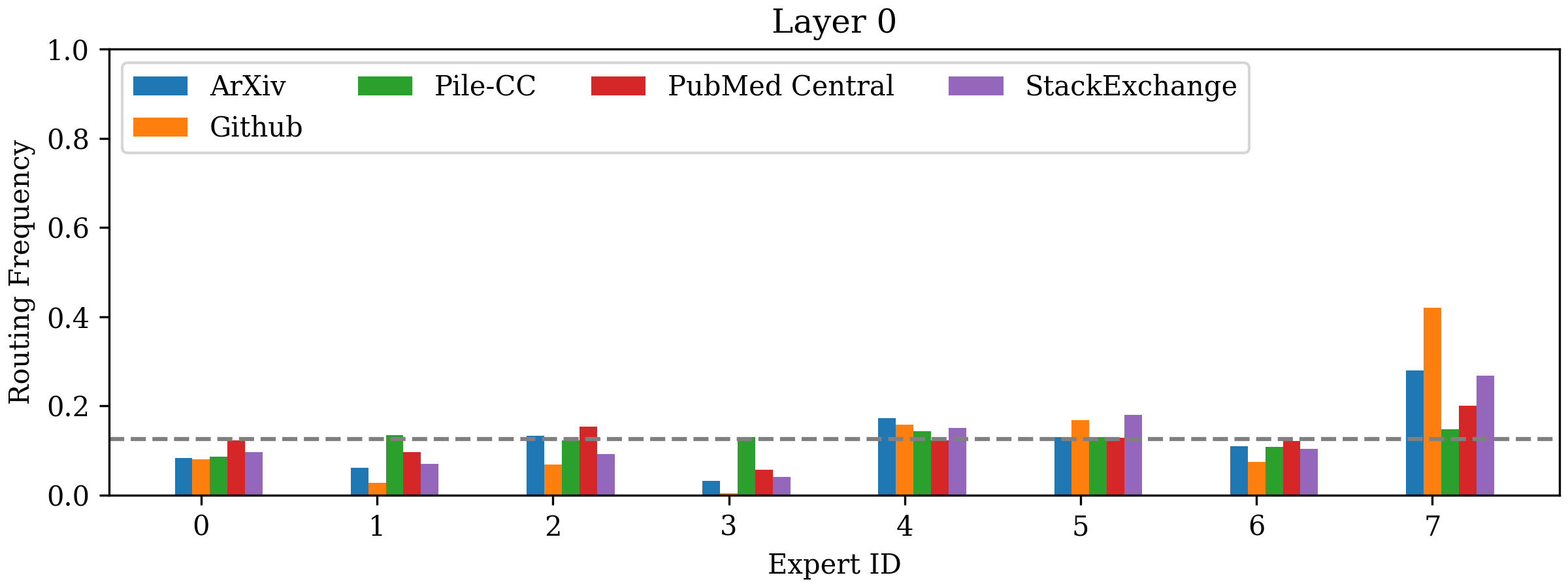}
    \caption{Domain Specialization for 8c1 DefaultMoE at Layer 0.}
    \label{fig:appen-spec-defvec-8c1-layer0}
\end{figure}

\begin{figure}[hbtp]
    \centering
    \includegraphics[width=0.5\linewidth]{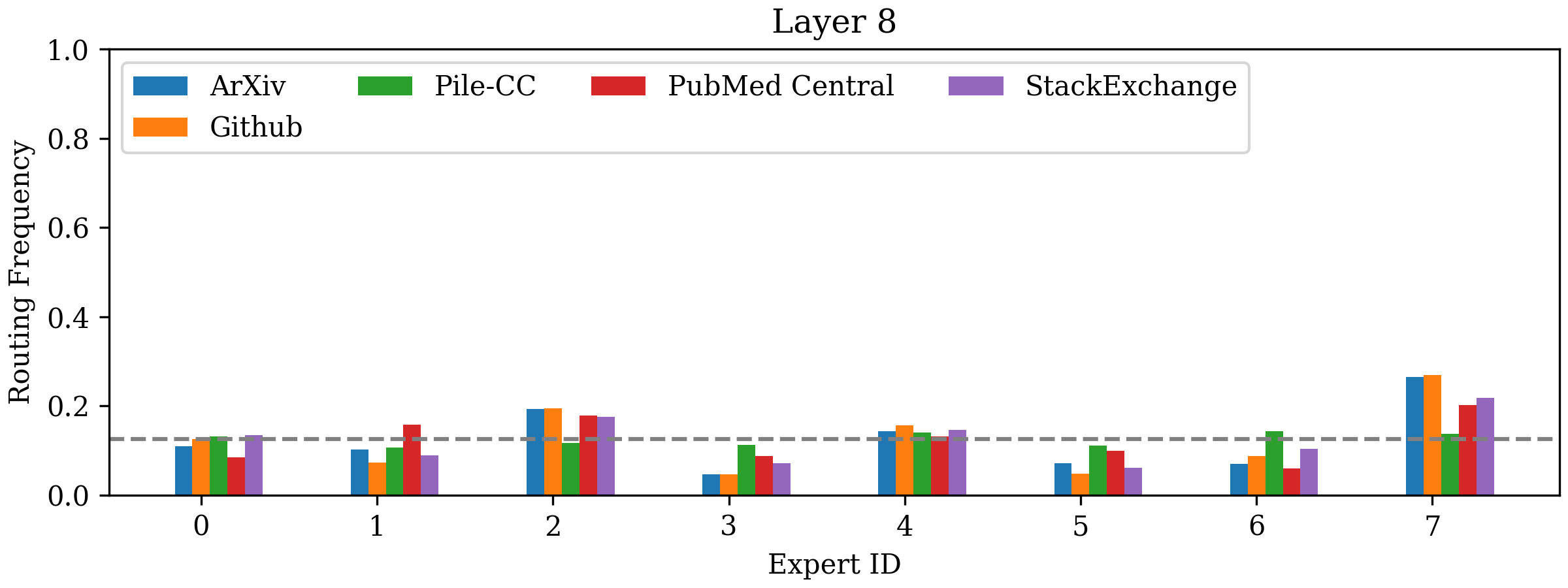}
    \caption{Domain Specialization for 8c1 DefaultMoE at Layer 8.}
    \label{fig:appen-spec-defvec-8c1-layer8}
\end{figure}

\begin{figure}[hbtp]
    \centering
    \includegraphics[width=0.5\linewidth]{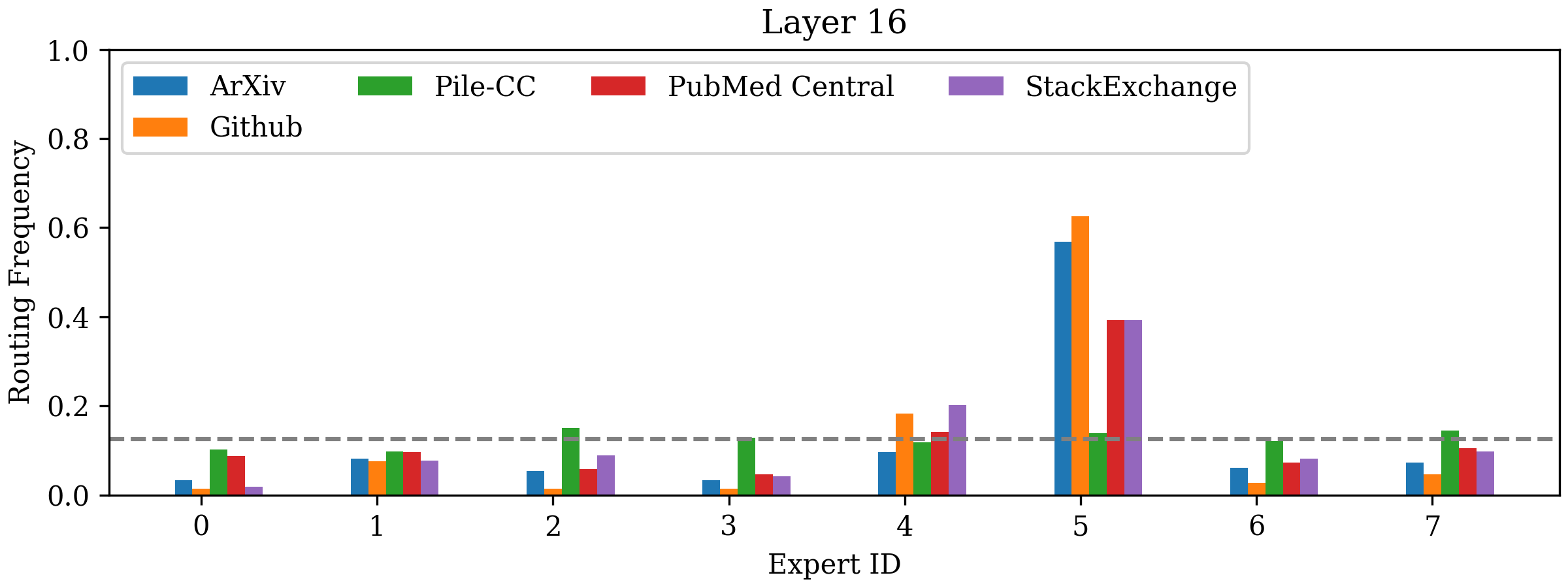}
    \caption{Domain Specialization for 8c1 DefaultMoE at Layer 16.}
    \label{fig:appen-spec-defvec-8c1-layer16}
\end{figure}

\begin{figure}[hbtp]
    \centering
    \includegraphics[width=0.5\linewidth]{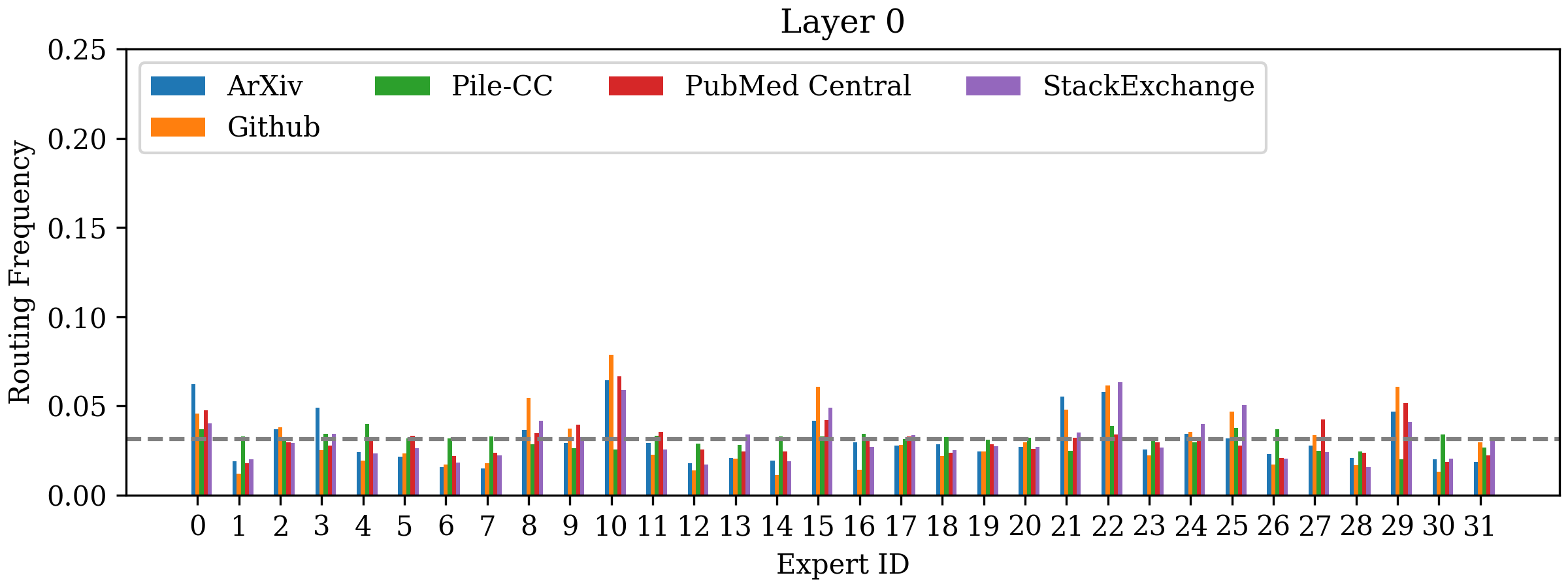}
    \caption{Domain Specialization for 32c4 DefaultMoE at Layer 0.}
    \label{fig:appen-spec-defvec-32c4-layer0}
\end{figure}

\begin{figure}[hbtp]
    \centering
    \includegraphics[width=0.5\linewidth]{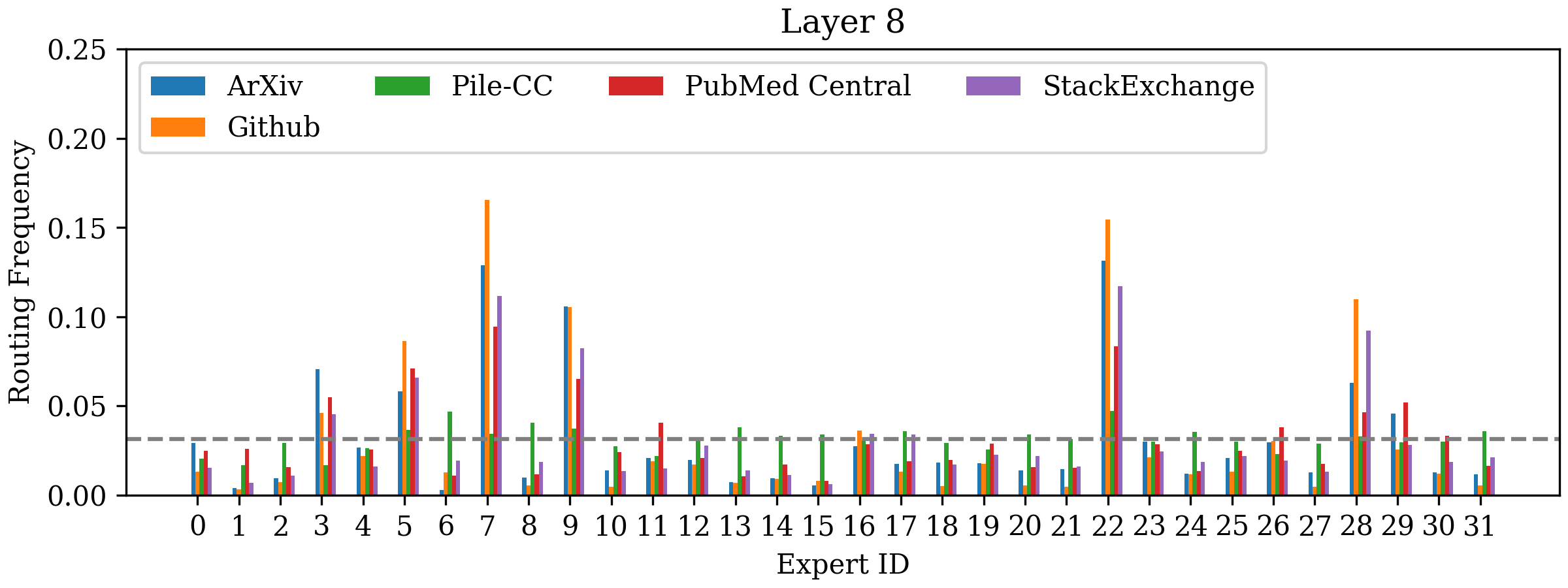}
    \caption{Domain Specialization for 32c4 DefaultMoE at Layer 8.}
    \label{fig:appen-spec-defvec-32c4-layer8}
\end{figure}

\begin{figure}[hbtp]
    \centering
    \includegraphics[width=0.5\linewidth]{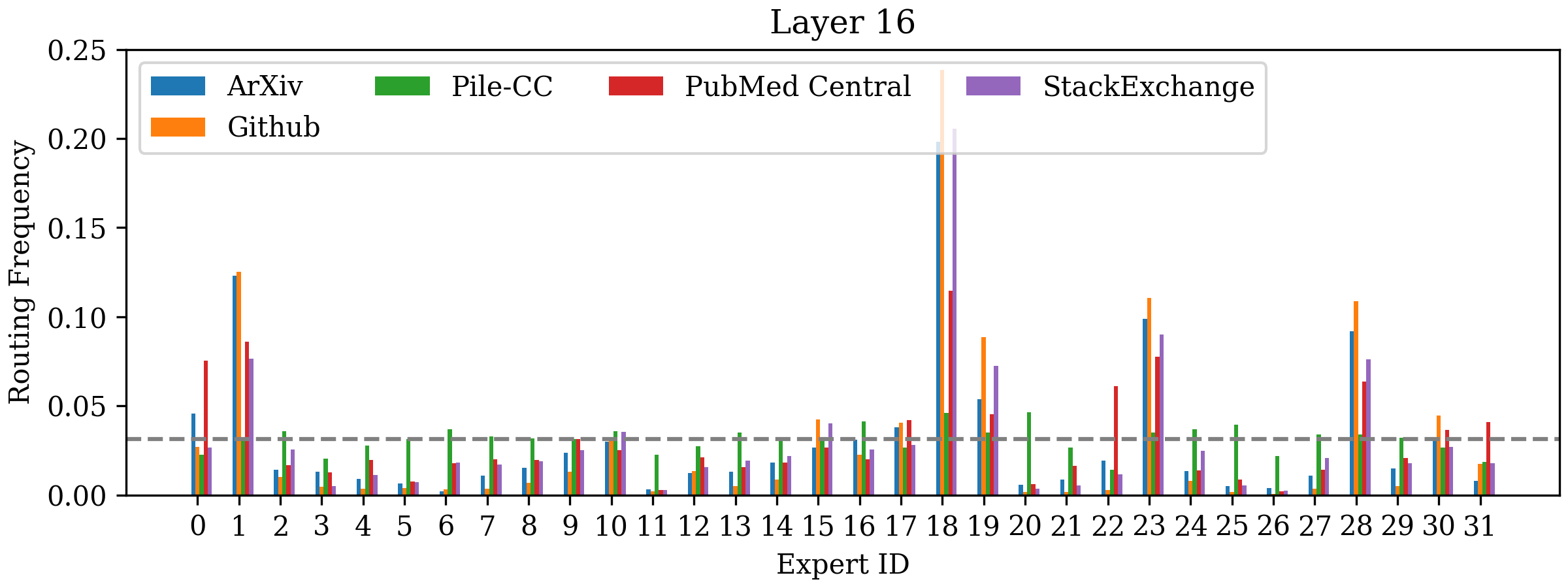}
    \caption{Domain Specialization for 32c4 DefaultMoE at Layer 16.}
    \label{fig:appen-spec-defvec-32c4-layer16}
\end{figure}

We plot the domain specialization for 8c1 TopKMoE~(\cref{fig:appen-spec-topk-8c1-layer0},\cref{fig:appen-spec-topk-8c1-layer8},\cref{fig:appen-spec-topk-8c1-layer16}), 8c1 DefaultMoE~(\cref{fig:appen-spec-defvec-8c1-layer0},\cref{fig:appen-spec-defvec-8c1-layer8}, \cref{fig:appen-spec-defvec-8c1-layer16}), 32c4 TopKMoE~(\cref{fig:appen-spec-topk-32c4-layer0},\cref{fig:appen-spec-topk-32c4-layer8},\cref{fig:appen-spec-topk-32c4-layer16}) and 32c4 DefaultMoE~(\cref{fig:appen-spec-defvec-32c4-layer0},\cref{fig:appen-spec-defvec-32c4-layer8},\cref{fig:appen-spec-defvec-32c4-layer16}). All models are pretrained on FineWeb-Edu for 160B tokens.

\subsection{Load Balancing}

\begin{figure}[hbtp]
    \centering
    \includegraphics[width=0.5\linewidth]{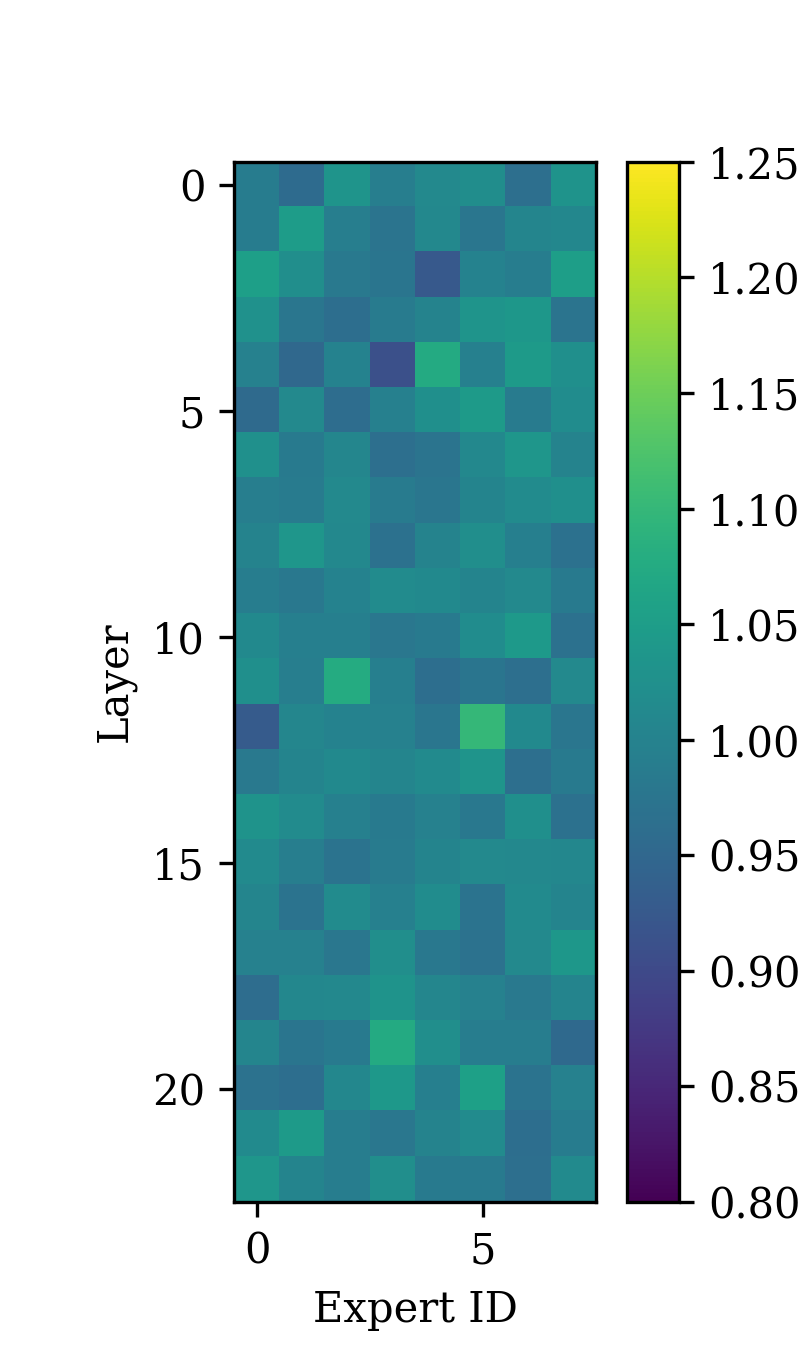}
    \caption{Routing Frequency for 8c1 TopKMoE.}
    \label{fig:appen-routing-topk-8c1}
\end{figure}

\begin{figure}[hbtp]
    \centering
    \includegraphics[width=0.5\linewidth]{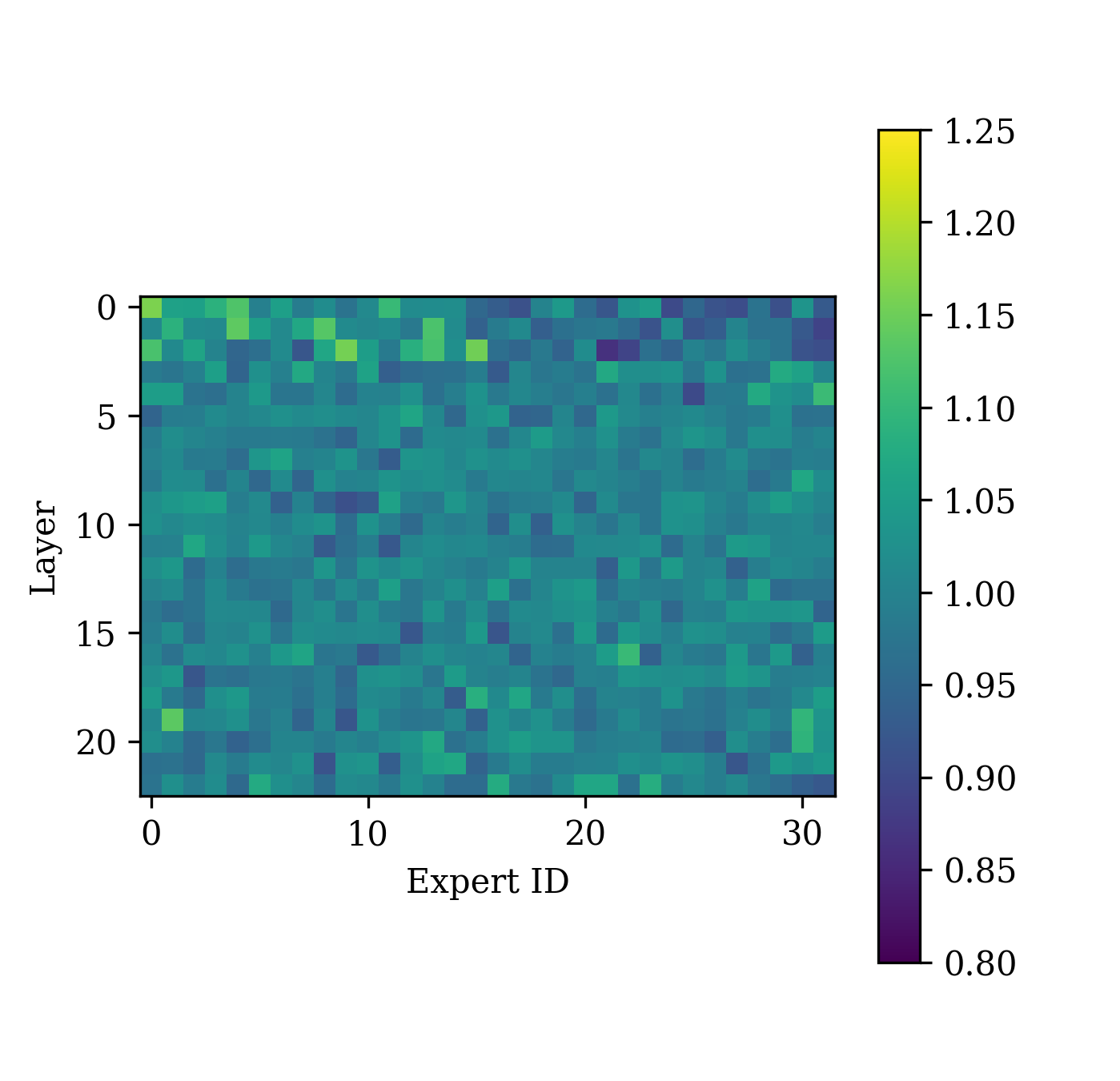}
    \caption{Routing Frequency for 32c4 TopKMoE.}
    \label{fig:appen-routing-topk-32c4}
\end{figure}

\begin{figure}[hbtp]
    \centering
    \includegraphics[width=0.5\linewidth]{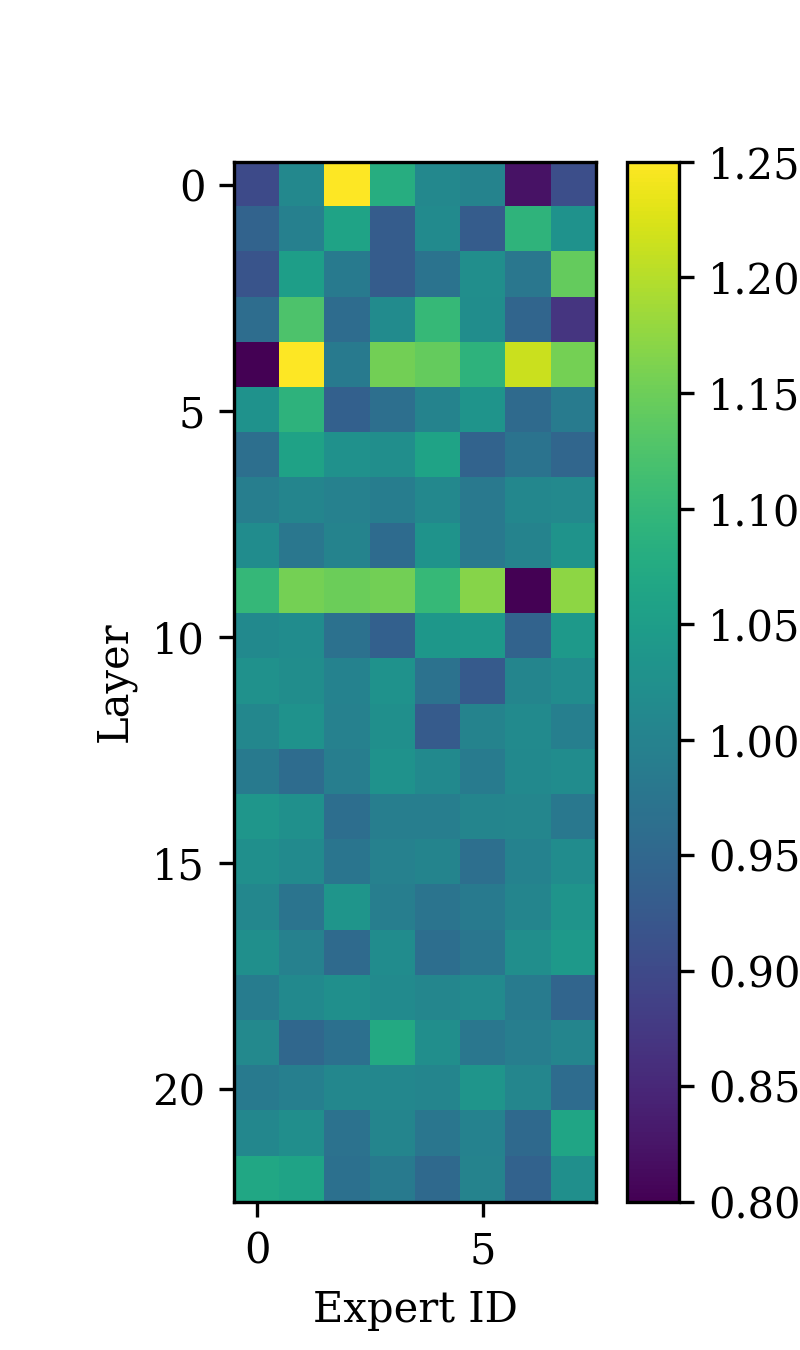}
    \caption{Routing Frequency for 8c1 DefaultMoE.}
    \label{fig:appen-routing-defvec-8c1}
\end{figure}

\begin{figure}[hbtp]
    \centering
    \includegraphics[width=0.5\linewidth]{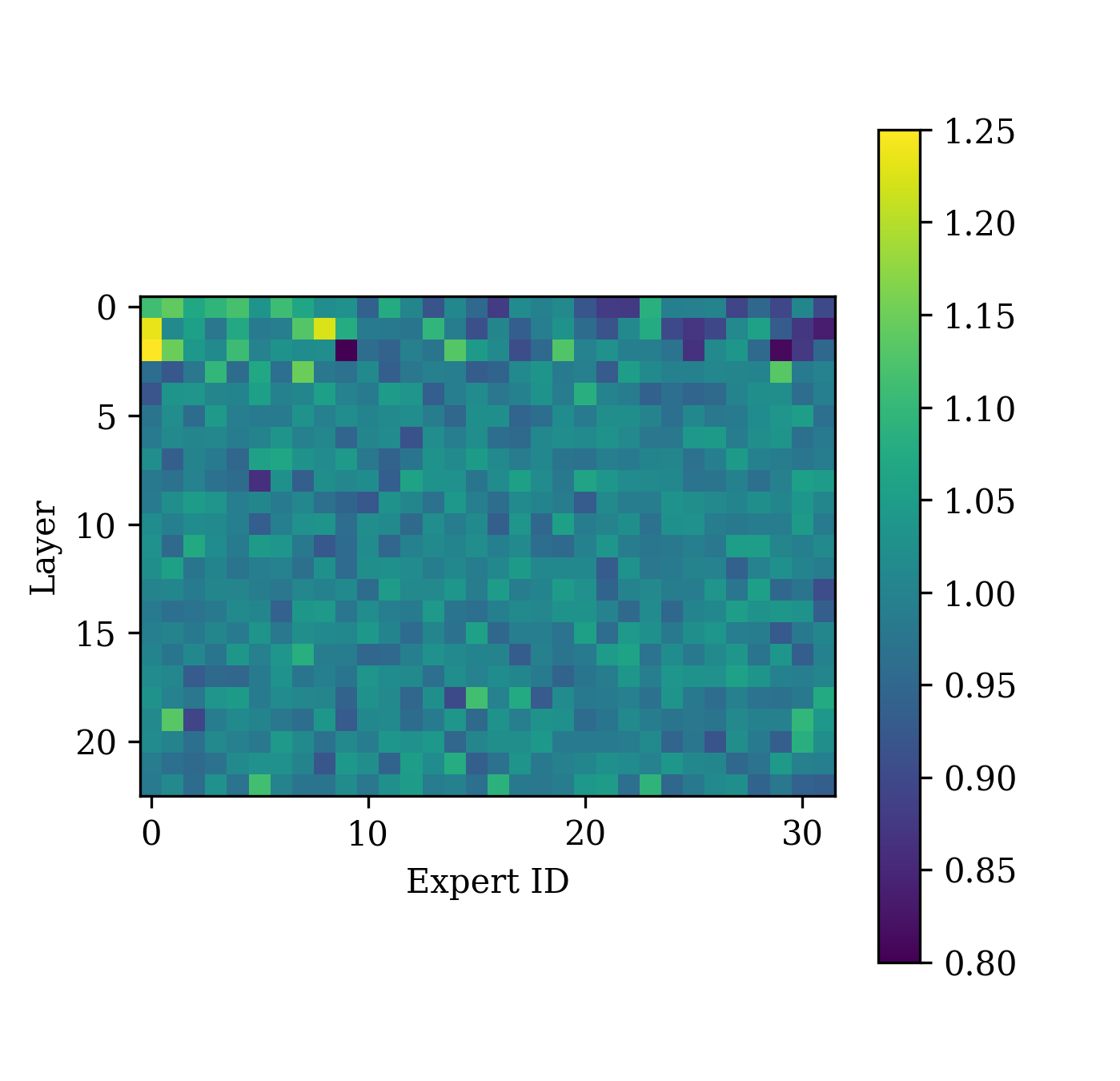}
    \caption{Routing Frequency for 32c4 DefaultMoE.}
    \label{fig:appen-routing-defvec-32c4}
\end{figure}

We plot the load balancing for 8c1 TopKMoE~(\cref{fig:appen-routing-topk-8c1}), 32c4 TopKMoE~(\cref{fig:appen-routing-topk-32c4}), 8c1 DefaultMoE~(\cref{fig:appen-routing-defvec-8c1}) and 32c4 DefaultMoE~(\cref{fig:appen-routing-defvec-32c4}). All models are pretrained on FineWeb-Edu for 160B tokens.

\subsection{Expert Coactivation}

\begin{figure}[hbtp]
    \centering
    \includegraphics[width=0.5\linewidth]{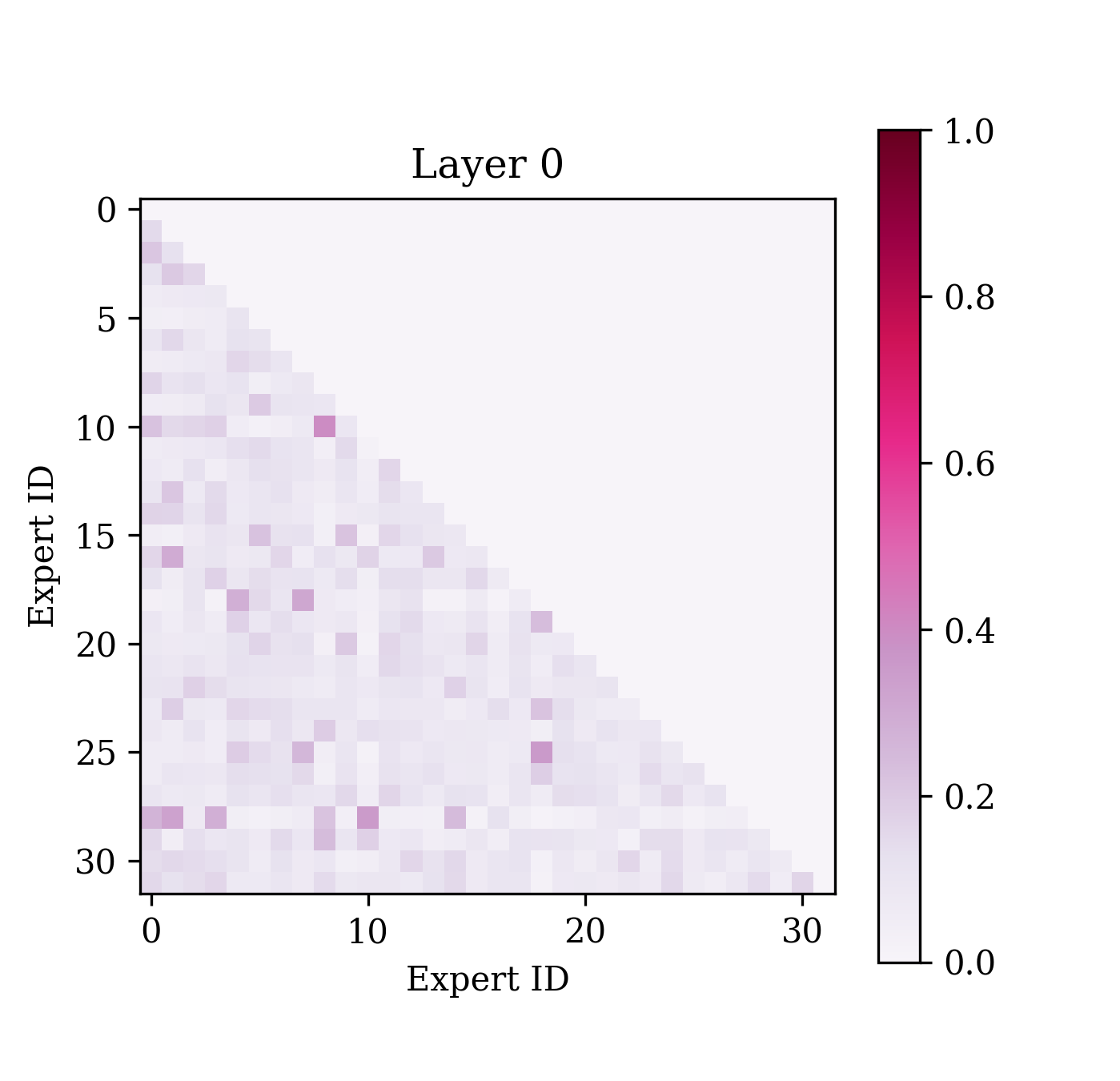}
    \caption{Expert Coactivation for 32c4 DefaultMoE at Layer 0.}
    \label{fig:appen-coact-defvec-32c4-layer0}
\end{figure}

\begin{figure}[hbtp]
    \centering
    \includegraphics[width=0.5\linewidth]{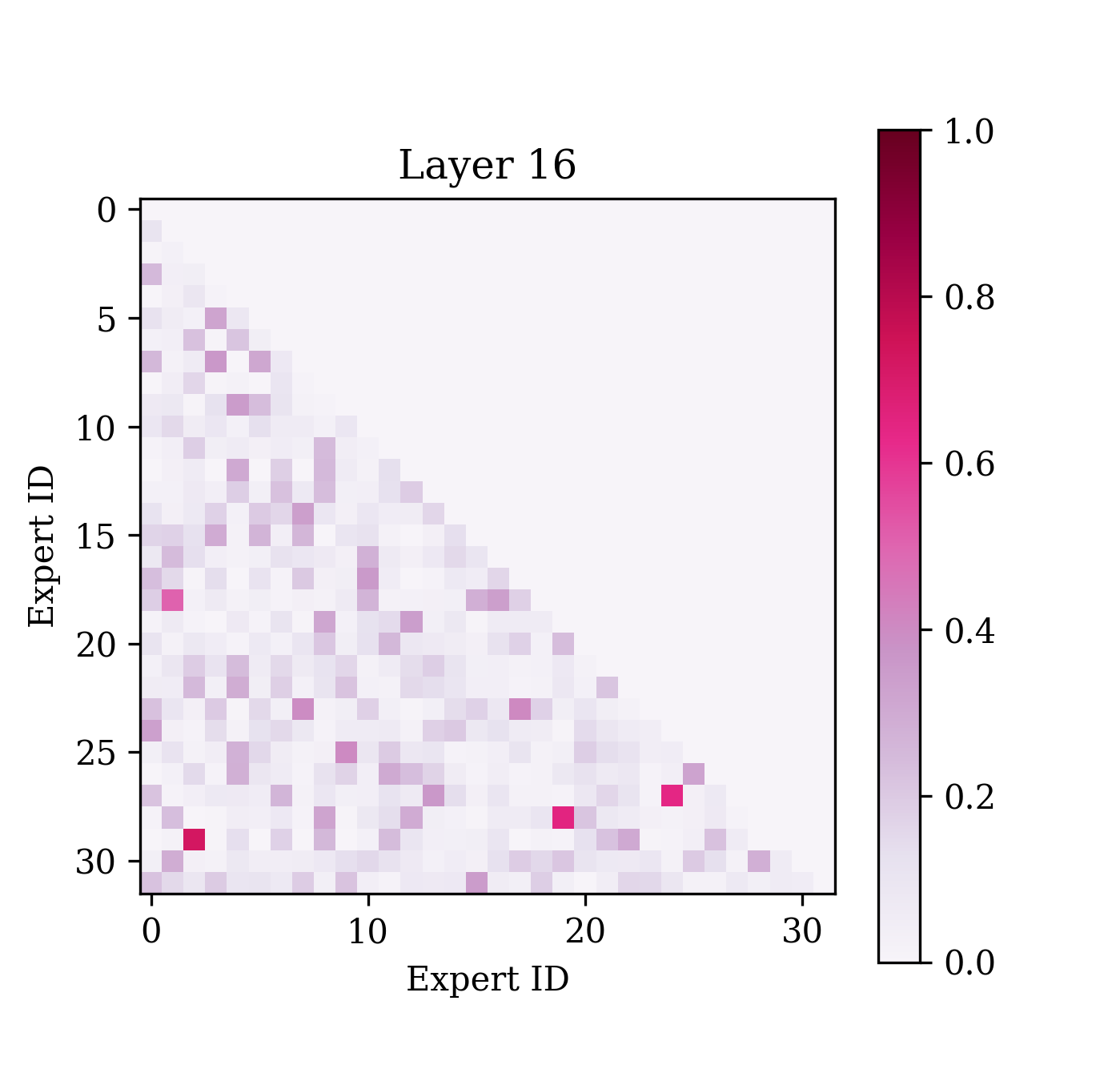}
    \caption{Expert Coactivation for 32c4 DefaultMoE at Layer 16.}
    \label{fig:appen-coact-defvec-32c4-layer16}
\end{figure}

We plot the expert coactivation for 32c4 DefaultMoE at Layer 0~(\cref{fig:appen-coact-defvec-32c4-layer0}) and Layer 16~(\cref{fig:appen-coact-defvec-32c4-layer16}). All models are pretrained on FineWeb-Edu for 160B tokens.

\subsection{Dense Gradient Similarity}

\begin{figure}[hbtp]
    \centering
    \includegraphics[width=0.5\linewidth]{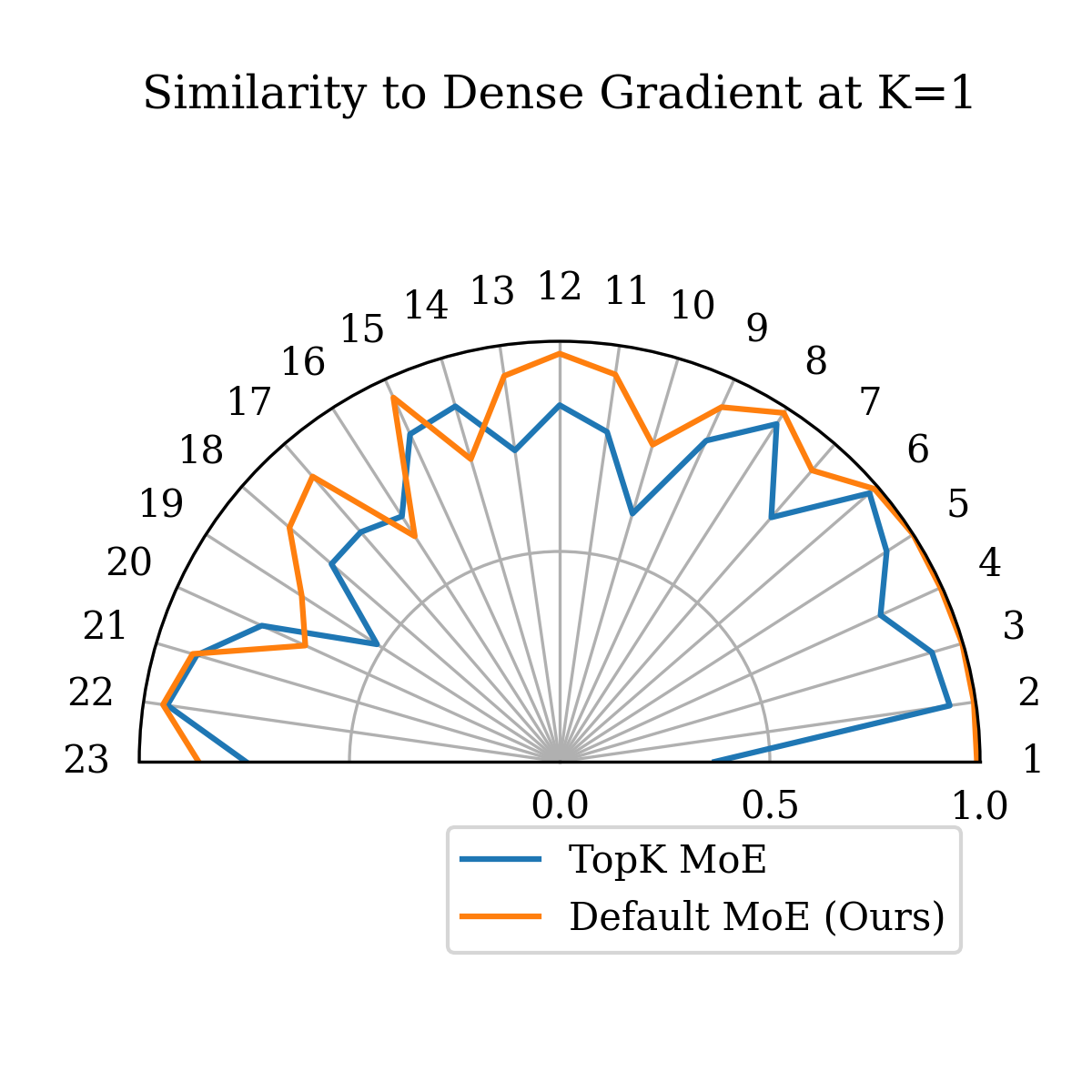}
    \caption{Similarity to Dense Gradient for 8c1 TopKMoE and DefaultMoE at K=1.}
    \label{fig:appen-denseapprox-8c1-k1}
\end{figure}

\begin{figure}[hbtp]
    \centering
    \includegraphics[width=0.5\linewidth]{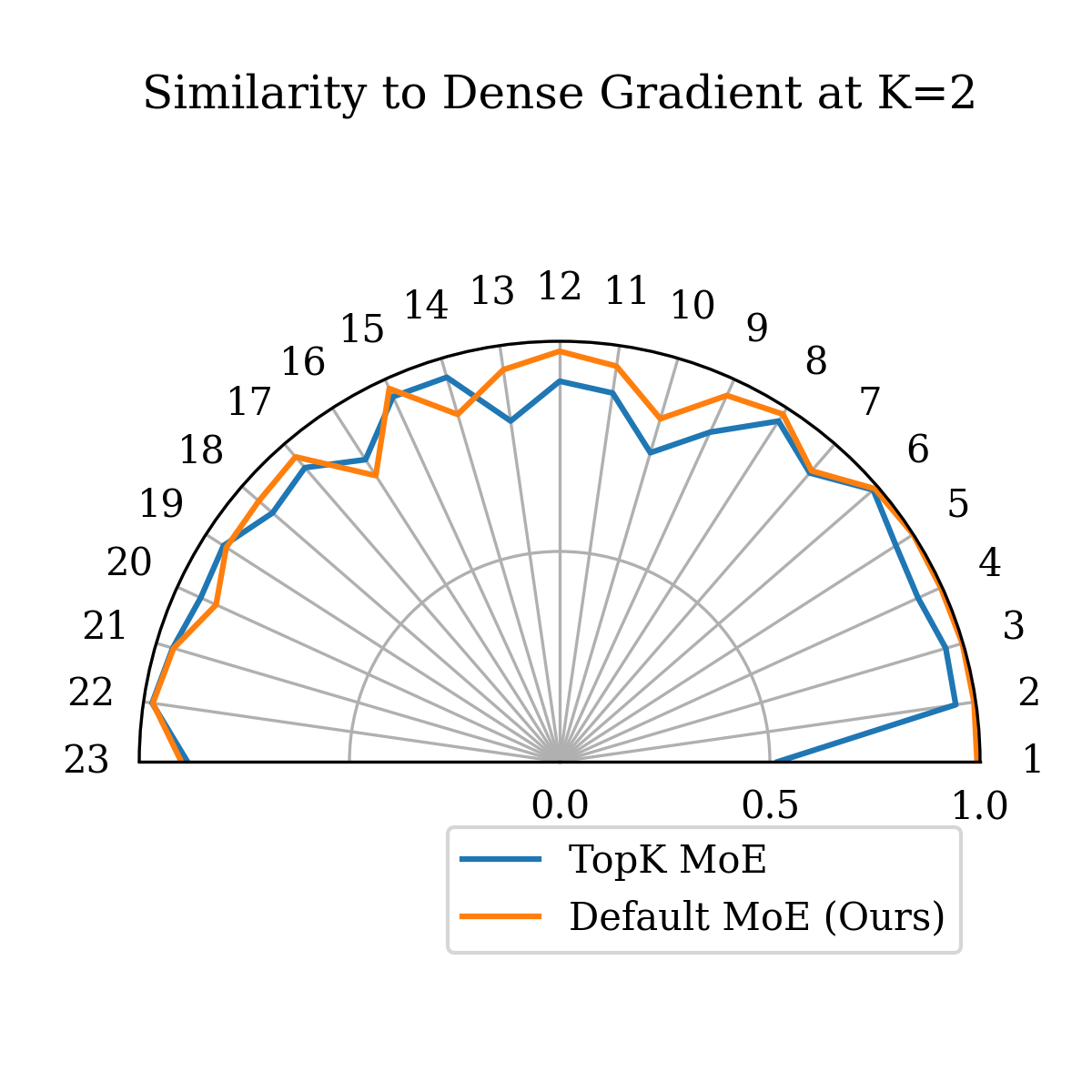}
    \caption{Similarity to Dense Gradient for 8c1 TopKMoE and DefaultMoE at K=2.}
    \label{fig:appen-denseapprox-8c1-k2}
\end{figure}

\begin{figure}[hbtp]
    \centering
    \includegraphics[width=0.5\linewidth]{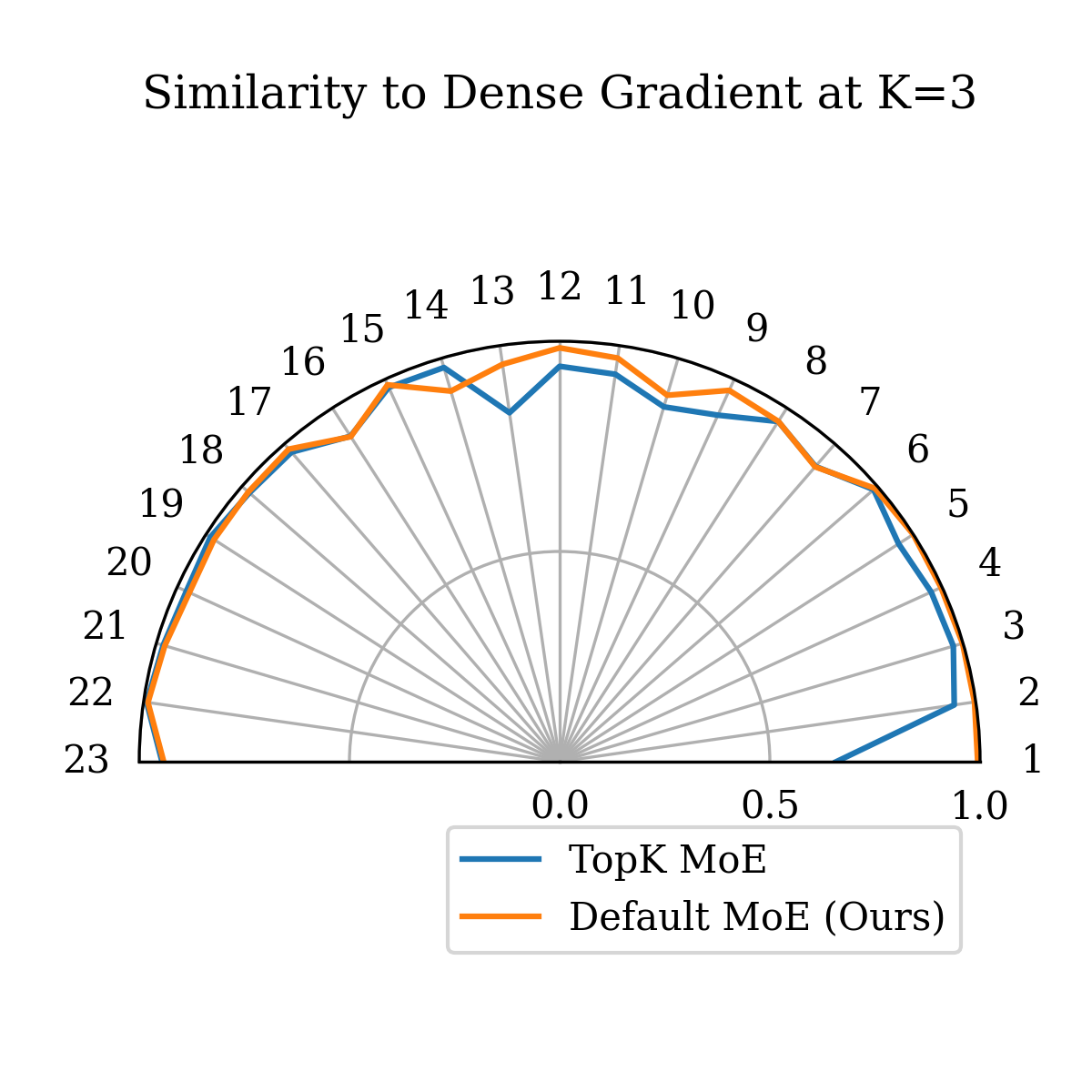}
    \caption{Similarity to Dense Gradient for 8c1 TopKMoE and DefaultMoE at K=3.}
    \label{fig:appen-denseapprox-8c1-k3}
\end{figure}

\begin{figure}[hbtp]
    \centering
    \includegraphics[width=0.5\linewidth]{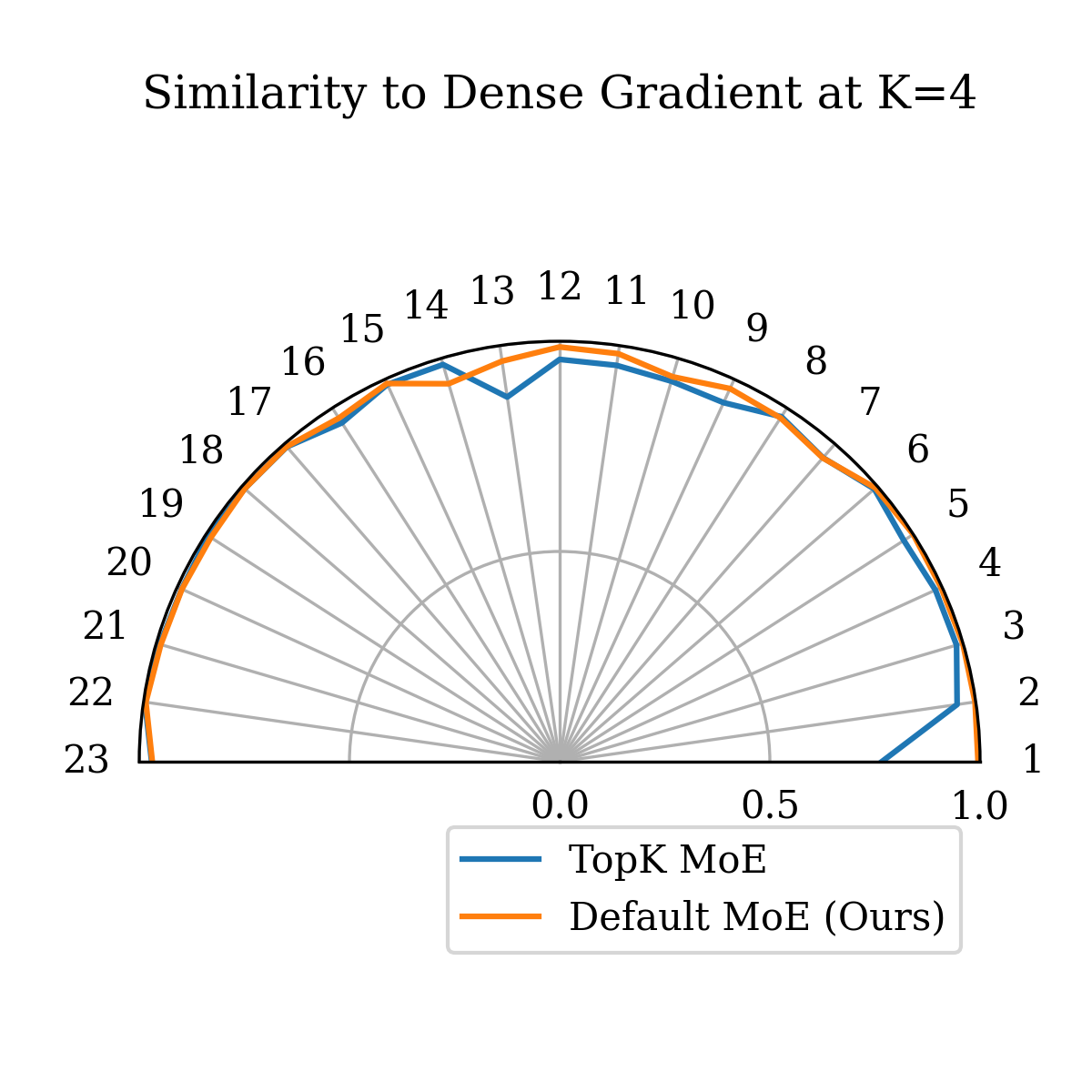}
    \caption{Similarity to Dense Gradient for 8c1 TopKMoE and DefaultMoE at K=4.}
    \label{fig:appen-denseapprox-8c1-k4}
\end{figure}

\begin{figure}[hbtp]
    \centering
    \includegraphics[width=0.5\linewidth]{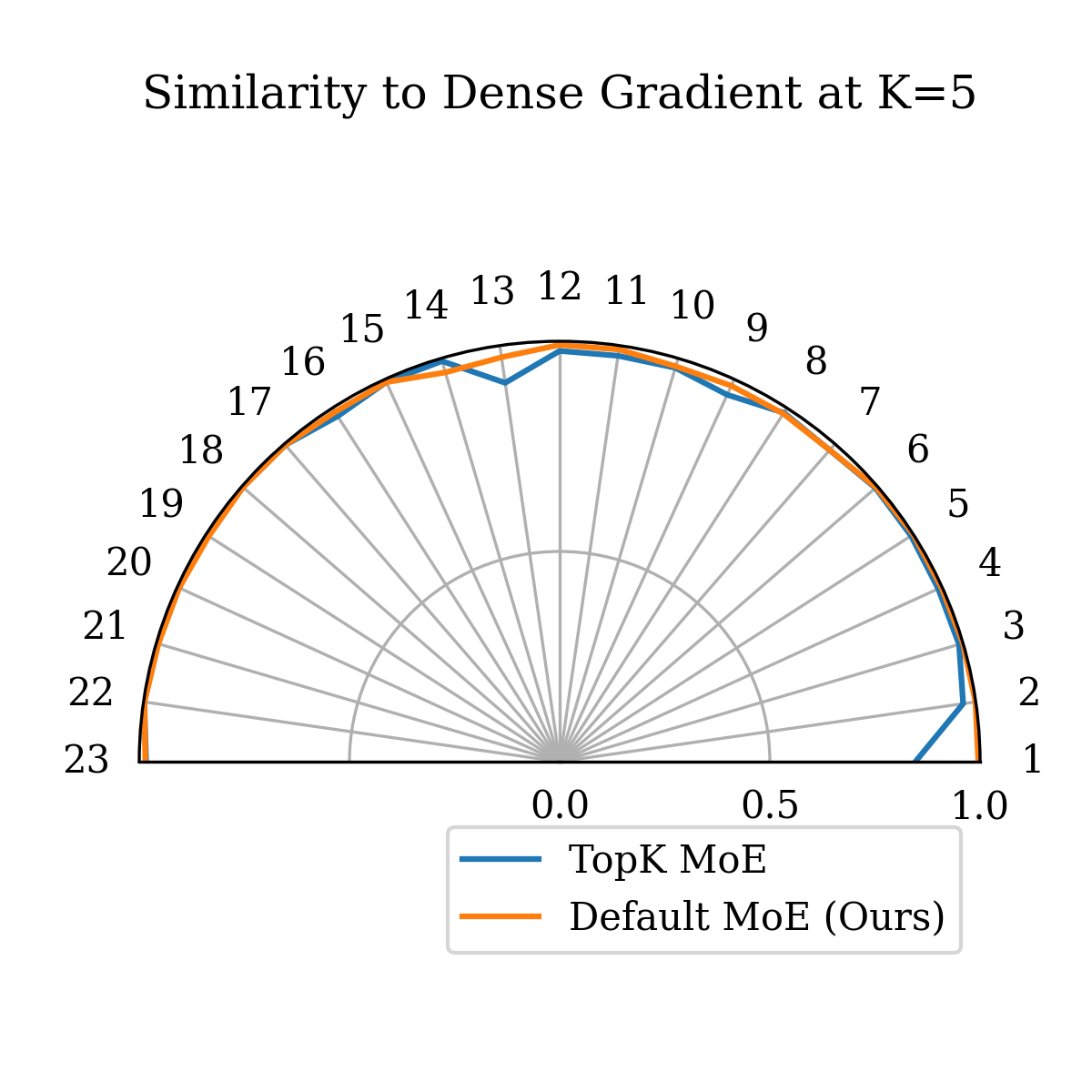}
    \caption{Similarity to Dense Gradient for 8c1 TopKMoE and DefaultMoE at K=5.}
    \label{fig:appen-denseapprox-8c1-k5}
\end{figure}

\begin{figure}[hbtp]
    \centering
    \includegraphics[width=0.5\linewidth]{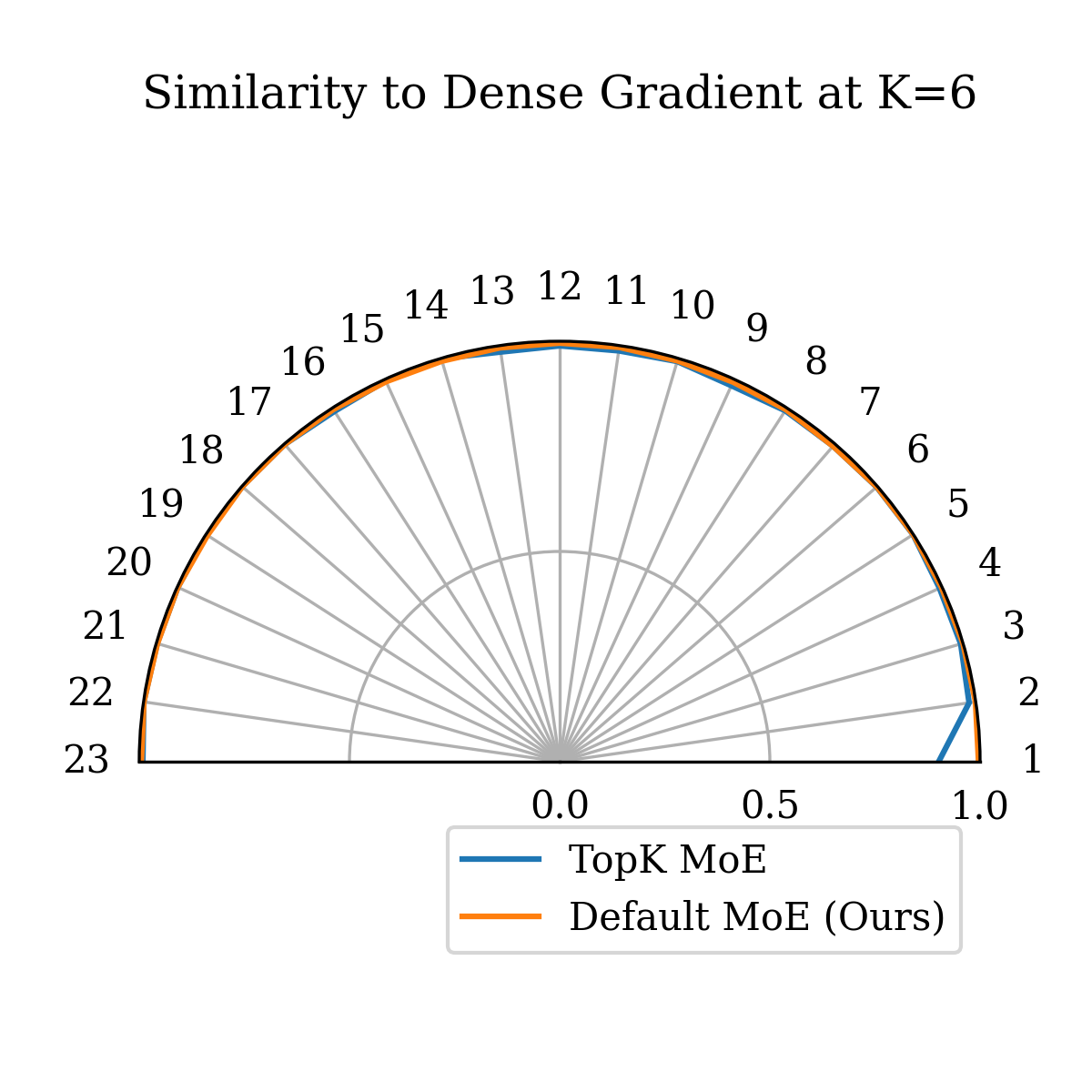}
    \caption{Similarity to Dense Gradient for 8c1 TopKMoE and DefaultMoE at K=6.}
    \label{fig:appen-denseapprox-8c1-k6}
\end{figure}

\begin{figure}[hbtp]
    \centering
    \includegraphics[width=0.5\linewidth]{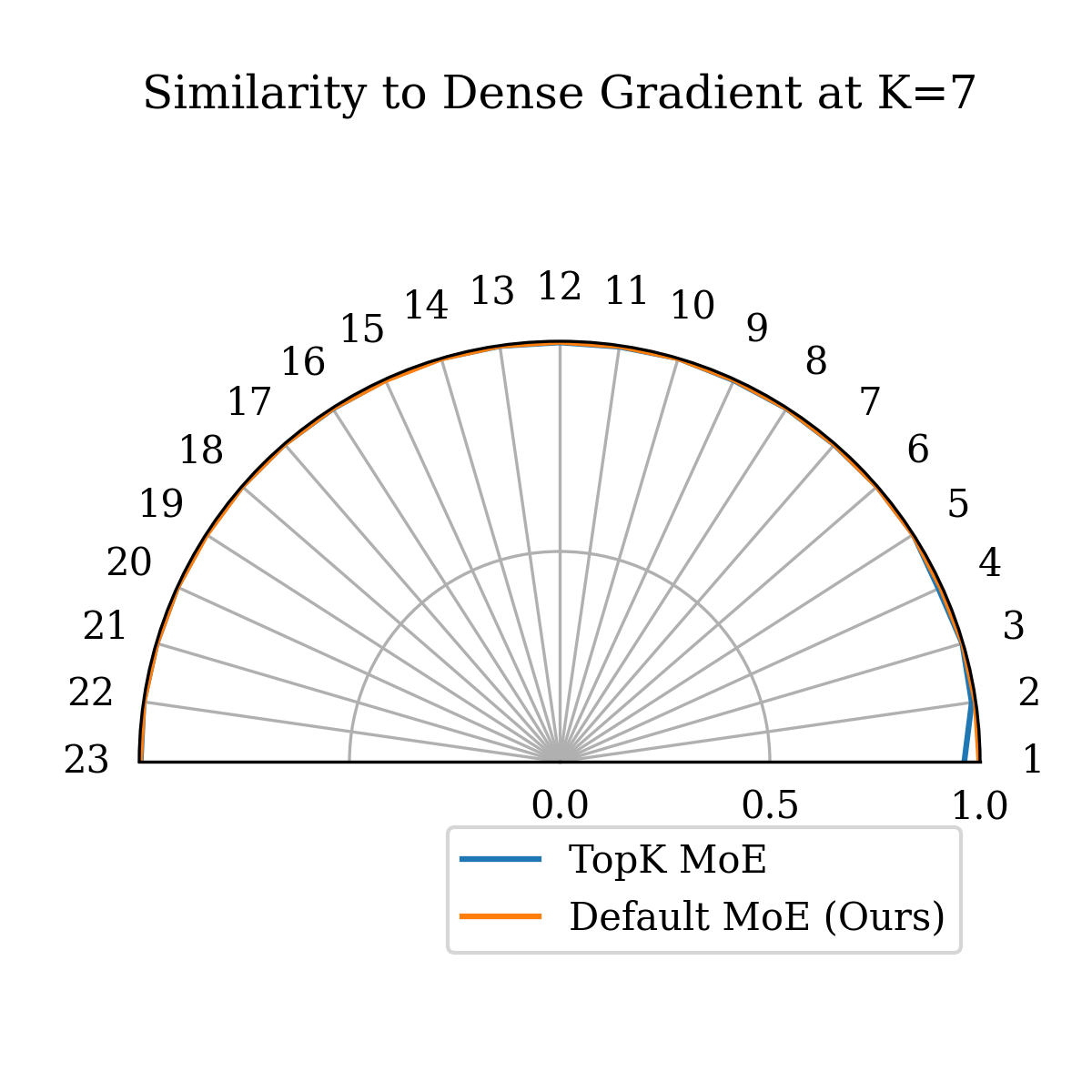}
    \caption{Similarity to Dense Gradient for 8c1 TopKMoE and DefaultMoE at K=7.}
    \label{fig:appen-denseapprox-8c1-k7}
\end{figure}

We plot the similarity to the dense gradient for 8c1 TopKMoE and DefaultMoE at K=1~(\cref{fig:appen-denseapprox-8c1-k1}), K=2~(\cref{fig:appen-denseapprox-8c1-k2}), K=3~(\cref{fig:appen-denseapprox-8c1-k3}), K=4~(\cref{fig:appen-denseapprox-8c1-k4}), K=5~(\cref{fig:appen-denseapprox-8c1-k5}), K=6~(\cref{fig:appen-denseapprox-8c1-k6}) and K=7~(\cref{fig:appen-denseapprox-8c1-k7}). All models are pretrained on FineWeb-Edu for 160B tokens.

\subsection{Default Vector Similarities}

\begin{figure}[hbtp]
    \centering
    \includegraphics[width=0.5\linewidth]{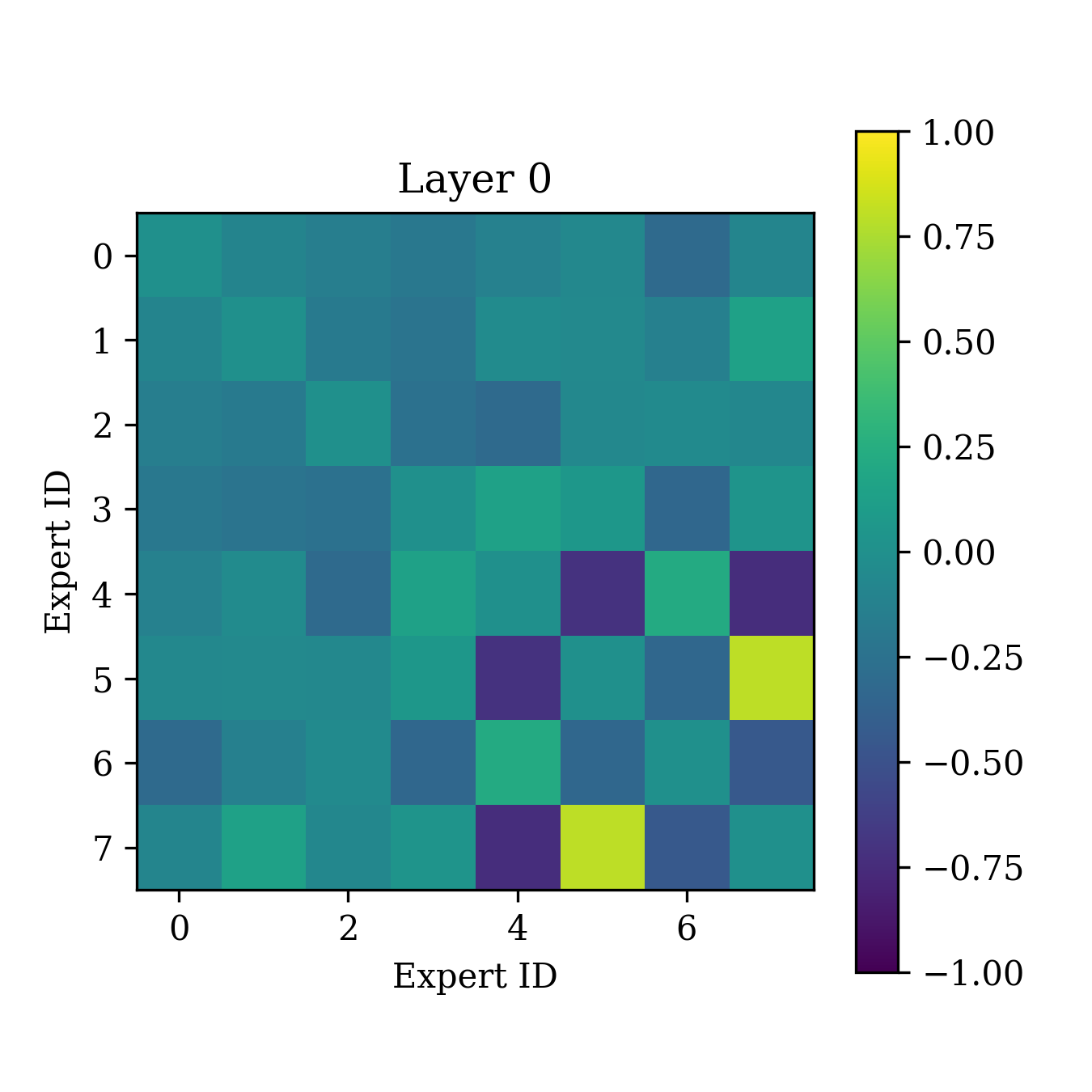}
    \caption{Similarities Between Default Vectors for 8c1 DefaultMoE at Layer 0.}
    \label{fig:appen-similarities-8c1-layer0}
\end{figure}

\begin{figure}[hbtp]
    \centering
    \includegraphics[width=0.5\linewidth]{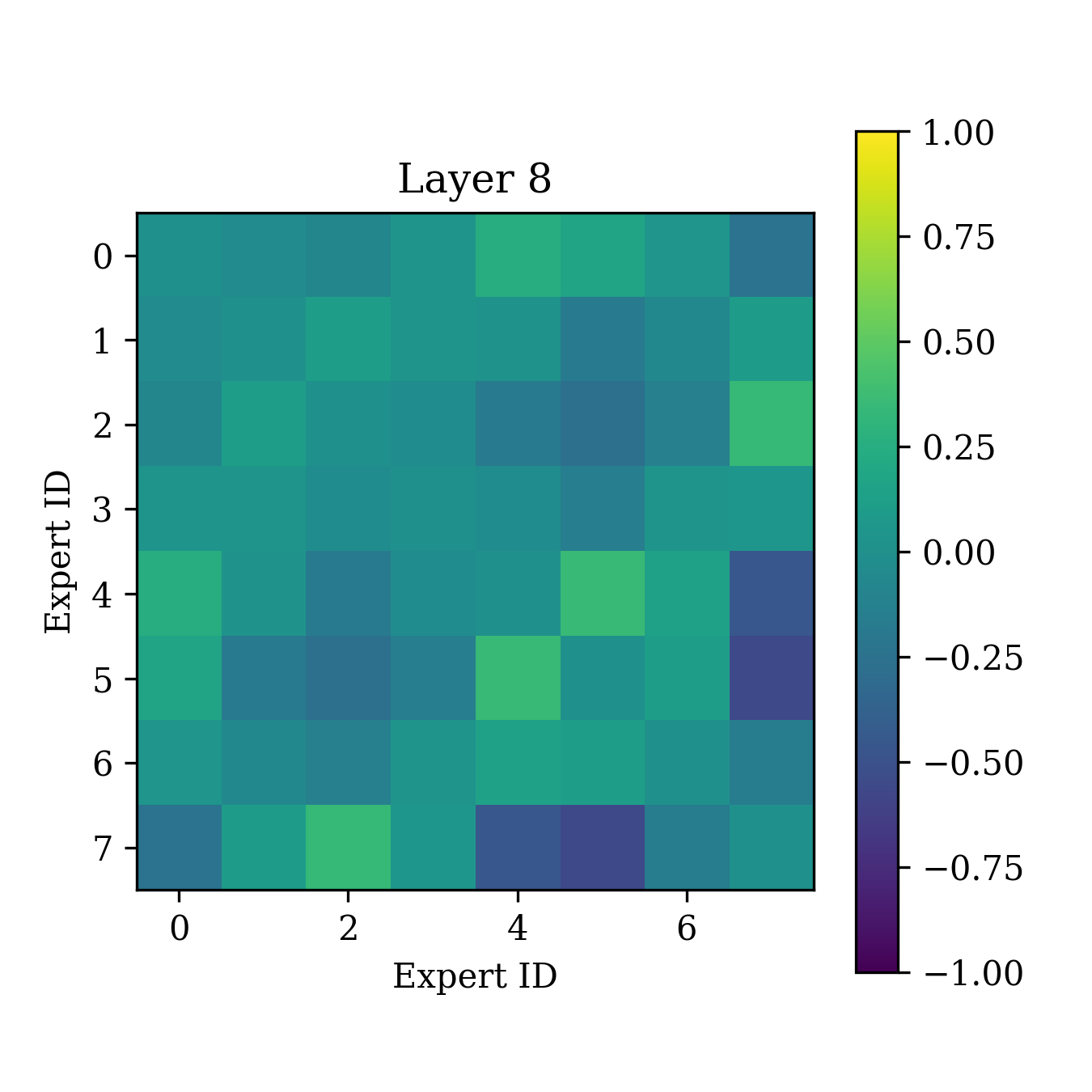}
    \caption{Similarities Between Default Vectors for 8c1 DefaultMoE at Layer 8.}
    \label{fig:appen-similarities-8c1-layer8}
\end{figure}

\begin{figure}[hbtp]
    \centering
    \includegraphics[width=0.5\linewidth]{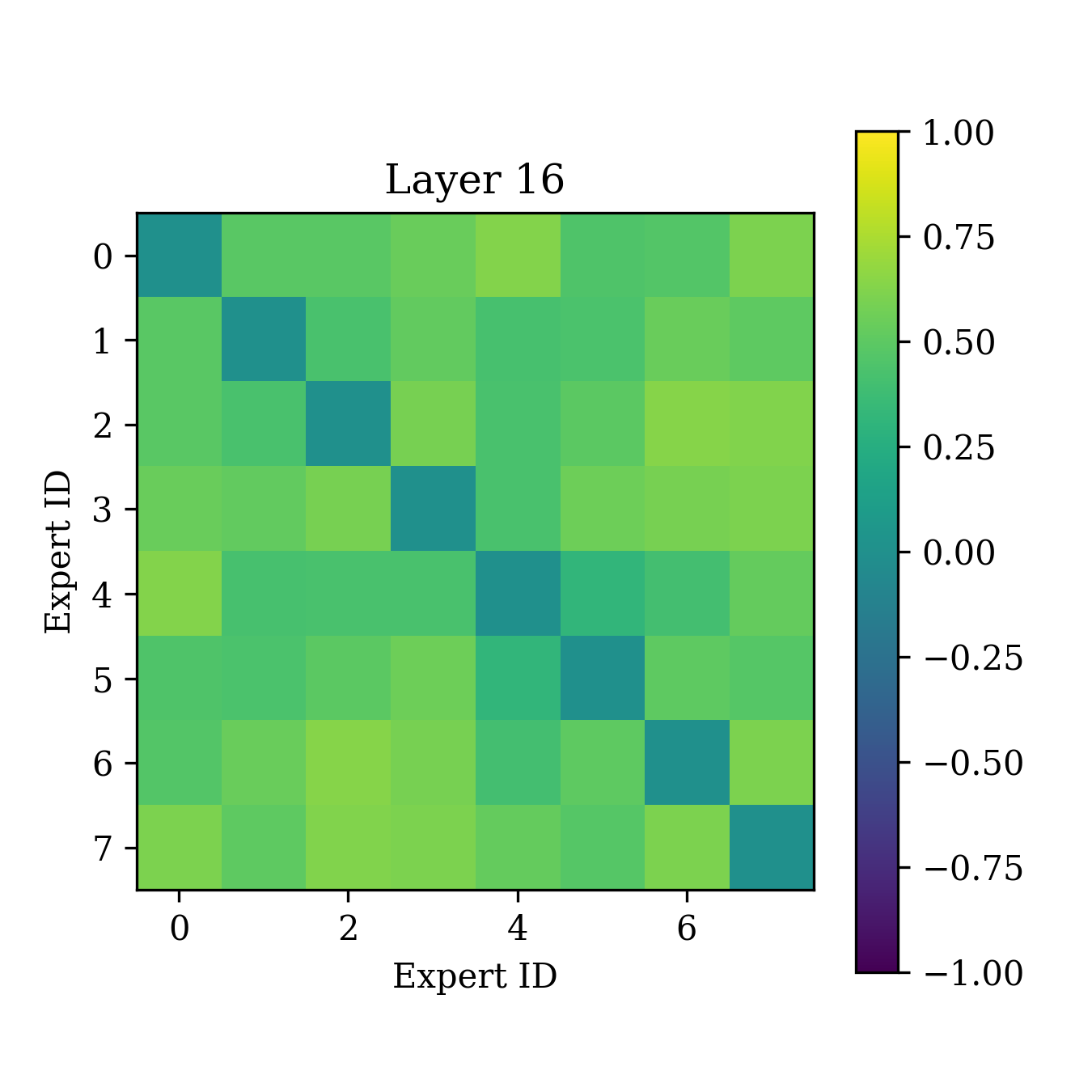}
    \caption{Similarities Between Default Vectors for 8c1 DefaultMoE at Layer 16.}
    \label{fig:appen-similarities-8c1-layer16}
\end{figure}

\begin{figure}[hbtp]
    \centering
    \includegraphics[width=0.5\linewidth]{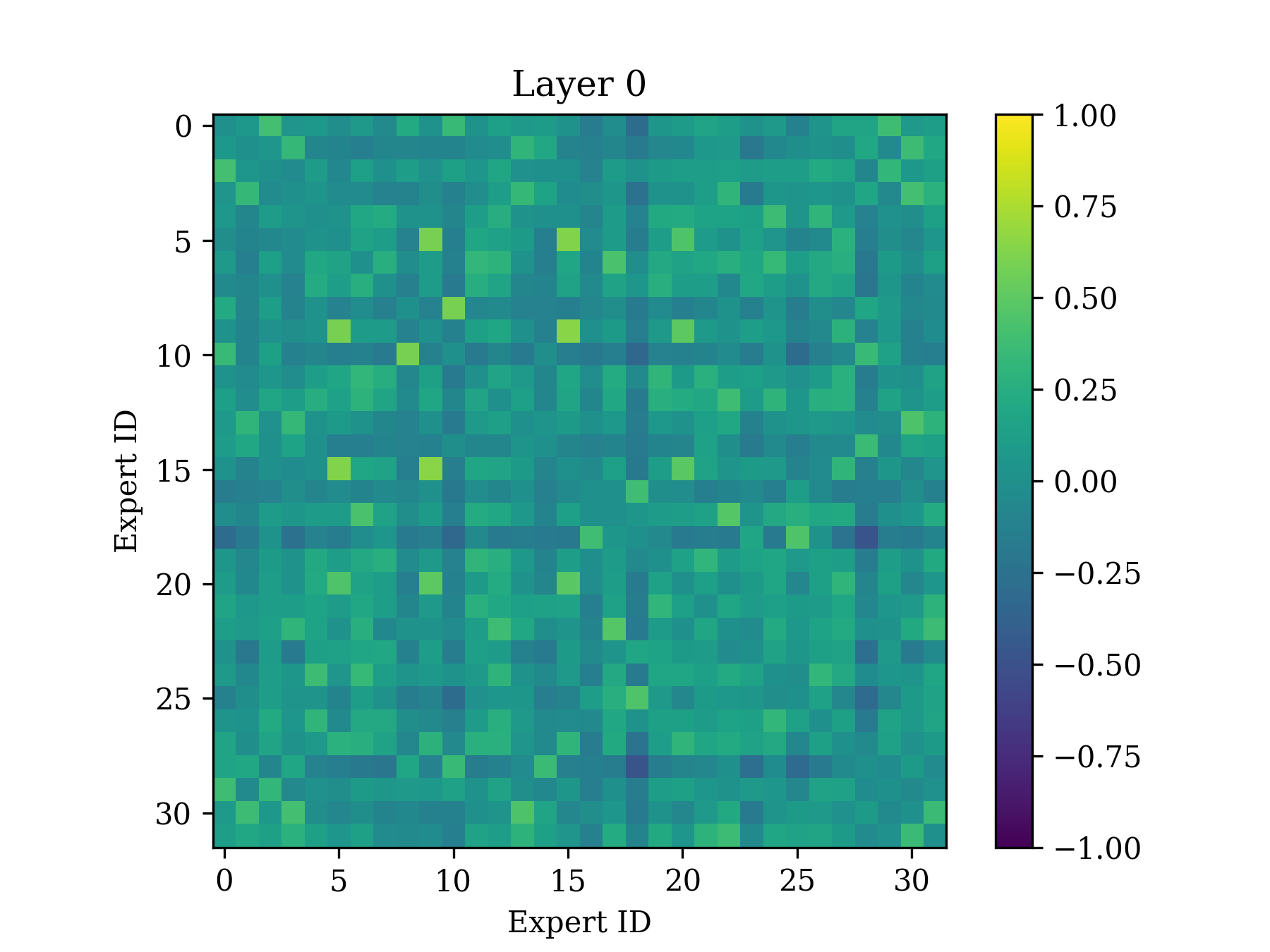}
    \caption{Similarities Between Default Vectors for 32c4 DefaultMoE at Layer 0.}
    \label{fig:appen-similarities-32c4-layer0}
\end{figure}

\begin{figure}[hbtp]
    \centering
    \includegraphics[width=0.5\linewidth]{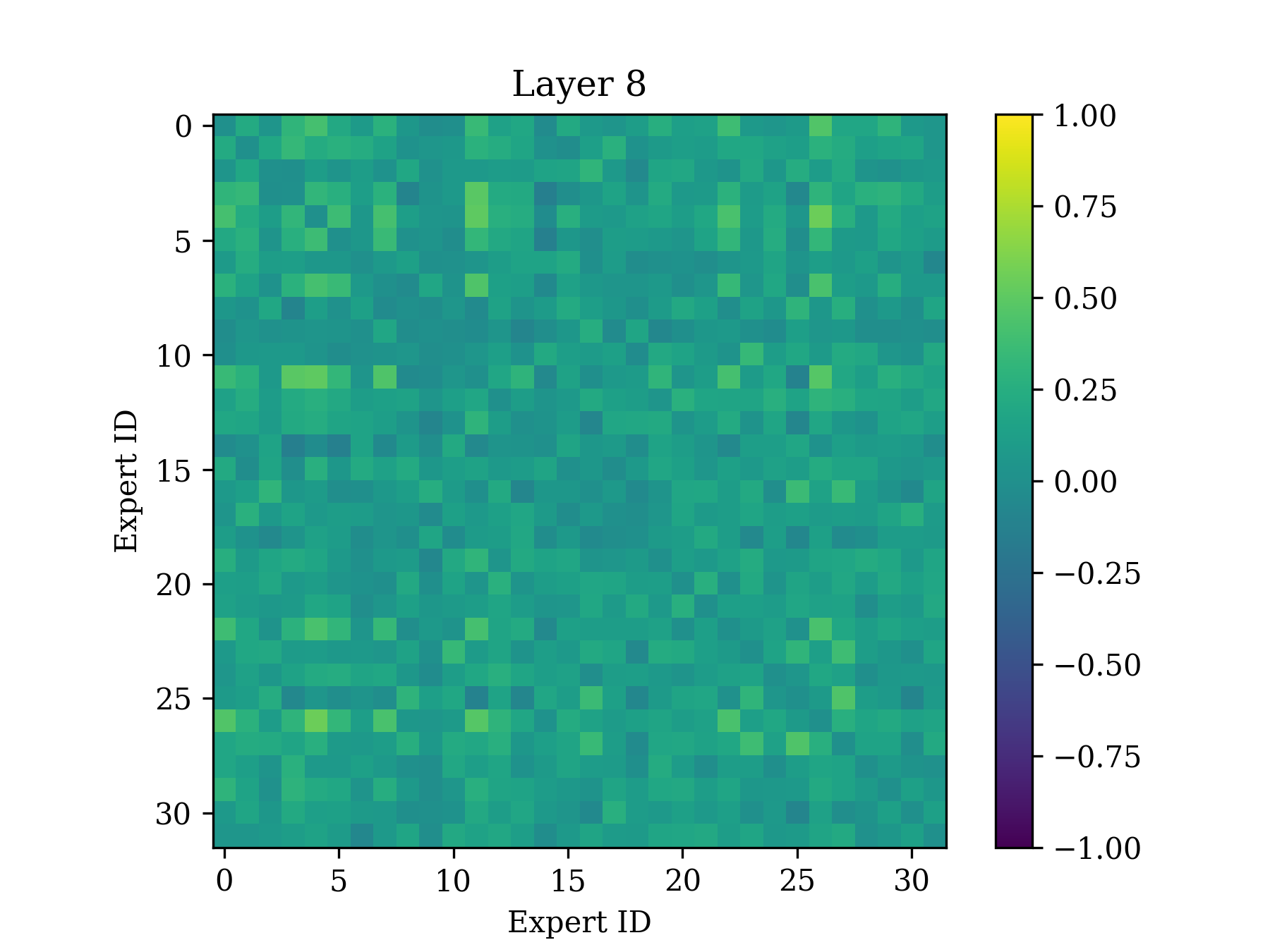}
    \caption{Similarities Between Default Vectors for 32c4 DefaultMoE at Layer 8.}
    \label{fig:appen-similarities-32c4-layer8}
\end{figure}

\begin{figure}[hbtp]
    \centering
    \includegraphics[width=0.5\linewidth]{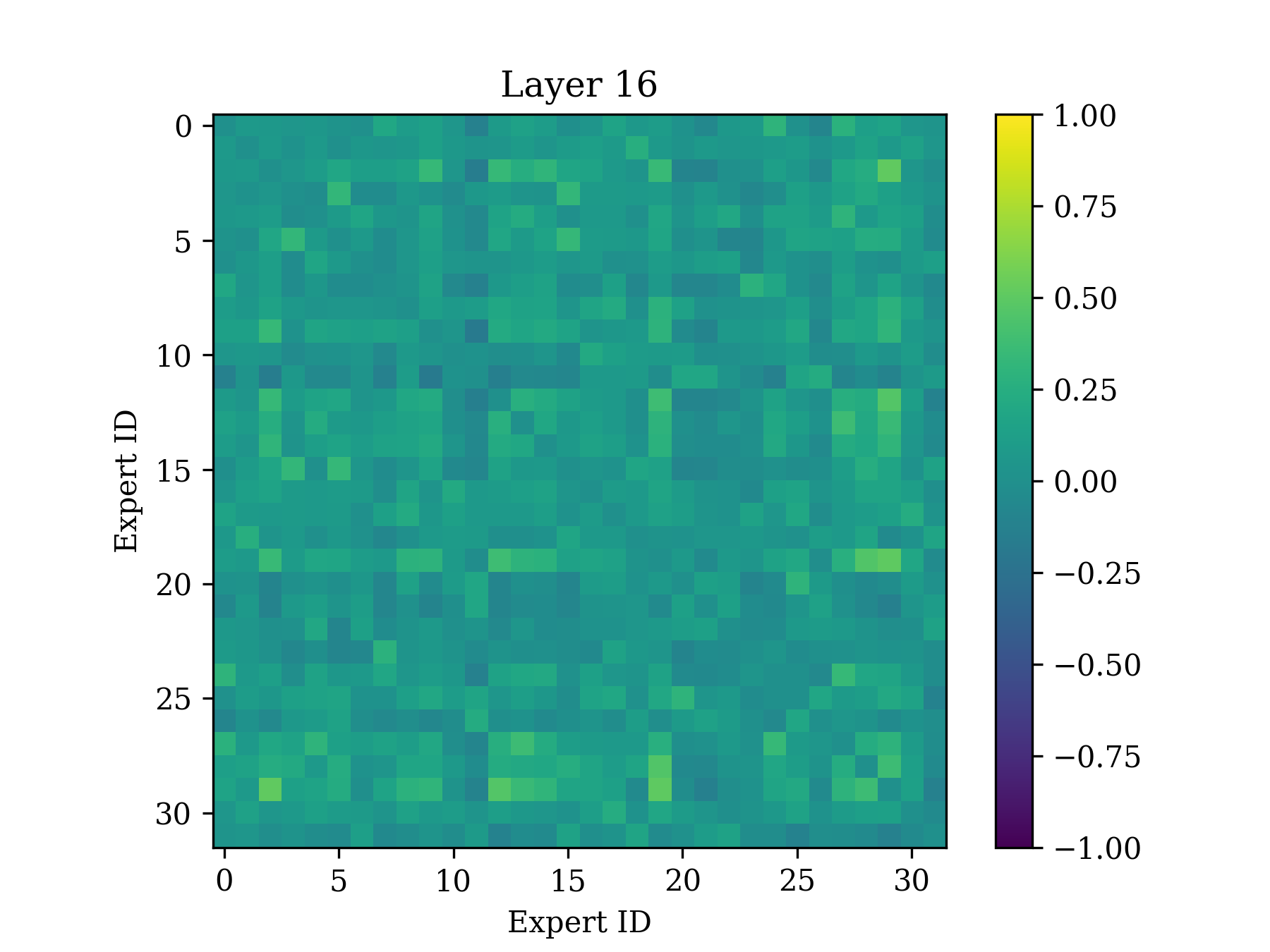}
    \caption{Similarities Between Default Vectors for 32c4 DefaultMoE at Layer 16.}
    \label{fig:appen-similarities-32c4-layer16}
\end{figure}

We plot the cosine similarity of default vectors in 8c1 DefaultMoE~(\cref{fig:appen-similarities-8c1-layer0}, \cref{fig:appen-similarities-8c1-layer8}, \cref{fig:appen-similarities-8c1-layer16}) and 32c4 DefaultMoE~(\cref{fig:appen-similarities-32c4-layer0}, \cref{fig:appen-similarities-32c4-layer8}, \cref{fig:appen-similarities-32c4-layer16}). All models are pretrained on FineWeb-Edu for 160B tokens.

\subsection{Note on References}
To avoid an extremely long bibliography, we abbreviated the bibliography entries for very large teams of authors \citep{deepseekv3, jiang2024mixtralexperts, abdin2024phi3technicalreporthighly, deepseekai2024deepseekv2strongeconomicalefficient, geminiteam2024gemini15unlockingmultimodal, openai2024gpt4technicalreport, dubey2024llama3herdmodels, sun2024hunyuanlargeopensourcemoemodel}. Full author lists are available at the links provided in the bibliography.


\newpage
\clearpage
\section*{NeurIPS Paper Checklist}

\begin{enumerate}

\item {\bf Claims}
    \item[] Question: Do the main claims made in the abstract and introduction accurately reflect the paper's contributions and scope?
    \item[] Answer: \answerYes{} 
    \item[] Justification: The abstract and introduction clearly state the motivation, summarize methodology, and state the key contributions made in the paper.

\item {\bf Limitations}
    \item[] Question: Does the paper discuss the limitations of the work performed by the authors?
    \item[] Answer: \answerYes{} 
    \item[] Justification: The first section of the Appendix states limitations. Other limitations, regarding the assumptions made are also discussed throughout the paper.

\item {\bf Theory assumptions and proofs}
    \item[] Question: For each theoretical result, does the paper provide the full set of assumptions and a complete (and correct) proof?
    \item[] Answer: \answerNA{} 
    \item[] Justification: We do not provide theoretical results.

    \item {\bf Experimental result reproducibility}
    \item[] Question: Does the paper fully disclose all the information needed to reproduce the main experimental results of the paper to the extent that it affects the main claims and/or conclusions of the paper (regardless of whether the code and data are provided or not)?
    \item[] Answer: \answerYes{} 
    \item[] Justification: In the main paper, each experiment details the setup and datasets used. In addition, we have open sourced our code.

\item {\bf Open access to data and code}
    \item[] Question: Does the paper provide open access to the data and code, with sufficient instructions to faithfully reproduce the main experimental results, as described in supplemental material?
    \item[] Answer: \answerYes{} 
    \item[] Justification: We provide link to the code and also explain the experimental setup in the main body. The data used in our experiments is publicly available.

\item {\bf Experimental setting/details}
    \item[] Question: Does the paper specify all the training and test details (e.g., data splits, hyperparameters, how they were chosen, type of optimizer, etc.) necessary to understand the results?
    \item[] Answer: \answerYes{} 
    \item[] Justification: We describe experimental details in the main body of the paper and also in our open sourced code.

\item {\bf Experiment statistical significance}
    \item[] Question: Does the paper report error bars suitably and correctly defined or other appropriate information about the statistical significance of the experiments?
    \item[] Answer: w\answerYes{} 
    \item[] Justification: The benchmark results report the standard error given by lm-eval-harness. The experiments focus on large-scale language model pretraining, where each run is computationally expensive and typically conducted once per configuration. As is common in this setting, we report complete training and validation results over training rather than multiple runs. 

\item {\bf Experiments compute resources}
    \item[] Question: For each experiment, does the paper provide sufficient information on the computer resources (type of compute workers, memory, time of execution) needed to reproduce the experiments?
    \item[] Answer: \answerYes{} 
    \item[] Justification: The LLM experiments were run on 64 GPUs on the AWS cluster. Our code contains configuration files which can be used to estimate compute resources.
    
\item {\bf Code of ethics}
    \item[] Question: Does the research conducted in the paper conform, in every respect, with the NeurIPS Code of Ethics \url{https://neurips.cc/public/EthicsGuidelines}?
    \item[] Answer: \answerYes{} 
    \item[] Justification: 

\item {\bf Broader impacts}
    \item[] Question: Does the paper discuss both potential positive societal impacts and negative societal impacts of the work performed?
    \item[] Answer: \answerNA{} 
    \item[] Justification: This work is methodological in nature, proposing improvements to training efficiency and stability in Mixture-of-Experts models. As such, it does not directly engage with societal use cases or their potential impacts.
    
\item {\bf Safeguards}
    \item[] Question: Does the paper describe safeguards that have been put in place for responsible release of data or models that have a high risk for misuse (e.g., pretrained language models, image generators, or scraped datasets)?
    \item[] Answer: \answerNA{} 
    \item[] Justification: This work does not involve the release of pretrained models, datasets, or tools with direct misuse risk. It focuses on architectural and training improvements for LLMS trained on publicly available data.

\item {\bf Licenses for existing assets}
    \item[] Question: Are the creators or original owners of assets (e.g., code, data, models), used in the paper, properly credited and are the license and terms of use explicitly mentioned and properly respected?
    \item[] Answer: \answerYes{} 
    \item[] Justification: We do cite creators and original owners of assets in our paper.

\item {\bf New assets}
    \item[] Question: Are new assets introduced in the paper well documented and is the documentation provided alongside the assets?
    \item[] Answer: \answerNA{} 
    \item[] Justification: No new assets are introduced as part of this work.

\item {\bf Crowdsourcing and research with human subjects}
    \item[] Question: For crowdsourcing experiments and research with human subjects, does the paper include the full text of instructions given to participants and screenshots, if applicable, as well as details about compensation (if any)? 
    \item[] Answer: \answerNA{} 
    \item[] Justification: The paper does not involve crowdsourcing nor research with human subjects.

\item {\bf Institutional review board (IRB) approvals or equivalent for research with human subjects}
    \item[] Question: Does the paper describe potential risks incurred by study participants, whether such risks were disclosed to the subjects, and whether Institutional Review Board (IRB) approvals (or an equivalent approval/review based on the requirements of your country or institution) were obtained?
    \item[] Answer: \answerNA{} 
    \item[] Justification: The paper does not involve crowdsourcing nor research involving human subjects.

\item {\bf Declaration of LLM usage}
    \item[] Question: Does the paper describe the usage of LLMs if it is an important, original, or non-standard component of the core methods in this research? Note that if the LLM is used only for writing, editing, or formatting purposes and does not impact the core methodology, scientific rigorousness, or originality of the research, declaration is not required.
    \item[] Answer: \answerNA{} 
    \item[] Justification: The research presented in this manuscript does not involve the use of LLMs as an important, original, or non-standard component. 

\end{enumerate}

\end{document}